%% file: manuscript.tex
\documentclass[pmlr, nosubfloats]{jmlr}

\usepackage{longtable}

 \usepackage{booktabs}
 
\usepackage[load-configurations=version-1]{siunitx} 

\input{requirements}

\input{notation}
\makeatletter
\def\set@curr@file#1{\def\@curr@file{#1}} 
\makeatother

\theorembodyfont{\upshape}
\theoremheaderfont{\scshape}
\theorempostheader{:}
\theoremsep{\newline}

\makeatletter
\def\blfootnote{\xdef\@thefnmark{}\@footnotetext}
\makeatother

\title[FedICU: Federated Learning in Multi-Center Critical Care Research]{Federated Learning in Multi-Center Critical Care Research: A Systematic Case Study using the eICU Database}

\author{\Name{Arash Mehrjou\textsuperscript{*}}
        \Email{arash.x.mehrjou@gsk.com}\\ 
        \addr GlaxoSmithKline, Artificial Intelligence \& Machine Learning\\
        \addr Max Planck Institue for Intelligent Systems, T\"ubingen, Germany\\
        \addr ETH Z\"urich, Z\"urich, Switzerland
        \AND
        \Name{Ashkan Soleymani\textsuperscript{*}}
        \Email{ashkanso@mit.edu}\\ 
        \addr Massachusetts Institute of Technology, Cambridge, US\\
        \vspace{-0.4cm} \AND
        \Name{Annika Buchholz}
        \Email{annika.buchholz@tuebingen.mpg.de}\\ 
        \addr Max Planck Institue for Intelligent Systems, T\"ubingen, Germany\\
        \vspace{-0.4cm} \AND
        \Name{Jürgen Hetzel}
        \Email{juergen.hetzel@web.de}\\ 
        \addr Department of Pneumology, Kantonsspital Winterthur, Winterthur, Switzerland\\
        \vspace{-0.4cm} \AND
        \Name{Patrick Schwab}
        \Email{patrick.x.schwab@gsk.com}\\ 
        \addr GlaxoSmithKline, Artificial Intelligence \& Machine Learning\\
        \vspace{-0.4cm} \AND
        \Name{Stefan Bauer}
        \Email{baue@kth.se}\\ 
        \addr KTH, Stockholm, Sweden \\
        \addr GlaxoSmithKline, Artificial Intelligence \& Machine Learning\\
        }

\begin{document}

\maketitle

\vspace{-1.2cm}

\begin{abstract}
    Federated learning (FL) has been proposed as a method to train a model on different units without exchanging data. This offers great opportunities in the healthcare sector, where large datasets are available but cannot be shared to ensure patient privacy. We systematically investigate the effectiveness of FL on the publicly available eICU dataset for predicting the survival of each ICU stay. We employ Federated Averaging as the main practical algorithm for FL and show how its performance changes by altering three key hyper-parameters, taking into account that clients can significantly vary in size.
    We find that in many settings, a large number of local training epochs improves the performance while at the same time reducing communication costs. 
    Furthermore, we outline in which settings it is possible to
    have only a low number of hospitals participating in each federated update round. When many hospitals with low patient counts are involved, the effect of overfitting can be avoided by decreasing the batchsize.
    This study thus contributes toward identifying suitable settings for running distributed algorithms such as FL on clinical datasets.
    \blfootnote{\hspace{-0.5cm}$*$ Equal contributions}
\end{abstract}

\section{Introduction}
\label{sec:introduction}

The progress of the past years in the field of machine learning (ML) offers unprecedented opportunities for data analysis and interpretation. A potentially impactful domain of ML application is the healthcare sector, where firstly, large datasets are available. Secondly, the underlying relations are extremely complex, and thus ML methods can help reveal new insights and a more profound understanding.

However, patient data is particularly sensitive information that cannot simply be shared with third parties. Some governments have imposed restrictions on the amount of clients' data that can be exchanged among various sites (e.g., see White House report on the privacy of consumer data~\cite{house2012consumer} or the European data protection regulation~\cite{EUdataregulations2018}). Data anonymization offers certain possibilities to analyze data for research purposes but reaches its limits, e.g., \ when face images are investigated or perfect unidentifiability is required.
To overcome this difficulty, various methods that ensure privacy are being developed.
One of these approaches is the so-called Federated Learning (FL)~\cite{mcmahan2017communication} where the exchanged information is \emph{not} in the form of data but instead of model parameters or gradients.

In our case, a neural network that predicts a specific medical outcome (survival of the current ICU stay) can be trained in parallel on the datasets of different hospitals, updating the model parameters in each round which are then aggregated on a central server. This way, training on large datasets becomes possible. 
However, the inclusion of data from different hospitals is accompanied by the \emph{client shift} problem common to FL algorithms, which means that hospital data distributions may differ from each other and complicate the federated learning approach. In particular, too many local training epochs can lead to models that are inconsistent with each other. On the other hand, fewer local training epochs increase the number of communications between clients and the central server. 

We investigate this trade-off, and one of our key findings is that the model performance increases with a large number of local training epochs, especially when many large clients are involved in the federation. This suggests that the data distributions of different hospitals are not too different after all. We also show that the training can succeed if only a fraction of the clients is involved in the aggregation round. In this case, involving many small hospitals in the federation brings along the risk of overfitting, which can be compensated by reducing the batchsize in the training process.
Our approach identifies suitable conditions for FL on clinical data, which provides an essential basis for reliable classification algorithms or synthetic data generation.

\paragraph{Related work.} 
Federated learning on medical data has been investigated in previous work. Different subsets of the eICU dataset have been used to evaluate the performance of FL for the prediction of different medical outcomes by~\cite{pfohl2019federated} and ~\cite{huang2020loadaboost}, investigating privacy-preserving methods, see e.g.~\cite{beaulieu2018privacy}, or including clinical meaningful clustering by~\cite{huang2019patient}.
A benchmark for machine learning models on other clinical data is presented in ~\cite{sheikhalishahi2020benchmarking}. A more sophisticated model of FL on clinical data is presented by~\cite{zec2021specialized}. In this work, we investigate the effect of varying different hyper-parameters of the FL on the eICU dataset. Since eICU is an important dataset to emulate a real-world scenario where a joint model is trained among the hospitals of various sizes, the insights gained from this work can guide the choice of the critical hyper-parameters in a federated discriminative or generative algorithm.

\section{Dataset}
\label{sec:data_description}

We use the publicly available \textit{eICU Collaborative Research Database} by~\cite{eicu2018} for our presented analyses. The dataset contains over 200,000 deidentified submissions to intensive care units (ICU) in the United States across 208 hospitals in 2014 and 2015. The dataset consists of separate tables that contain different categories of information. The smallest unit of information is a single ICU stay that takes a unique identification number (ID) called~\emph{patientUnitStayID}. The information of all tables can be merged into a single dataset using this ID. 
In the analysis, the ID that identifies the hospital of the ICU stay can be included or ignored, which allows for the comparison of local, federated, and global models.
For the survival prediction task, we use the \emph{Apache variables} by~\cite{knaus1985apache} together with \emph{age} from the~\emph{patient} table in the eICU dataset.

\section{Method}

\label{sec:method}

A popular learning scenario in privacy-sensitive or large-scale distributed settings is~\emph{federated learning} (FL), where the data originates from different sources to train a centralized model. Each local source of data is called a~\emph{client}, and a~\emph{server} often hosts the centralized model. In applications such as mobile devices, the communication cost between the clients and the server becomes a major challenge. However, in the application of this paper's interest, the number of clients (hospitals) does not exceed a few hundred. The major challenge, though, is the efficient use of the available data in all hospitals without compromising privacy. The conceptual idea of FL is that a central model is transmitted to the clients, each client updates the received model with its locally available data, and the local updates are sent back to the server and aggregated to make a new central model.

\def\NoNumber#1{{\def\alglinenumber##1{}\State #1}\addtocounter{ALG@line}{-1}}
\begin{algorithm*}[!t]
    \caption{Federated Averaging (\texttt{FedAvg})}
    \label{alg:canonical_fed_learning}
    \hspace*{\algorithmicindent} \textbf{Input:} $N$: Total number of clients, $\{S_1, \ldots, S_N\}$:  Local datasets available to the clients, $E$: Number of training epochs on each local client, $\mathtt{batchsize}$: The size of local mini-batches used for training the client models, $\eta$: Learning rate, $C$: The ratio of all clients who participate at each round,  $\ell$: Loss function, Hypothesis set $\mathcal{H}=\{h(\cdot;\theta)\}$ parameterized by the weight vector $\theta$.\\
    
    \textbf{TrainClient}$(k, \theta):$
    \begin{algorithmic}[1]
        \STATE $\mathcal{B} \leftarrow$ Divide $S_k$ into batches of size $\texttt{batchsize}$
        \FOR{each epoch $i:1$ to $E$}
        \FOR{each batch $b\in\mathcal{B}$}
        \STATE $\theta\leftarrow \theta-\frac{\eta}{|b|}\sum_{(x, y)\in b} \nabla\ell(h(x;\theta), y)$
        \ENDFOR
        \ENDFOR
    \STATE return $\theta$
    \end{algorithmic}
    
    \texttt{\\}
    \textbf{Server executes:}
    \begin{algorithmic}[1]
        \WHILE{Stopping criterion not satisfied}
        \STATE $n\leftarrow \min(NC, 1)$
        \STATE $\pi_r\leftarrow$ (random set of $n$ clients)
        \FOR{each client $k\in \pi_r$ \textbf{in parallel}}
            \STATE $\theta^k_{r+1}\leftarrow$ \textbf{TrainClient}$(k, \theta_r)$
            \STATE $\theta_{r+1}\leftarrow 1/n \sum_{k=1}^n \theta^k_r$
        \ENDFOR
        \ENDWHILE

    \end{algorithmic}

\end{algorithm*}

 Based on the goal of training and assumptions about the clients, different training strategies have been devised.  
 A famous aggregation method is called federated averaging (\texttt{FedAvg})~\cite{mcmahan2017communication} that learns the local models for a few epochs $E$, then averages them to update the central model. The pseudocode of the~\texttt{FedAvg} method is presented in~\Cref{alg:canonical_fed_learning} (see \cite{kairouz2019advances} for an in-depth overview of federated learning topics.).
 
 Theoretically speaking, for arbitrarily distant clients' distributions, the performance of the model obtained by \texttt{FedAvg} can become arbitrarily poor. This effect, known as~\emph{client shift}, exacerbates when the local clients are trained for too long at each round of FL (see~\cite{goodfellow2014qualitatively}). On the other hand, excessive under-training of the local models necessitates too many rounds of FL, which incurs a considerable communication cost. We investigate this trade-off in the eICU dataset where the clients are different hospitals.

When the data distribution of clients are not the same, the target distribution is a mixture of the client's data distributions as $\mathcal{D}^\tau=\sum_{k=1}^N \frac{m_k}{m}\mathcal{D}_k$, where $m_k$ and $\mathcal{D}_k$ are the dataset size and the data distribution of the client $k$. The sum of the size of local datasets is represented by $m$, and $N$ shows the number of clients (hospitals). This
makes the learned hypothesis $h_{\hat{\mathcal{D}}^\tau}$ favor clients with larger sample sizes. This can be restrictive in our application, where we assume distinct data distributions across hospitals. 

There are multiple reasons to assume some difference among the hospital data distributions in the eICU dataset. i) The hospitals are geographically distributed in a large area. Consequently, hospitals are likely to be exposed to patients with different demographic and clinical features. ii) Even in nearby hospitals, the equipment, and measurement facilities can vary between ICU units of distinct hospitals, resulting in an uneven distribution of measured features. iii) Even though the hospitals support the range of diseases present in the dataset, some of them may be specialized in some topics that allow them to measure and diagnose the health condition of the patients more extensively and with higher resolution. All these conditions result in hospitals with different distributions of patients and diseases that necessitate the non-IID assumption among clients in FL.
However, the experimental results in~\Cref{sec:experiments_canonical_fedlearning} indirectly imply that the difference between hospital data distributions is not too large within the eICU dataset.  

There are a few hyper-parameters in a FL algorithm that need to be chosen carefully. In the experiments, we investigate the effect of the number of epochs for which every client is trained at each round of federated learning ($E$), the ratio of the clients that participate in federated learning at each round ($C$), and the batchsize used when training the local models ($B$). 

\section{Results}
\label{sec:experiments_canonical_fedlearning}
We consider a prediction task where the survival of the patients in an ICU admission is predicted given the Apache variables~\cite{knaus1985apache} together with the patient's age. An important factor in the success or failure of FL is the cohort of clients that participate in FL rounds. To study the effect of this factor alongside other factors, every experiment is repeated for a number of cohorts of hospitals whose size falls within a number of patients of $\texttt{Lower\_Size}\in \{5,10,50,100,500,1000\}$ and $\texttt{Upper\_Size}\in \{50,500,1000,5000\}$.
Each choice of cohorts is called a~\emph{scenario} which are detailed in \cref{tab:scenarios} in App.~\ref{app:scenarios}. In principle, the purpose of defining various scenarios of this kind is to make sure the relative size of the clients is taken into account while studying the outcome of a FL process.

 The trained model is evaluated on the test dataset, which is 30\% of the full dataset
 and is left out during the rounds of FL. We use a 3-layer MLP as the classifier, binary cross-entropy as the loss function, and the Adam optimizer with a learning rate of 0.0001 for training.

\paragraph{On the computational barriers to FL simulations} Simulating distributed computations is known to be computationally expensive since all the local computations at clients alongside the central computations at the server should be simulated at the same time~\cite{attiya2004distributed}. Meanwhile, a significant amount of runtime for computer programs is because of the I/O overhead of memory access instructions~\cite{mano1993computer}. Graphics processing units (GPU) and their memory access overheads are not an exception to this~\cite{kim2014gpudmm}. In the case of FL, at each round, all of the client and server models should be fetched from memory, updated, and saved back to the memory, which causes each FL round to be prohibitively expensive even for the most simplistic algorithmic settings. To illustrate the case, let's consider the most covering scenarios (5) and (18). Each round updates roughly $\sim200$ models, each model taking between 5 to 10 minutes, amounts to $\sim25$ hours totally for a single round. Accordingly, each experiment alone takes a couple of days, e.g., the corresponding experiment for~\cref{fig:bc1_subfig15} takes up to 2 weeks for each setting (10 rounds). Thus, simulating the FL algorithms for realistic scenarios is very challenging to do in a sequential setting. 

We overcome this barrier by parallelizing the computation: Having access to hundreds of computation nodes, we use the MapReduce framework~\cite{dean2008mapreduce} adjusted with fault-tolerant techniques in our high-level development of the FL algorithms. In our MapReduce FL framework, at each round, the central computation node allocates an auxiliary computation node for each client and provides the data, model, and necessary instructions as our Map phase. By synchronization techniques, the central computation node will be notified of the termination of auxiliary nodes and assess their outputs to avoid possible errors. Afterward, the central node aggregates the client's results during the Reduce phase and moves to the next round. This way, using 200 computation nodes, the corresponding experiment for~\cref{fig:bc1_subfig15} is conducted in a day and a half. These computational barriers, both in connection with time and scale of computational resources, make it necessary to thoroughly investigate different aspects of classical FL algorithms on the dataset to provide a firm standpoint for future works.

\subsection{Investigate the effect of the number of local epochs  $E$}

\begin{figure}[!t]
    \centering
    \begin{minipage}{0.5\textwidth}
        \centering
        \includegraphics[width=\linewidth]{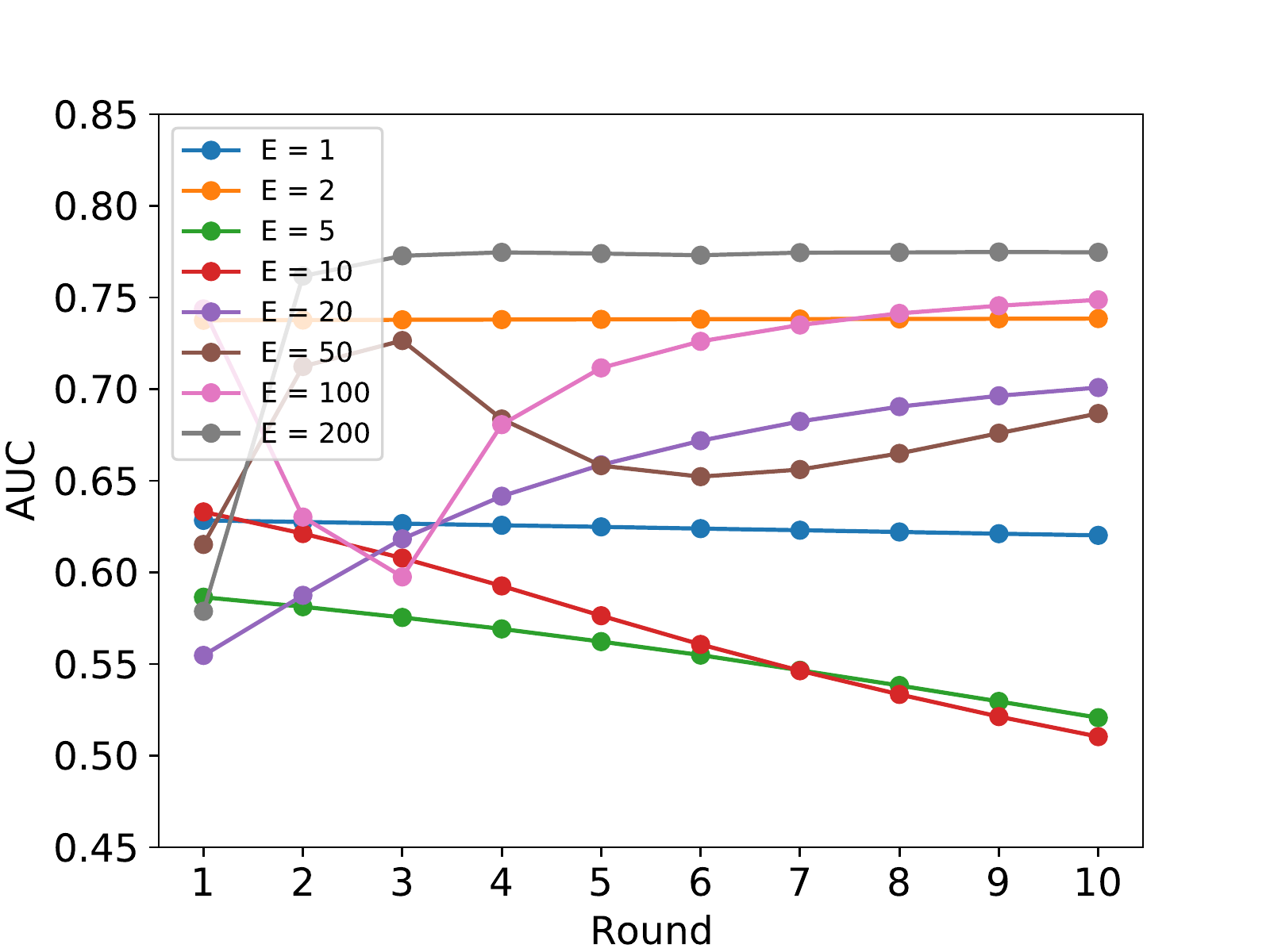}
        \label{fig:e_main1_subfig1}
    \end{minipage}%
    \begin{minipage}{0.5\textwidth}
        \centering
        \includegraphics[width=\linewidth]{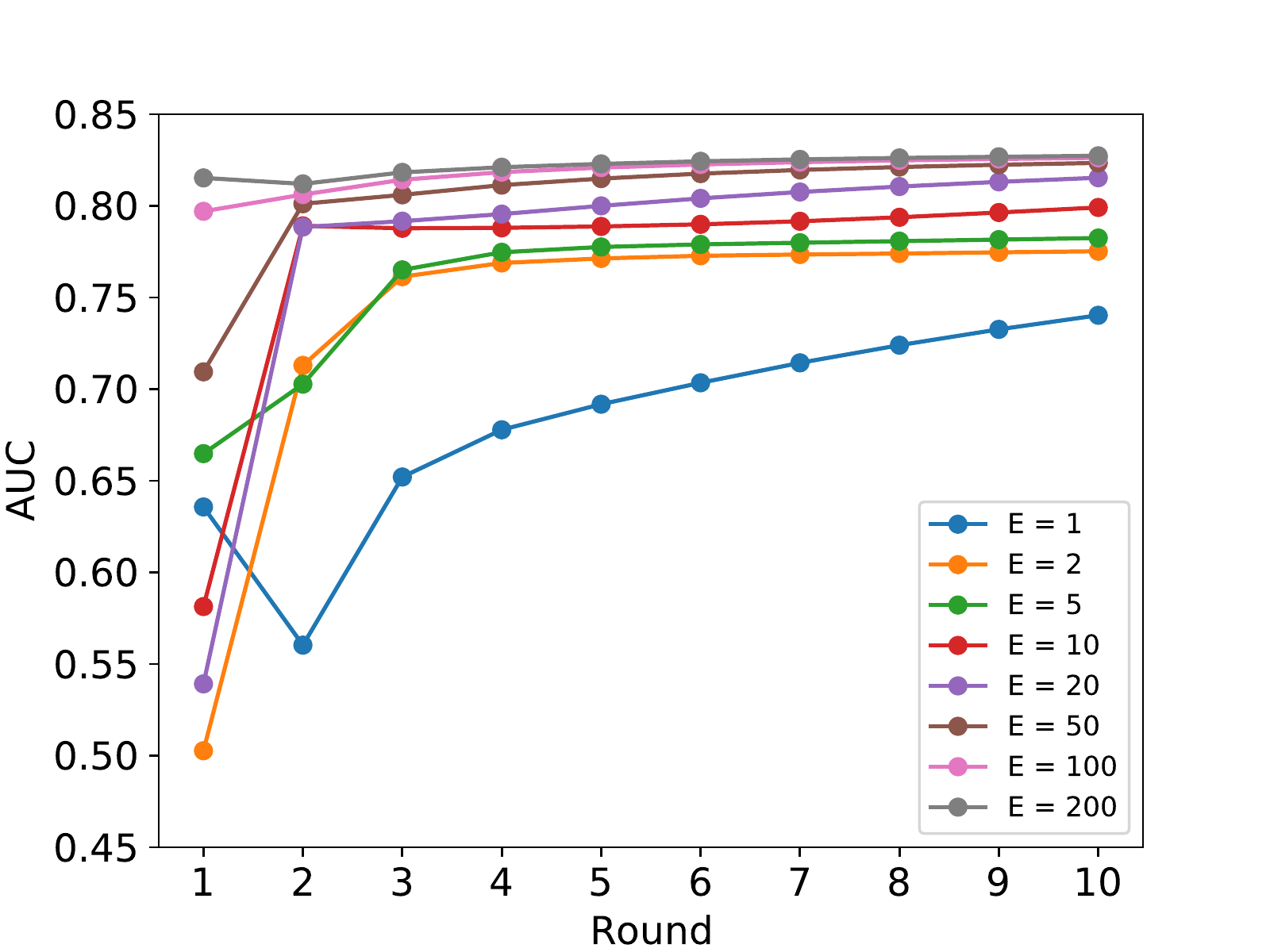}
       \label{fig:e_main1+subfig2}
    \end{minipage}
    \caption{The effect of the number of local training epochs $E$ is illustrated for two different scenarios: Left: including only small hospitals (scenario (1) in table~\ref{tab:scenarios}). Right: mixing small and large hospitals (scenario (5)) 
    }
    \label{fig:e_main}
\end{figure}

As mentioned earlier, it is hypothesized that there would be a trade-off between the depth of training on each client's dataset that results in a fewer number of communication rounds in FL and the risk of client shift that biases the overall learned model. Here, we investigate this effect on the eICU dataset for the Electronic Health Record (EHR) data distributed among hospitals. We employ the \texttt{FedAvg} algorithm and evaluate the final model on a 30\% fraction of 
the dataset that did not participate in the training of the clients. We consider scenarios (1) to (15) each of which corresponds to a choice of $(\texttt{Lower\_Size}, \texttt{Upper\_Size})$ on the size of the hospital datasets. The numerous choices for the values of these bounds cover scenarios such as (1) cohorts with only small hospitals, (15) cohorts with only large hospitals, and (5) cohorts with a mixture of small and large hospitals. The experiments suggest that generally, increasing the number of local training epochs improves the prediction performance of the model measured by the area under the curve (AUC) (See \cref{fig:e_main} and other scenarios presented in App.~\ref{app:plots}, \Cref{fig:e}). If the clients had considerably different distributions, extensive local training could deteriorate the final aggregated model. However, this effect was not observed by increasing $E$, possibly because the distribution of data across different hospitals is not too different. This suggests that a great number of communication rounds can be saved by increasing the depth of training for each local model without being worried about getting biased towards small local datasets.

\subsection{Investigate the joint effect of the fraction of clients $C$ and the batchsize $B$}

\begin{figure}[!h]
    \centering
    \begin{minipage}{0.5\textwidth}
        \centering
        \includegraphics[width=\linewidth]{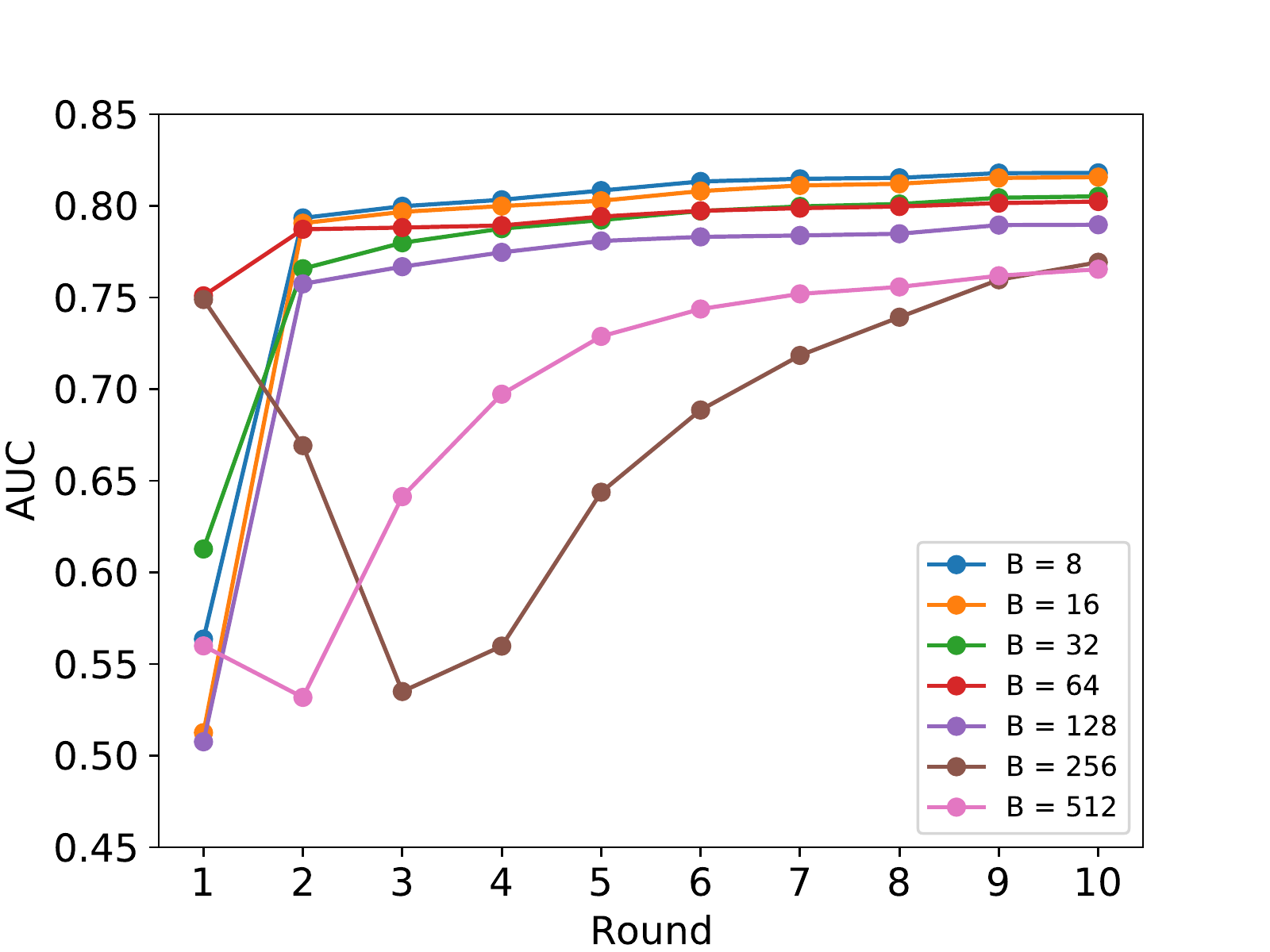}
        \label{fig:bc_main1_subfig1}
    \end{minipage}%
    \begin{minipage}{0.5\textwidth}
        \centering
        \includegraphics[width=\linewidth]{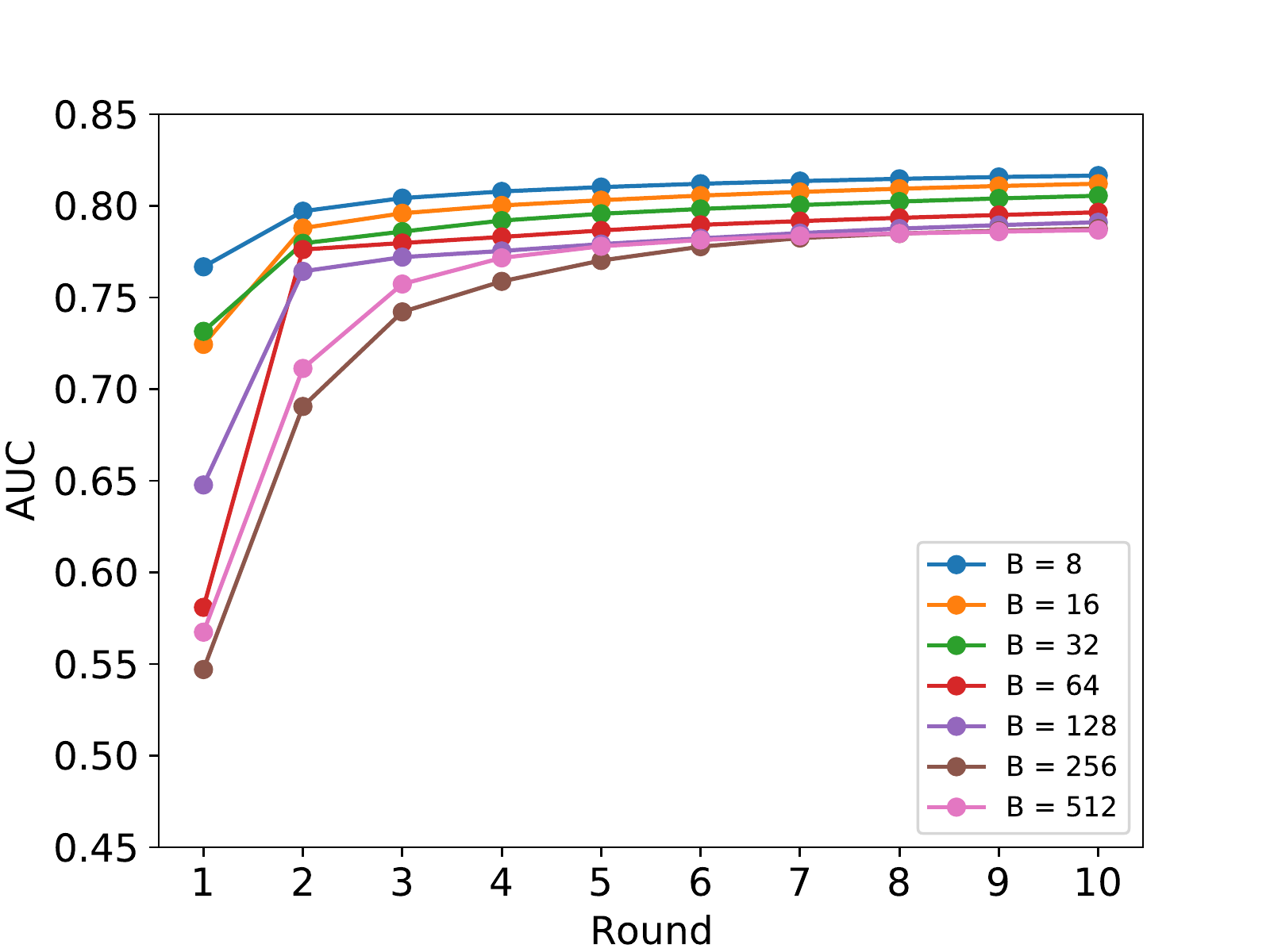}
       \label{fig:bc_main1_subfig2}
    \end{minipage}
    \caption{The effect of different batchsizes $B$ and the fraction of clients involved in each aggregation round with $C=0.2$ (left) and $C=1$ (right) for Scenario (17), a mixture of small and medium-sized hospitals.}
    \label{fig:bc_main1}
\end{figure}

\begin{figure}[!t]
    \centering
    \begin{minipage}{0.5\textwidth}
        \centering
        \includegraphics[width=\linewidth]{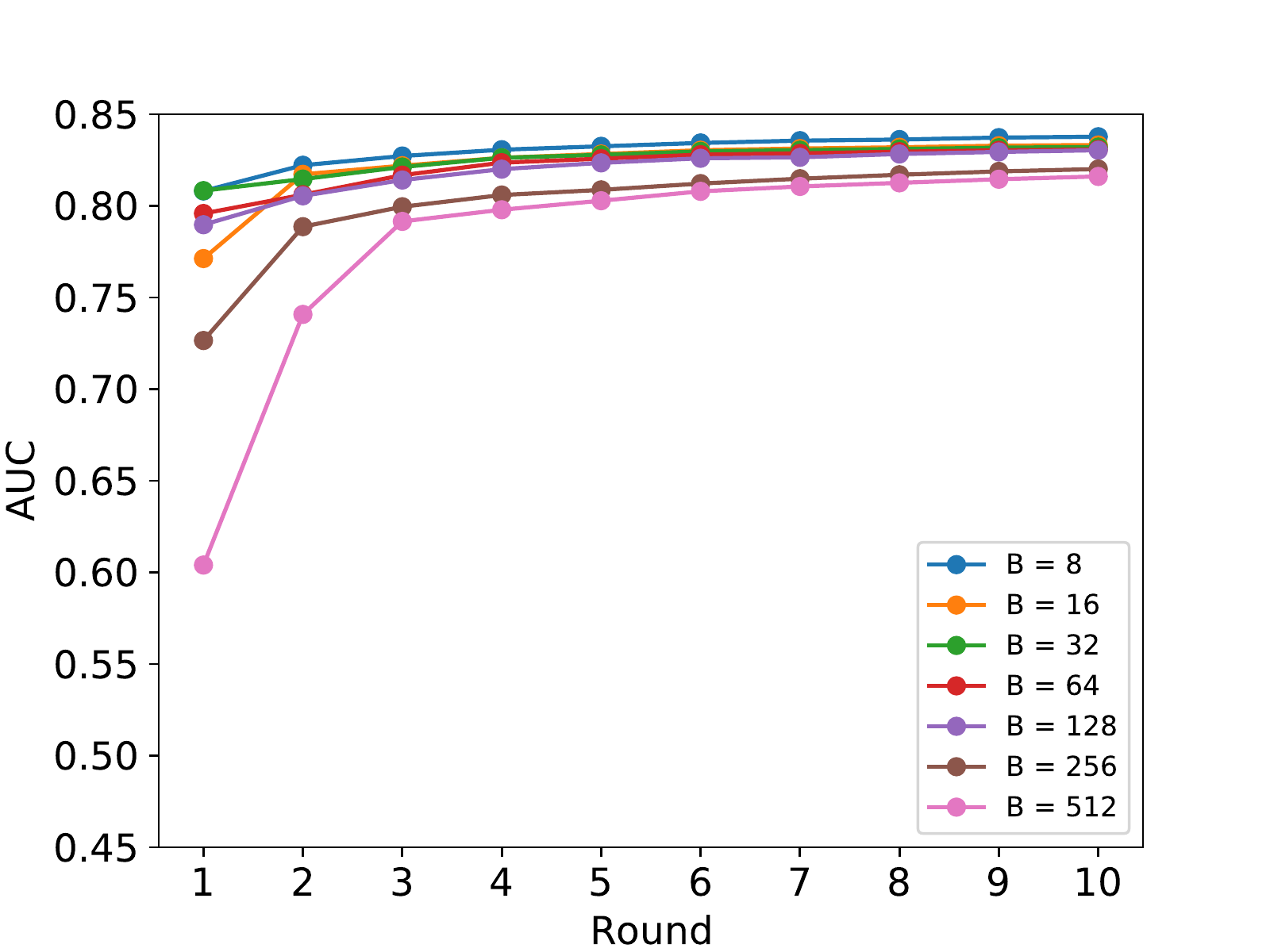}
        \label{fig:bc_main2_subfig1}
    \end{minipage}%
    \begin{minipage}{0.5\textwidth}
        \centering
        \includegraphics[width=\linewidth]{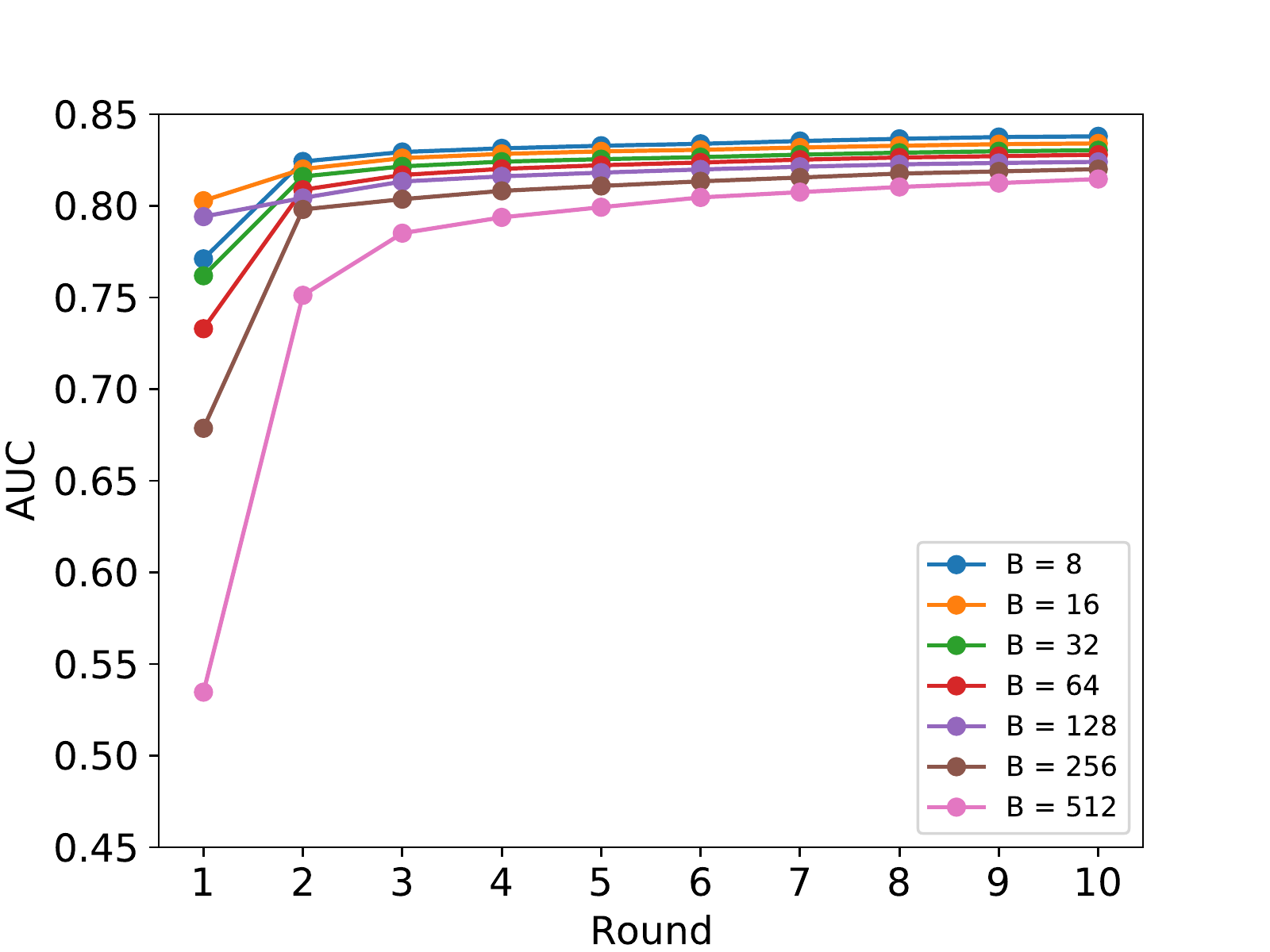}
       \label{fig:bc_main2_subfig2}
    \end{minipage}
    \caption{The effect of different batchsize $B$ and the fraction of clients involved in each aggregation round with $C=0.2$ (left) and $C=0.4$ (right) for Scenario (14), hospitals with large patient counts.}
    \label{fig:bc_main2}
\end{figure}

Choosing a ratio $C<1$ of the clients for each round of FL is a source of randomness that occurs at the level of federated update rounds. Another source of randomness that occurs at the level of the training of each client is batchifying data for SGD-type optimization algorithms. These sources of randomness should be carefully harnessed to improve the performance of the output model of a FL process. We observe that for cohorts consisting of smaller hospitals (5-500), larger $C$ leads to improved performance almost regardless of the batchsize (See \cref{fig:bc_main1} and the other scenarios presented in App.~\ref{app:plots},  \cref{fig:bc1,fig:bc2}). Therefore, if the hospitals own small datasets, it is best to include more of them in each round of federated updates. The effect of $C$ becomes less important when the cohort consists of hospitals with larger datasets. For example it can be seen in \cref{fig:bc_main2_subfig1,fig:bc_main2_subfig2,fig:bc_main2_subfig1}. that for different values of $C\in\{0.2, 0.4\}$, the performance is equally good when the cohort consists of hospitals whose datasets' sizes are within (500, 5000). In the heterogeneous scenario where the cohort is a mixture of large and small hospitals whose datasets' sizes are within (5, 5000), the choice of batchsizes becomes important for smaller values of $C$ (\cref{fig:bc_main2_subfig2}), especially in the middle rounds of federated updates. This observation is expected in a heterogeneous scenario because of the following reason: When $C$ is large, at every round of federated update, both small and large hospitals are likely to be chosen and contribute to the update of that phase. Therefore, if a choice of batchsize is bad for small hospitals, its effect will be counteracted by the contribution of the large hospital. However, when $C$ is smaller, it's likely that in some rounds, only small hospitals are chosen, and there will be no large hospital to counteract the contribution of their overfitted model to the update of that round. This effect will be magnified for the scenarios in that only small hospitals are chosen. The takeaway message would be: In cohorts that consist of small or small+large hospitals, try to use more hospitals in each round of federated updates. If the participation ratio cannot be increased due to technical reasons, choose smaller batchsizes for training to decrease the chance of the models trained on small hospitals to overfit. This will of course come with the cost of slower convergence and requires more rounds of federated updates.

\subsection{Conclusion}
\label{sec:conclusion}
We used the eICU dataset as a benchmark for FL as a privacy-preserving method. The critical hyper-parameters of the \texttt{FedAvg} algorithm are studied on this dataset and evaluated on the performance of predicting the survival probability of each ICU stay.
It was observed that large numbers of local training epochs improve the performance of the FL without overfitting the local datasets. The effect is particularly strong if hospitals with low patient counts are included. This insight allows for reducing the number of communication rounds.
The joint effect of the fraction of clients in the training of each federated update round $C$ and the size of the local batchsizes turns out to depend on the homogeneity of the federation: for scenarios where clients with large patient counts dominate, the performance does not strongly depend on $C$ and the chosen batchsizes. However, if many clients with low patient counts are included, our results indicate that one should either pick a large number for $C$ or compensate by choosing smaller batchsizes for training to avoid overfitting on small units.

Our contributions aim to improve and accelerate the efficient use of FL in medical research. Nevertheless, there are limitations to the generalizability of our work. Although the considered dataset is a very extensive one, its representativeness is expected to have its limits for various reasons. For example, hospitals with a lack of digitization are not represented and may differ in their technical equipment and thus also in their data distribution. Also, if hospitals from different countries were combined in a federation, we would expect a much more significant client shift, which would also alter our results.

Studying the performance of other FL aggregation methods, such as \texttt{FedSGD}~\cite{mcmahan2017communication}, and also investigating the similar effects on other EHR datasets such as \cite{Johnson2021mimic4} are postponed to future work.

\bibliography{refs}

\newpage
\appendix

\input{appendix}

\end{document}

%% file: requirements.tex
\usepackage{xr-hyper}
\usepackage{hyperref}
\usepackage{url}
\usepackage{graphicx}

\usepackage{amsfonts,amssymb,amsmath}
\usepackage{graphicx}
\usepackage[english]{babel}
\usepackage{numprint}
\usepackage{float}
\usepackage{placeins}

\usepackage{xr-hyper}
\usepackage{hyperref}     
\usepackage[capitalise]{cleveref}
\usepackage{subcaption}
\usepackage{xstring}
\usepackage{setspace}
\usepackage{microtype} 
\usepackage{booktabs}  
\usepackage{longtable} 
\usepackage{rotating}
\usepackage{fancyhdr}

\usepackage{multirow}
\usepackage{colortbl}
\usepackage{pgfplotstable}
\usepackage{bm}
\usepackage[utf8]{inputenc}
\usepackage{csquotes}
\usepackage{lipsum}
\usepackage{cleveref}
\usepackage{textcomp}
\usepackage{algorithm}
\usepackage{algorithmic}
\usepackage{tikz}
\usetikzlibrary{matrix,fit}
\usepackage{breqn}

\usepackage{amsmath}
\usepackage{amssymb}
\usepackage{amsfonts}
\usepackage{amssymb}
\usepackage[mathscr]{euscript}
\usepackage{subfiles}
\usepackage[utf8]{inputenc}

%% file: appendix.tex
\section{Scenarios of hospital sizes in the federation}\label{app:scenarios}
\begin{table}[h]

	\renewcommand{\arraystretch}{1.3}

	\caption{The number of hospitals chosen for every pair of (\texttt{Lower\_Size}, \texttt{Upper\_Size}) and the distributional information of how the size of data is scattered among them. The columns $\mu$ and $\sigma$ show the mean and standard deviation of the size of hospitals' datasets in the cohort. Larger $\mu$ suggest that more populated hospitals are present in the cohorts and larger $\sigma$ suggests that the heterogeneity of the population sizes in the selected cohort is higher. The column $n$ shows the number of hospitals in each cohort.} 
	\label{tab:scenarios}
	\centering

	\footnotesize
    \begin{tabular}{ |l||>{\centering\arraybackslash}p{2.4cm}|>{\centering\arraybackslash}p{2.4cm}|>{\centering\arraybackslash}p{1.55cm}|>{\centering\arraybackslash}p{1.2cm}|>{\centering\arraybackslash}p{1.55cm}|}
        \hline
        \multicolumn{6}{|c|}{Data Scenarios} \\
        \hline
        Scenario & \texttt{Lower\_Size ($l$)} & \texttt{Upper\_Size ($u$)} & $n$ & $\mu$ & $\sigma$\\
        \hline
        \hline
        Scenario (1)   & 10 & 50 & 19 & 24.95 & 13.42\\
        Scenario (2)   & 10 & 100 & 29 & 39.41 & 24.14\\
        Scenario (3)   & 10 & 500 & 103 & 210.42 & 139.57\\
        Scenario (4)  & 10 & 1000 & 148 & 364.62 & 273.76\\
        Scenario (5)    & 10 & 5000 & 202 & 813.06 & 932.94\\
        Scenario (6)   & 50 & 100 & 10 & 66.90 & 13.93\\
        Scenario (7) & 50 & 500 & 84 & 252.37 & 119.60\\
        Scenario (8) & 50 & 1000 & 129 & 414.65 & 257.80\\
        Scenario (9) & 50 & 5000 & 183 & 894.89 & 943.16\\
        Scenario (10) & 100 & 500 & 74 & 277.43 & 104.56\\
        Scenario (11) & 100 & 1000 & 119 & 443.87 & 247.01\\
        Scenario (12) & 100 & 5000 & 173 & 942.75 & 948.18\\
        Scenario (13) & 500 & 1000 & 45 & 717.58 & 151.32\\
        Scenario (14) & 500 & 5000 & 99 & 1440.06 & 992.32\\
        Scenario (15) & 1000 & 5000 & 54 & 2042.13 & 994.34\\
        Scenario (16) & 5 & 50 & 20 & 24.15 & 13.54\\
        Scenario (17) & 5 & 500 & 104 & 208.48 & 140.28\\
        Scenario (18) & 5 & 5000 & 203 & 809.10 & 932.35\\
        \hline
    \end{tabular}
\end{table}

\section{Additional Plots}\label{app:plots}
Additional plots on effect of $E$ for all possible scenarios are given in~\Cref{fig:e}. Further plots regarding effects of $B$ and $C$ for all possible scenarios are provided in~\Cref{fig:bc1,fig:bc2}.

\input{figures}

%% file: figures.tex
 \begin{figure*}
    \vspace{-1.5cm}
        \centering
        \makebox[0.6\paperwidth]{%
            \begin{tikzpicture}[ampersand replacement=\&]
            \matrix (fig) [matrix of nodes, row sep=-1.1em, column sep=-3em]{ 
                \begin{subfigure}{0.35\columnwidth}
                    \centering
                    \resizebox{\linewidth}{!}{
                        \begin{tikzpicture}
                            \node (img)  {\includegraphics[width=\textwidth]{figs/Effect_of_E/predict_death_from_apache/10/50/dense.pdf}};
                        \end{tikzpicture}
                    }
                    \vspace{-0.7cm}
                    \caption{\scriptsize Scenario (1) \\$ (l = 10, u = 50)$} 
                    \label{fig:e_subfig1}
                \end{subfigure}
                \&
                \begin{subfigure}{0.35\columnwidth}
                    \centering
                    \resizebox{\linewidth}{!}{
                        \begin{tikzpicture}
                            \node (img)  {\includegraphics[width=\textwidth]{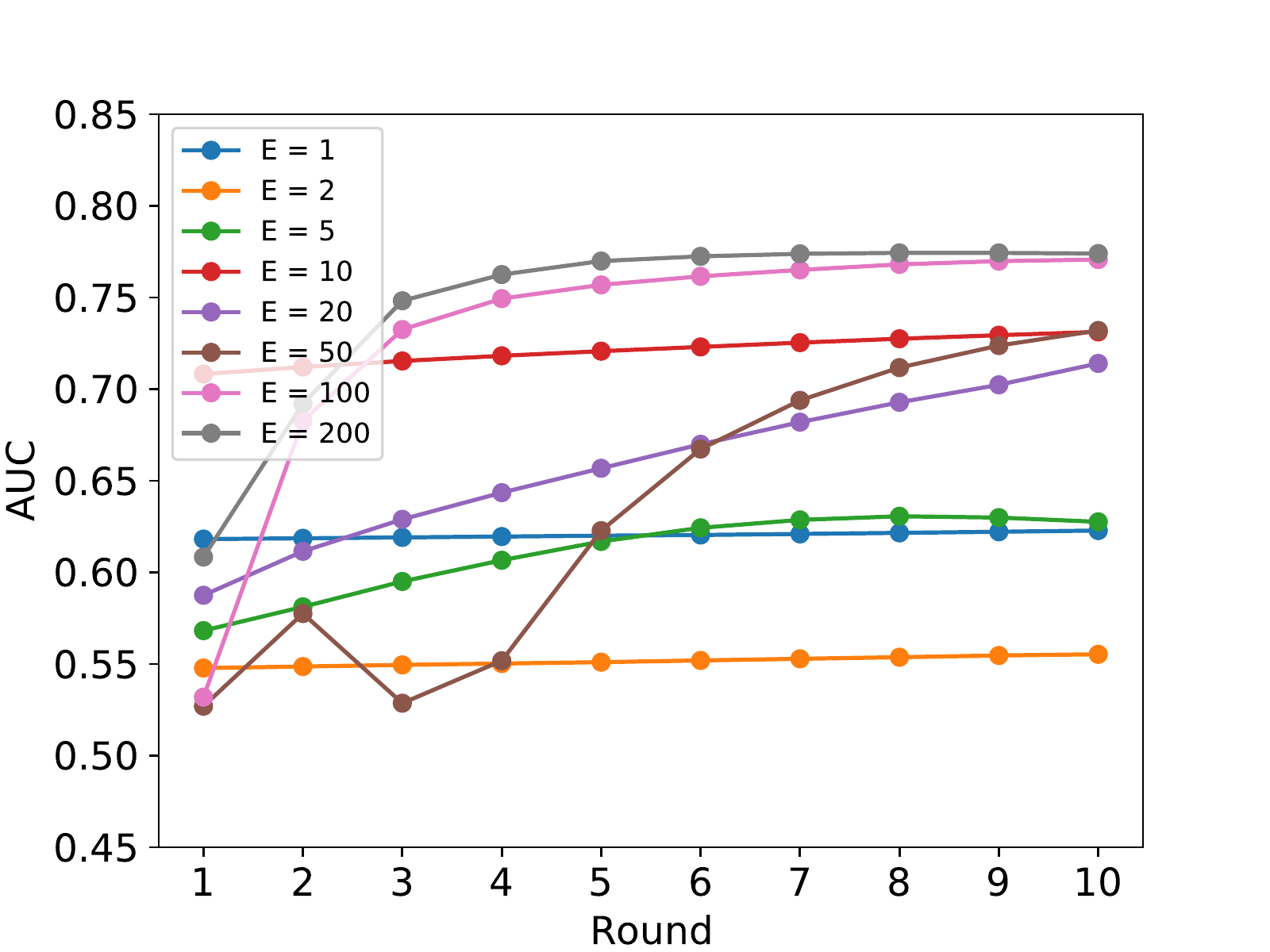}};
                        \end{tikzpicture}
                    }
                    \vspace{-0.7cm}
                    \caption{\scriptsize Scenario (2) \\$ (l = 10, u = 100)$} 
                    \label{fig:e_subfig2}
                \end{subfigure}
                \&
                \begin{subfigure}{0.35\columnwidth}
                    \centering
                    \resizebox{\linewidth}{!}{
                        \begin{tikzpicture}
                            \node (img)  {\includegraphics[width=\textwidth]{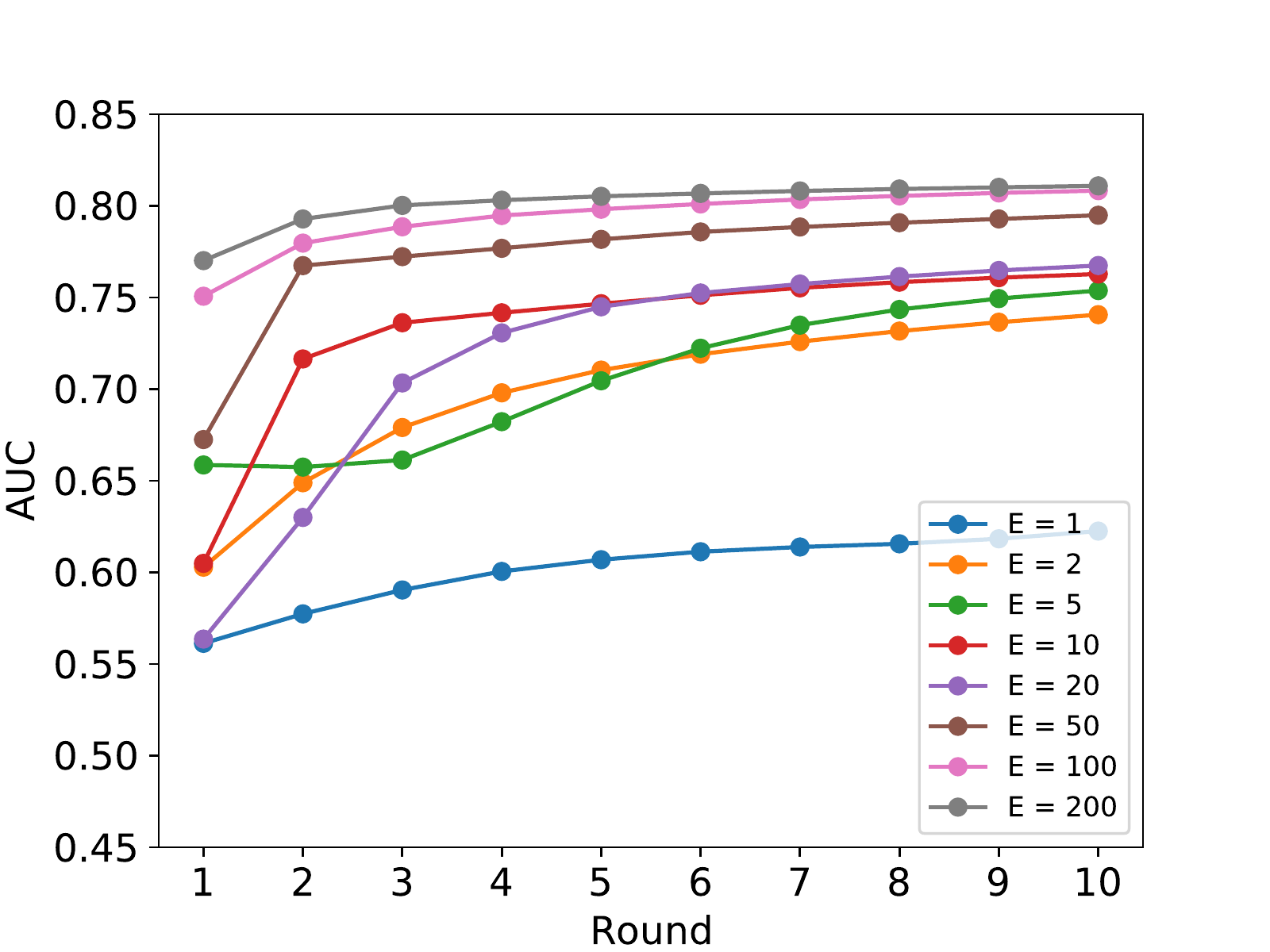}};
                        \end{tikzpicture}
                    }
                    \vspace{-0.7cm}
                    \caption{\scriptsize Scenario (3) \\$ (l = 10, u = 500)$} 
                    \label{fig:e_subfig3}
                \end{subfigure}
                \&
            \\
                \begin{subfigure}{0.35\columnwidth}
                    \centering
                    \resizebox{\linewidth}{!}{
                        \begin{tikzpicture}
                            \node (img)  {\includegraphics[width=\textwidth]{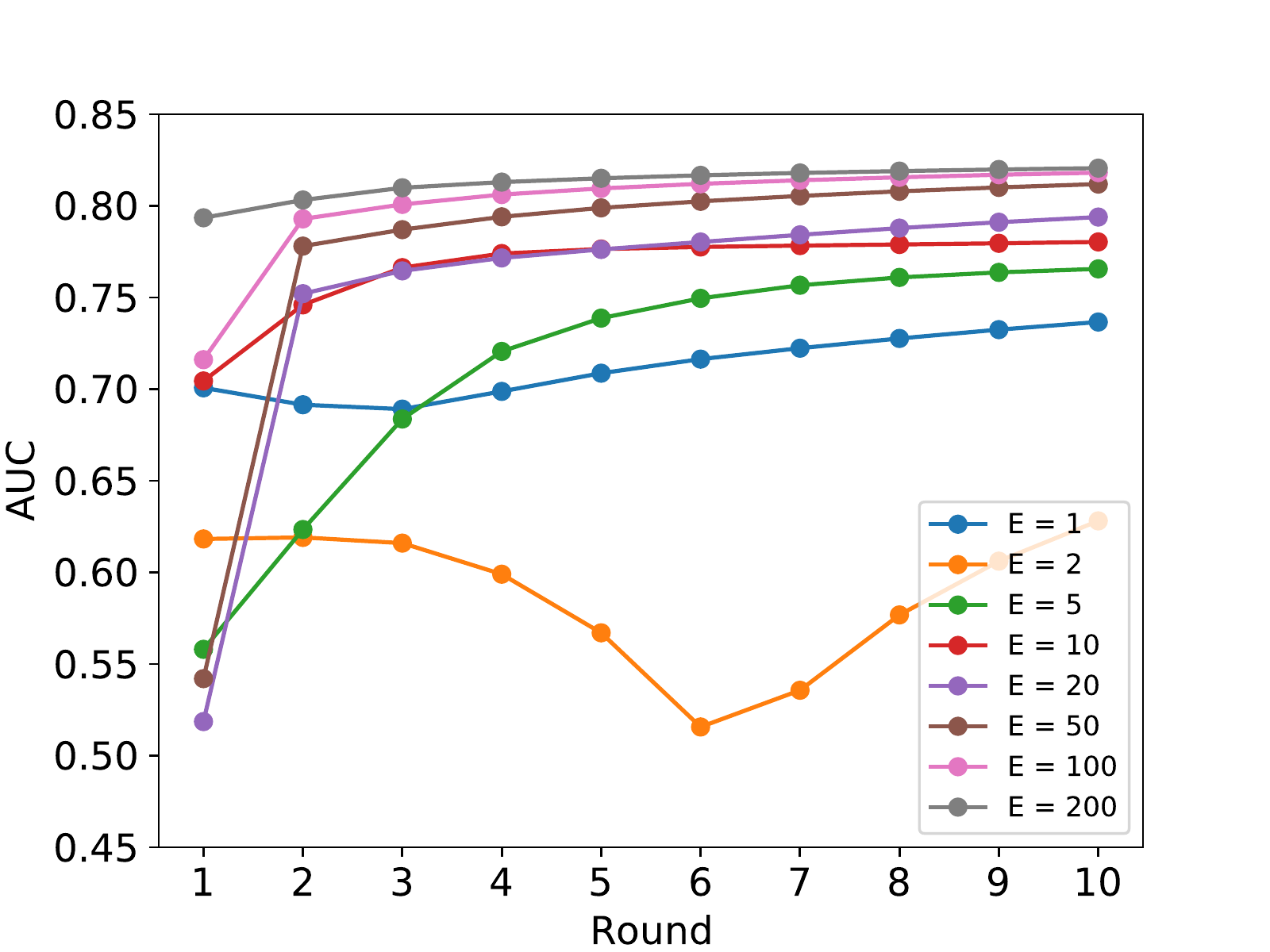}};
                        \end{tikzpicture}
                    }
                    \vspace{-0.7cm}
                    \caption{\scriptsize Scenario (4) \\$ (l = 10, u = 1000)$} 
                    \label{fig:e_subfig4}
                \end{subfigure}
                \&
                \begin{subfigure}{0.35\columnwidth}
                    \centering
                    \resizebox{\linewidth}{!}{
                        \begin{tikzpicture}
                            \node (img)  {\includegraphics[width=\textwidth]{figs/Effect_of_E/predict_death_from_apache/10/5000/dense.pdf}};
                        \end{tikzpicture}
                    }
                    \vspace{-0.7cm}
                    \caption{\scriptsize Scenario (5) \\$ (l = 10, u = 5000)$} 
                    \label{fig:e_subfig5}
                \end{subfigure}
                \&
                \begin{subfigure}{0.35\columnwidth}
                    \centering
                    \resizebox{\linewidth}{!}{
                        \begin{tikzpicture}
                            \node (img)  {\includegraphics[width=\textwidth]{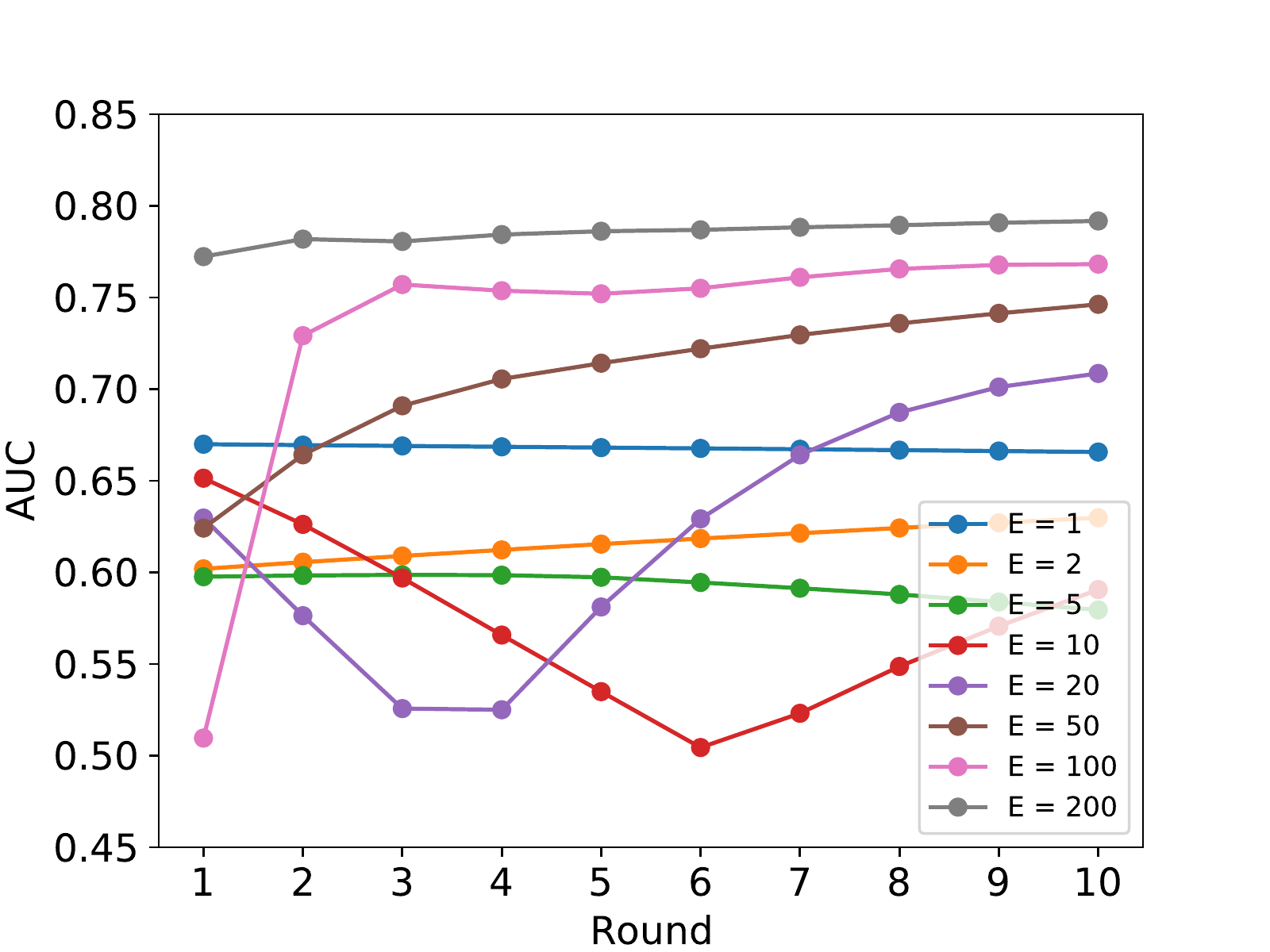}};
                        \end{tikzpicture}
                    }
                    \vspace{-0.7cm}
                    \caption{\scriptsize Scenario (6) \\$ (l = 50, u = 100)$} 
                    \label{fig:e_subfig6}
                \end{subfigure}
                \&
                \\
                \begin{subfigure}{0.35\columnwidth}
                    \centering
                    \resizebox{\linewidth}{!}{
                        \begin{tikzpicture}
                            \node (img)  {\includegraphics[width=\textwidth]{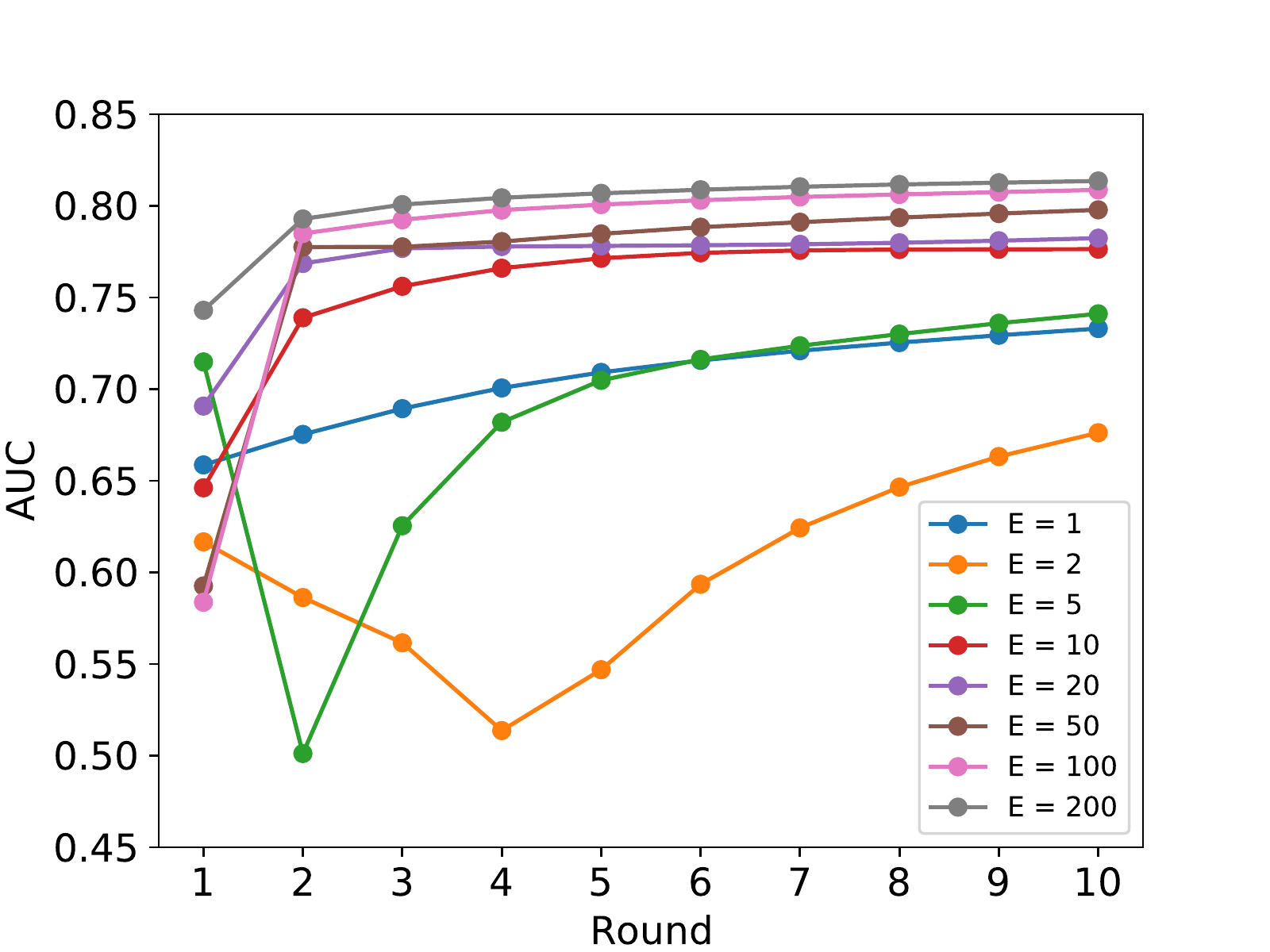}};
                        \end{tikzpicture}
                    }
                    \vspace{-0.7cm}
                    \caption{\scriptsize Scenario (7) \\$ (l = 50, u = 500)$} 
                    \label{fig:e_subfig7}
                \end{subfigure}
                \&
               \begin{subfigure}{0.35\columnwidth}
                    \centering
                    \resizebox{\linewidth}{!}{
                        \begin{tikzpicture}
                            \node (img)  {\includegraphics[width=\textwidth]{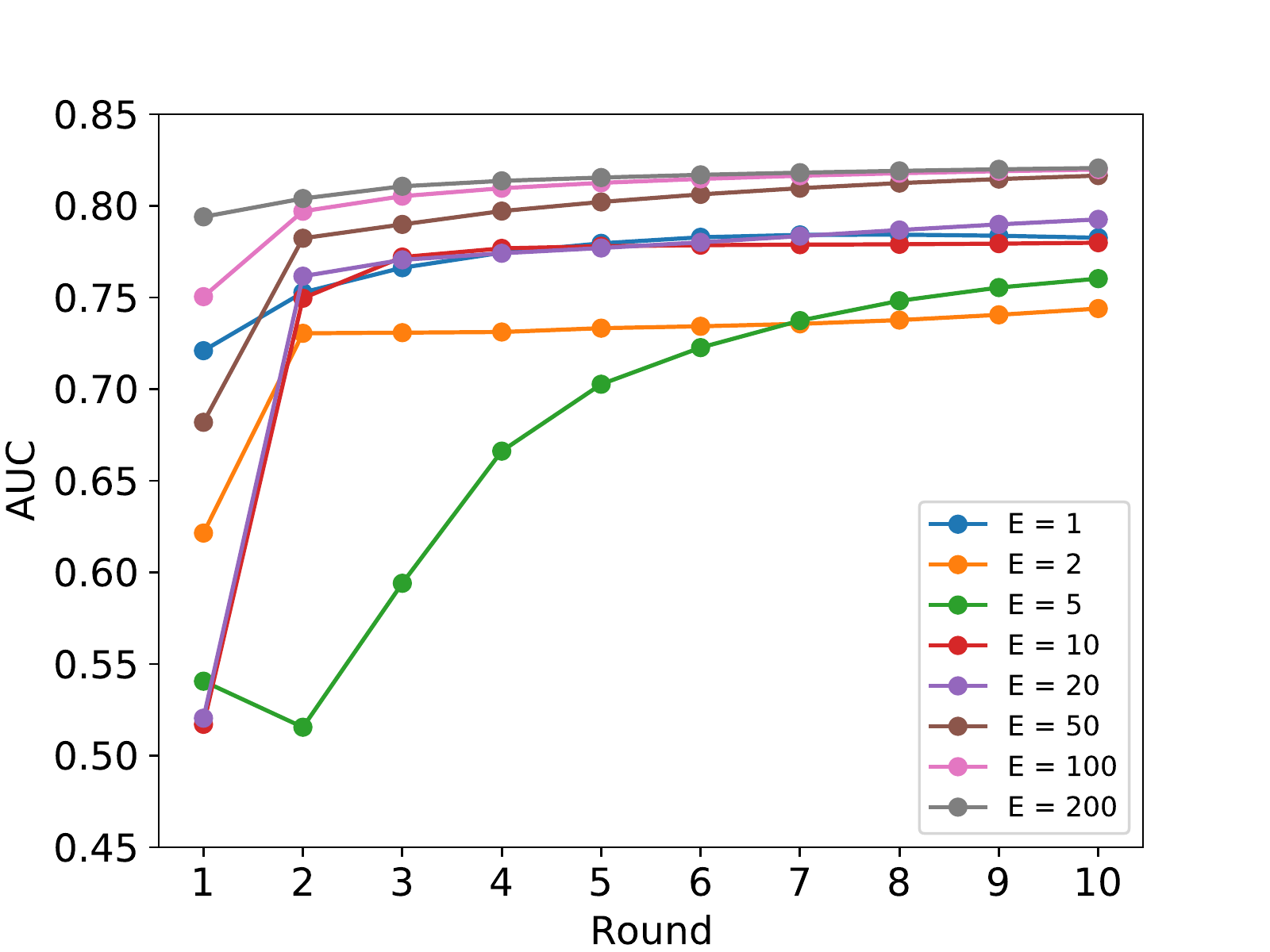}};
                        \end{tikzpicture}
                    }
                    \vspace{-0.7cm}
                    \caption{\scriptsize Scenario (8) \\$ (l = 50, u = 1000)$} 
                    \label{fig:e_subfig8}
                \end{subfigure}
                \&
                \begin{subfigure}{0.35\columnwidth}
                    \centering
                    \resizebox{\linewidth}{!}{
                        \begin{tikzpicture}
                            \node (img)  {\includegraphics[width=\textwidth]{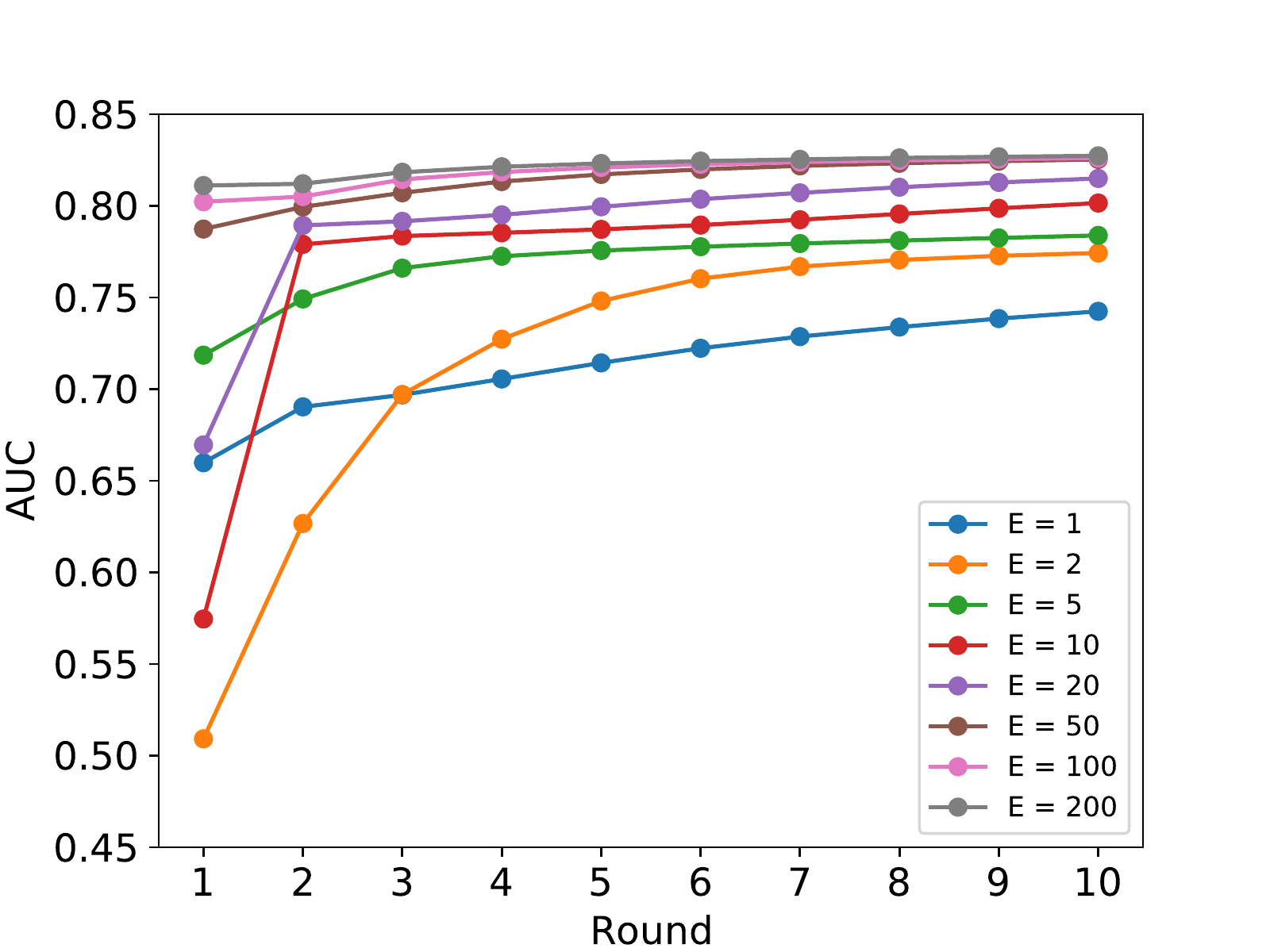}};
                        \end{tikzpicture}
                    }
                    \vspace{-0.7cm}
                    \caption{\scriptsize Scenario (9) \\$ (l = 50, u = 5000)$} 
                    \label{fig:e_subfig9}
                \end{subfigure}
                \&
                \\
                \begin{subfigure}{0.35\columnwidth}
                    \centering
                    \resizebox{\linewidth}{!}{
                        \begin{tikzpicture}
                            \node (img)  {\includegraphics[width=\textwidth]{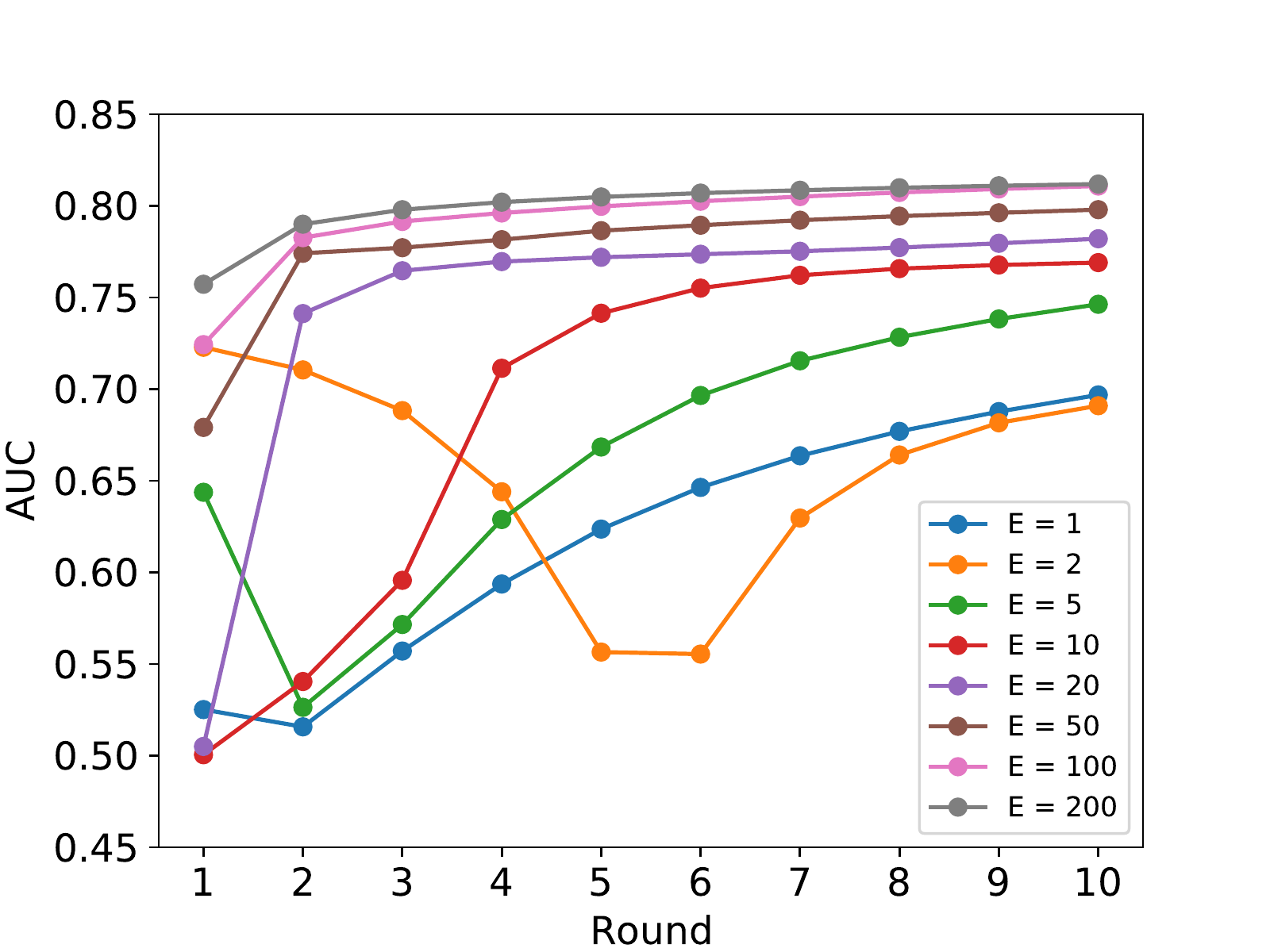}};
                        \end{tikzpicture}
                    }
                    \vspace{-0.7cm}
                    \caption{\scriptsize Scenario (10) \\$ (l = 100, u = 500)$} 
                    \label{fig:e_subfig10}
                \end{subfigure}
                \&
                \begin{subfigure}{0.35\columnwidth}
                    \centering
                    \resizebox{\linewidth}{!}{
                        \begin{tikzpicture}
                            \node (img)  {\includegraphics[width=\textwidth]{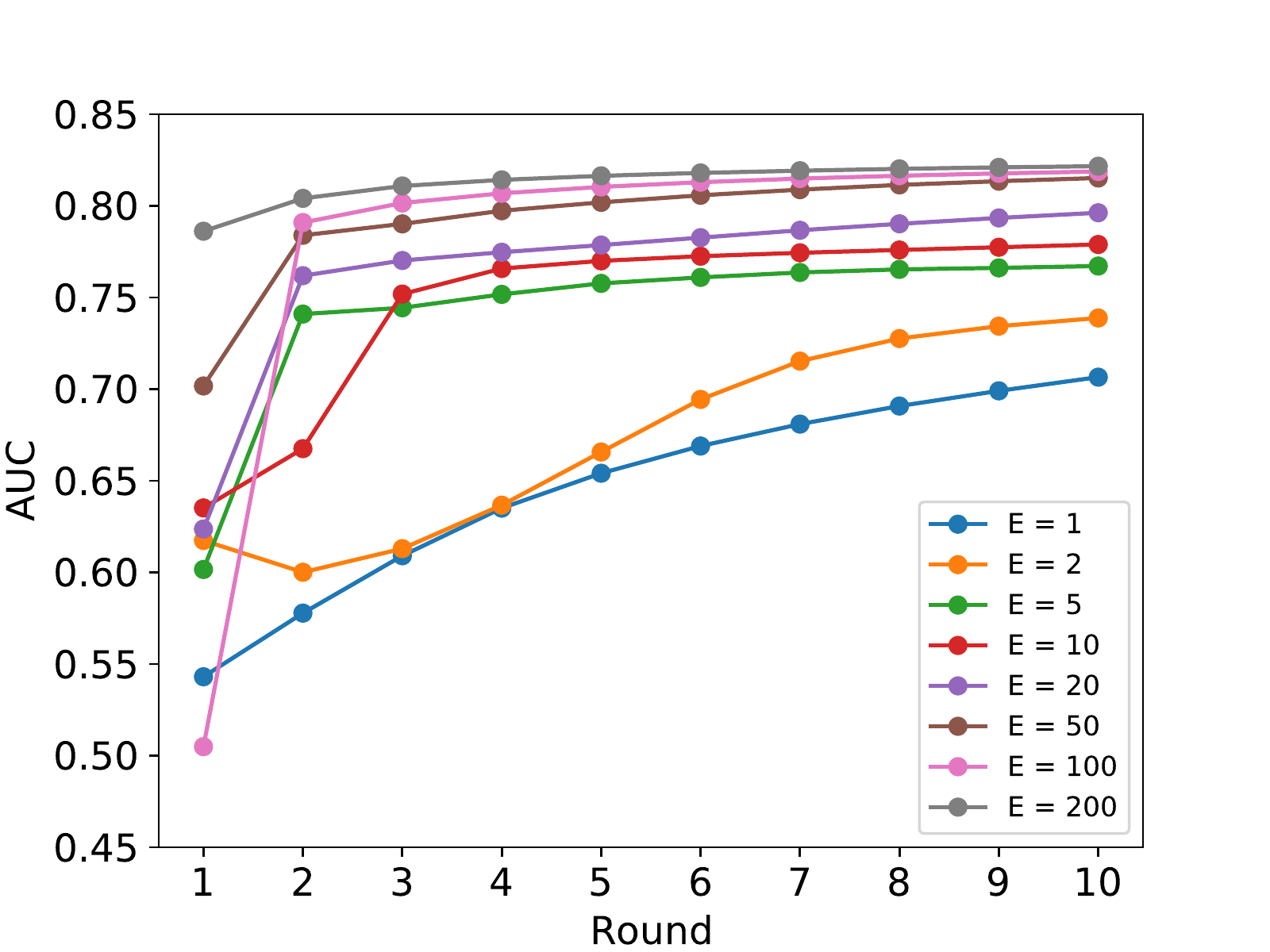}};
                        \end{tikzpicture}
                    }
                    \vspace{-0.7cm}
                    \caption{\scriptsize Scenario (11) \\$ (l = 100, u = 1000)$} 
                    \label{fig:e_subfig11}
                \end{subfigure}
                \&
                \begin{subfigure}{0.35\columnwidth}
                    \centering
                    \resizebox{\linewidth}{!}{
                        \begin{tikzpicture}
                            \node (img)  {\includegraphics[width=\textwidth]{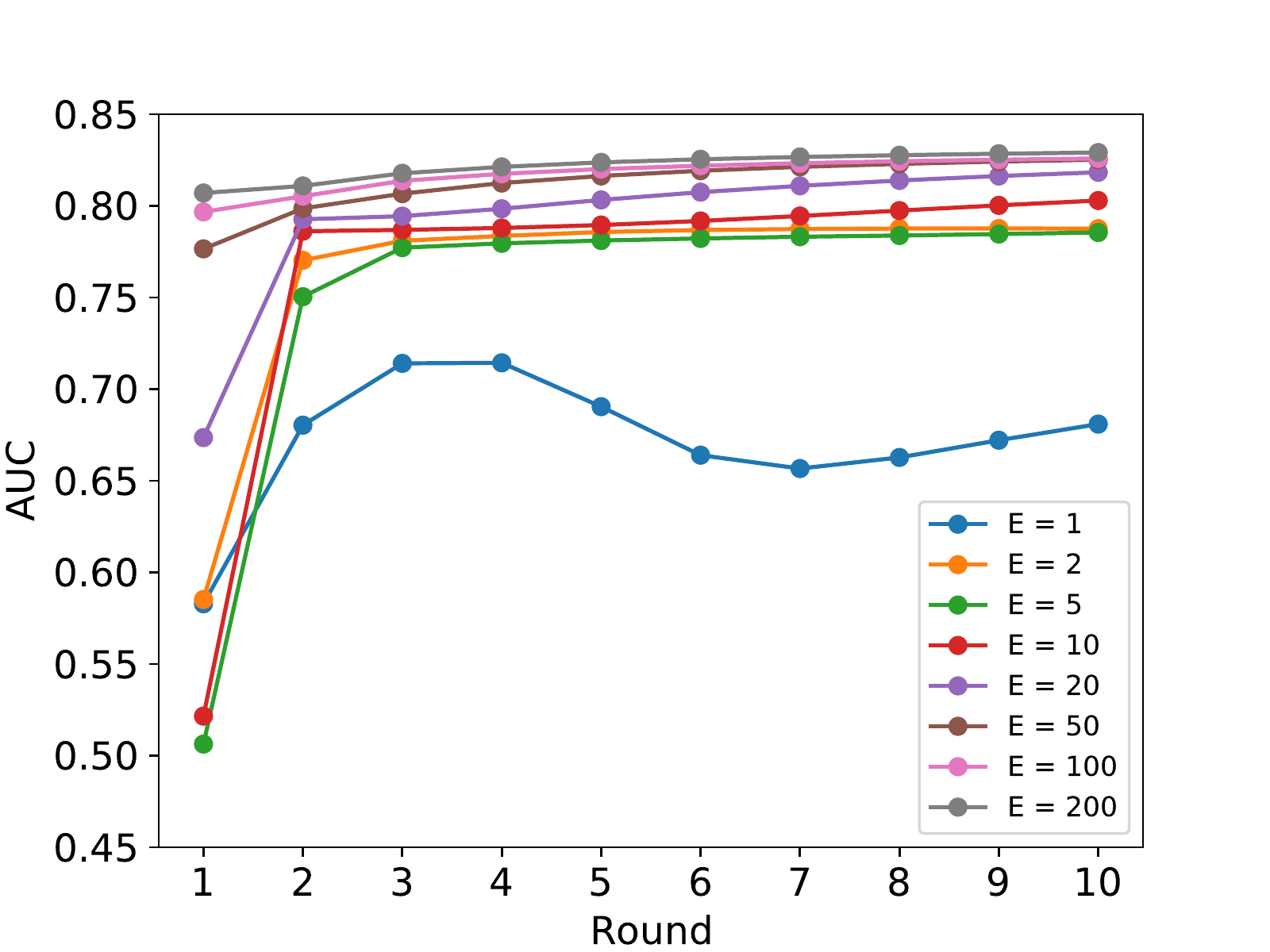}};
                        \end{tikzpicture}
                    }
                    \vspace{-0.7cm}
                    \caption{\scriptsize Scenario (12) \\$ (l = 100, u = 5000)$} 
                    \label{fig:e_subfig12}
                \end{subfigure}
                \&
                \\
                \begin{subfigure}{0.35\columnwidth}
                    \centering
                    \resizebox{\linewidth}{!}{
                        \begin{tikzpicture}
                            \node (img)  {\includegraphics[width=\textwidth]{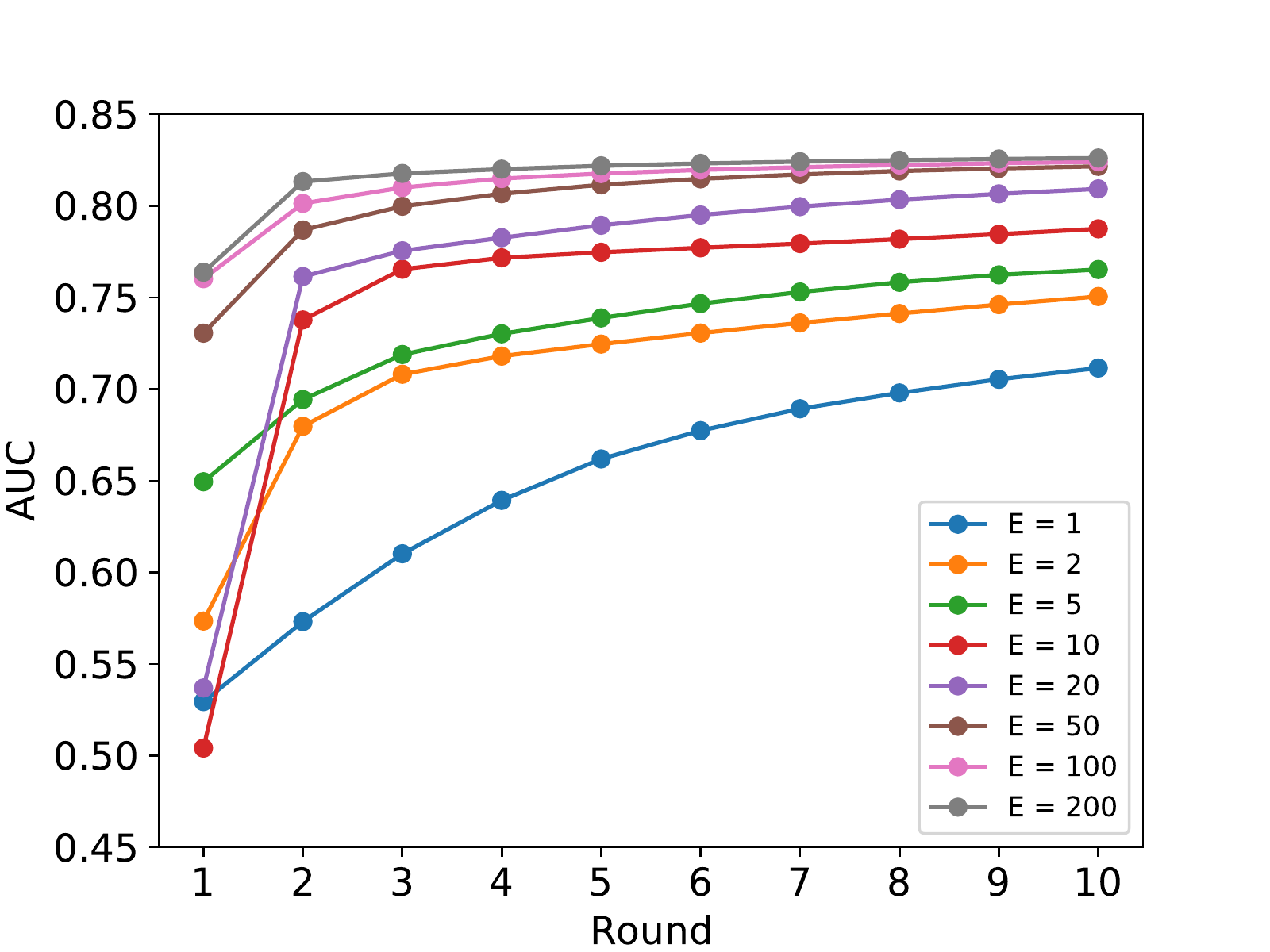}};
                        \end{tikzpicture}
                    }
                    \vspace{-0.7cm}
                    \caption{\scriptsize Scenario (13) \\$ (l = 500, u = 1000)$} 
                    \label{fig:e_subfig13}
                \end{subfigure}
                \&
                \begin{subfigure}{0.35\columnwidth}
                    \centering
                    \resizebox{\linewidth}{!}{
                        \begin{tikzpicture}
                            \node (img)  {\includegraphics[width=\textwidth]{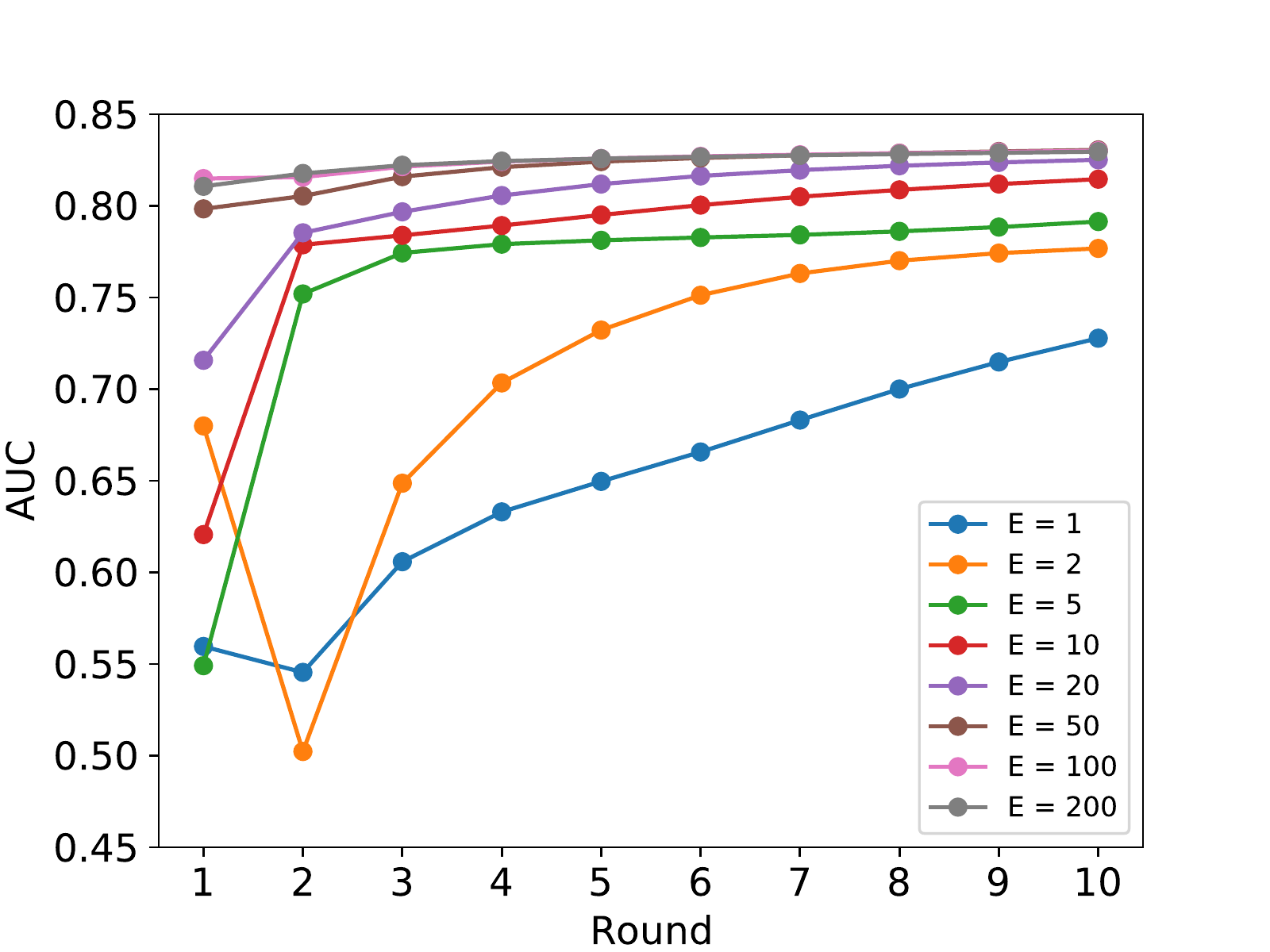}};
                        \end{tikzpicture}
                    }
                    \vspace{-0.7cm}
                    \caption{\scriptsize Scenario (14) \\$ (l = 500, u = 5000)$} 
                    \label{fig:e_subfig14}
                \end{subfigure}
                \&
                \begin{subfigure}{0.35\columnwidth}
                    \centering
                    \resizebox{\linewidth}{!}{
                        \begin{tikzpicture}
                            \node (img)  {\includegraphics[width=\textwidth]{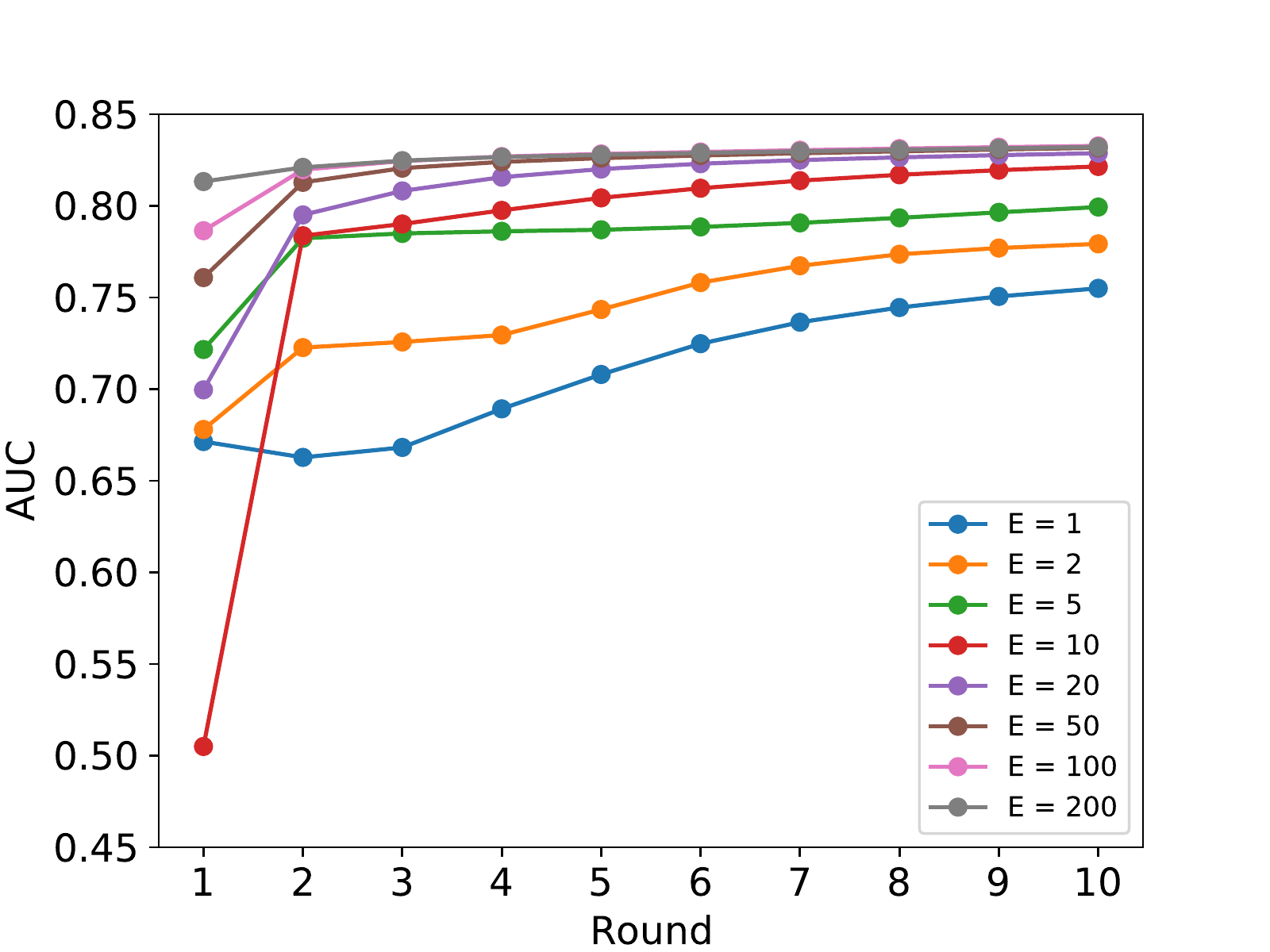}};
                        \end{tikzpicture}
                    }
                    \vspace{-0.7cm}
                    \caption{\scriptsize Scenario (15) \\$ (l = 1000, u = 5000)$} 
                    \label{fig:e_subfig15}
                \end{subfigure}
                \&
                \\
            \\
            };
        \end{tikzpicture}}
        \vspace{-0.5mm}
        \caption{Effect of E on Federated Learning (Scenarios 1 to 15).}
        \vspace{-5mm}
        \label{fig:e}
    \end{figure*}

 \begin{figure*}
    \vspace{-1.5cm}
        \centering
        \makebox[0.7\paperwidth]{%
            \begin{tikzpicture}[ampersand replacement=\&]
            \matrix (fig) [matrix of nodes, row sep=-1.1em, column sep=-3em]{ 
                \begin{subfigure}{0.35\columnwidth}
                    \centering
                    \resizebox{\linewidth}{!}{
                        \begin{tikzpicture}
                            \node (img)  {\includegraphics[width=\textwidth]{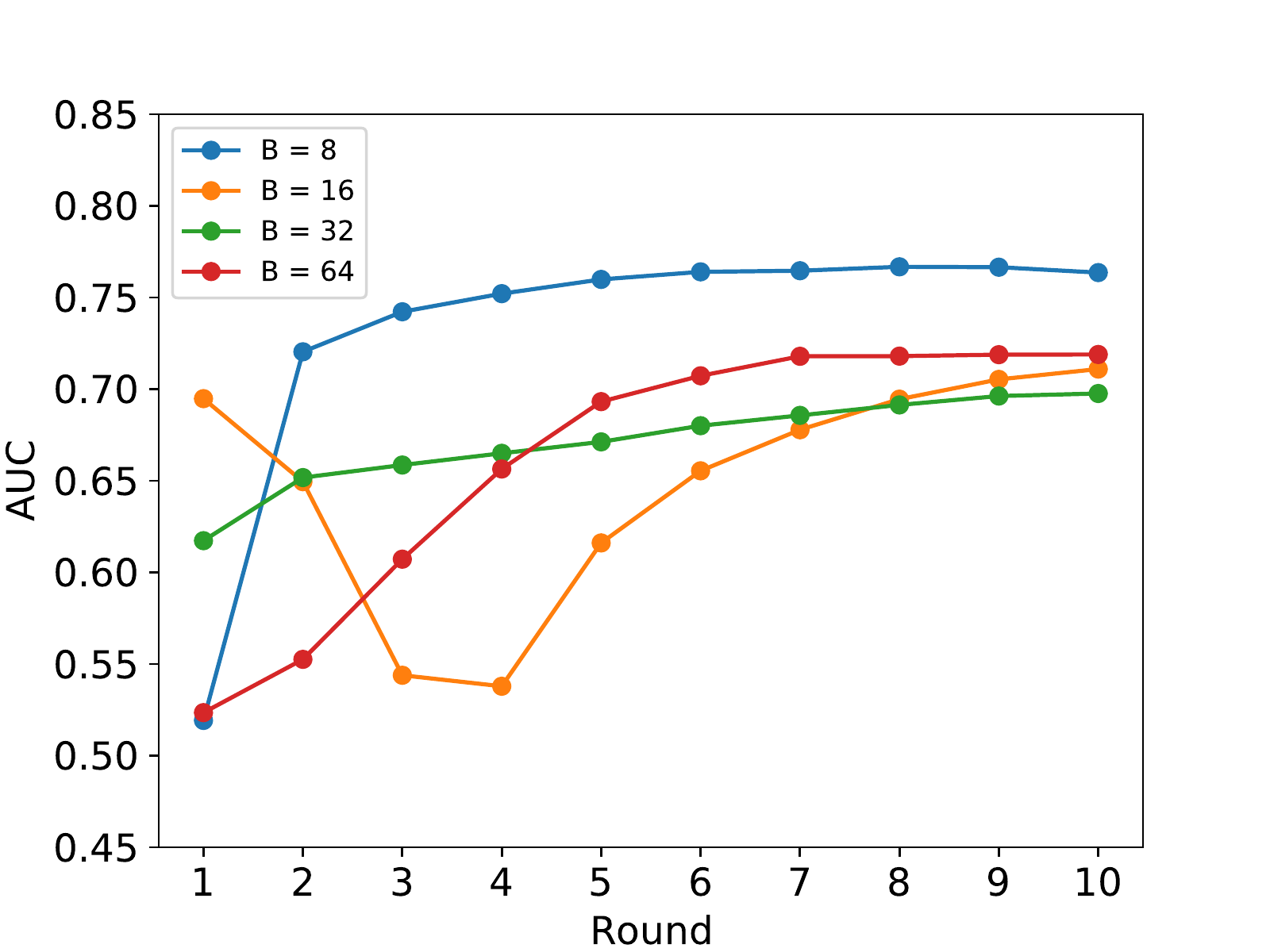}};
                        \end{tikzpicture}
                    }
                    \vspace{-0.7cm}
                    \caption{\scriptsize Scenario (16) \\ $(l = 5, u = 50)$} 
                    \label{fig:bc1_subfig1}
                \end{subfigure}
                \&
                \begin{subfigure}{0.35\columnwidth}
                    \centering
                    \resizebox{\linewidth}{!}{
                        \begin{tikzpicture}
                            \node (img)  {\includegraphics[width=\textwidth]{figs/Effect_of_B_C/predict_death_from_apache/5/500/0.2/dense.pdf}};
                        \end{tikzpicture}
                    }
                    \vspace{-0.7cm}
                    \caption{\scriptsize Scenario (17) \\$ (l = 5, u = 500)$} 
                    \label{fig:bc1_subfig2}
                \end{subfigure}
                \&
                \begin{subfigure}{0.35\columnwidth}
                    \centering
                    \resizebox{\linewidth}{!}{
                        \begin{tikzpicture}
                            \node (img)  {\includegraphics[width=\textwidth]{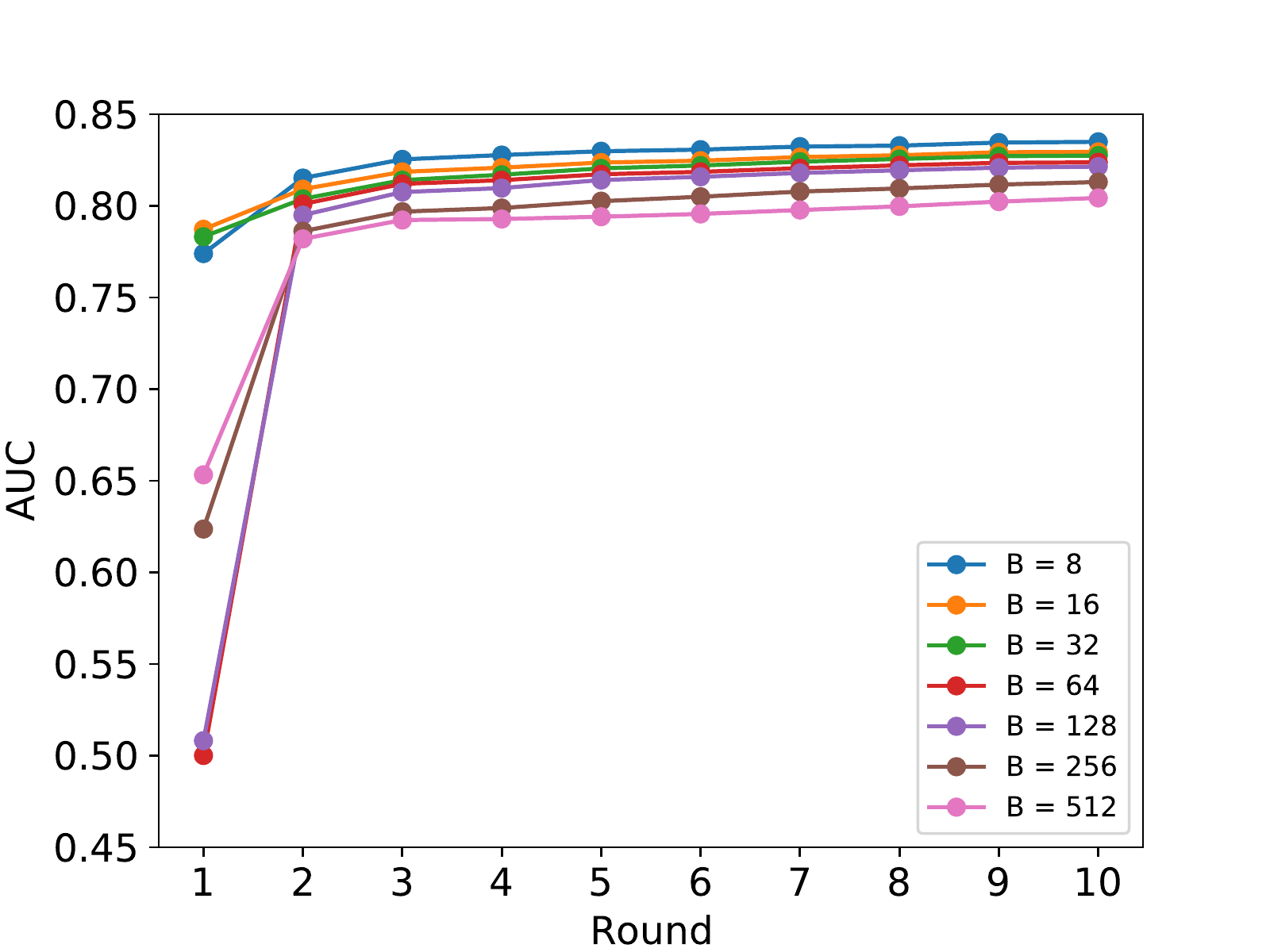}};
                        \end{tikzpicture}
                    }
                    \vspace{-0.7cm}
                    \caption{\scriptsize Scenario (18) \\$ (l = 5, u = 5000)$} 
                    \label{fig:bc1_subfig3}
                \end{subfigure}
                \&
            \\
                \begin{subfigure}{0.35\columnwidth}
                    \centering
                    \resizebox{\linewidth}{!}{
                        \begin{tikzpicture}
                            \node (img)  {\includegraphics[width=\textwidth]{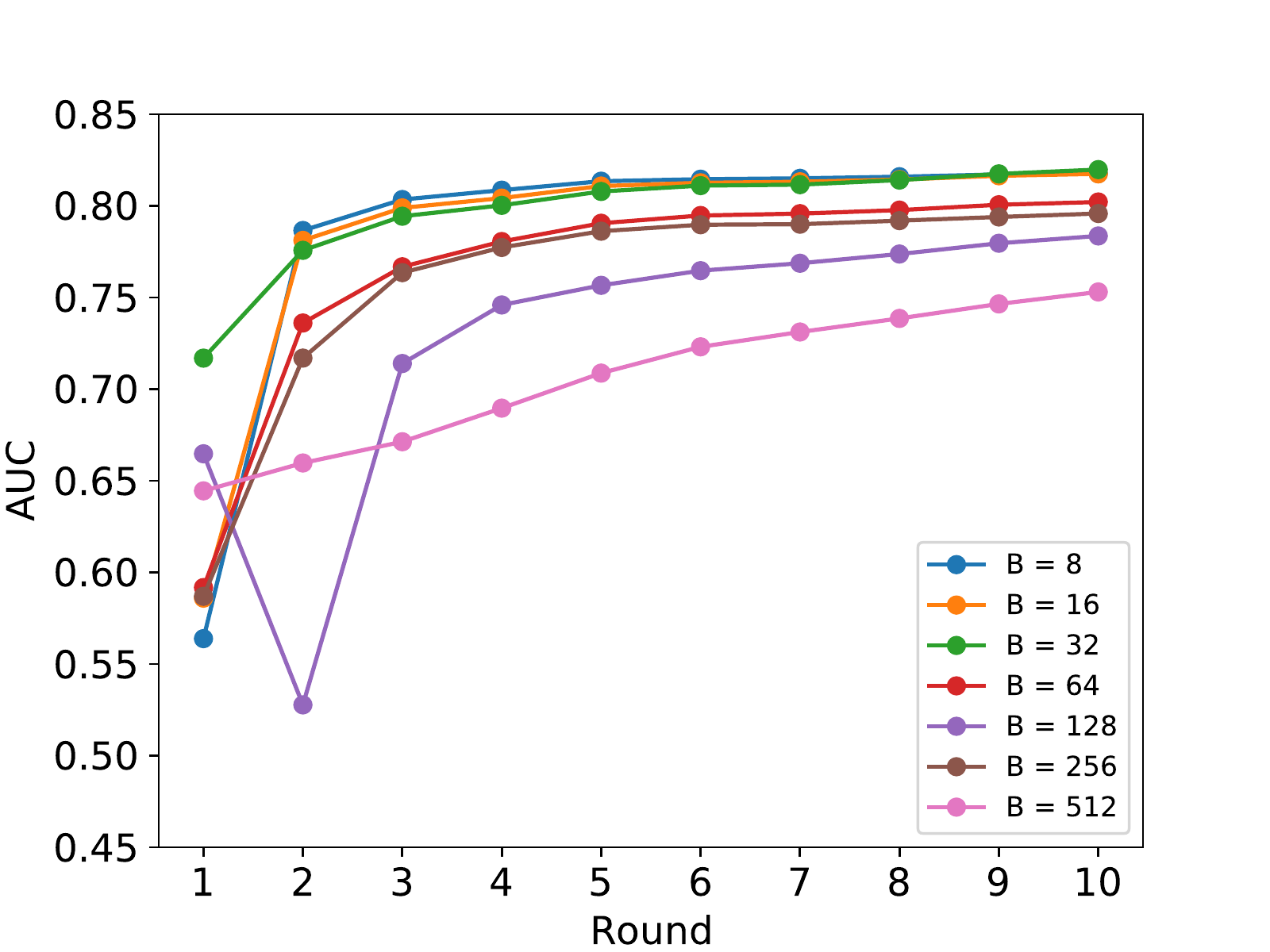}};
                        \end{tikzpicture}
                    }
                    \vspace{-0.7cm}
                    \caption{\scriptsize Scenario (7) \\$ (l = 50, u = 500)$} 
                    \label{fig:bc1_subfig4}
                \end{subfigure}
                \&
                \begin{subfigure}{0.35\columnwidth}
                    \centering
                    \resizebox{\linewidth}{!}{
                        \begin{tikzpicture}
                            \node (img)  {\includegraphics[width=\textwidth]{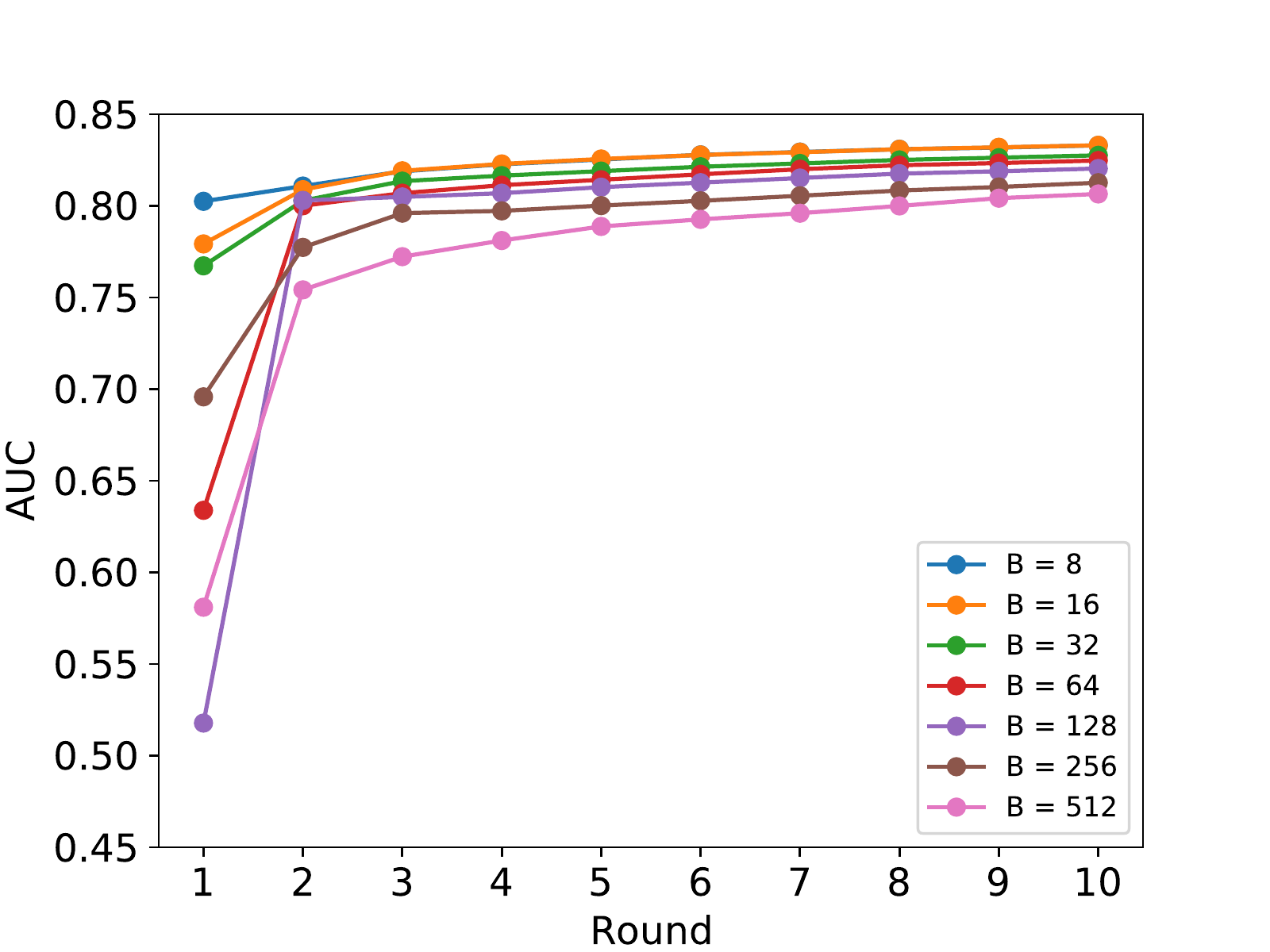}};
                        \end{tikzpicture}
                    }
                    \vspace{-0.7cm}
                    \caption{\scriptsize Scenario (9) \\$ (l = 50, u = 5000)$} 
                    \label{fig:bc1_subfig5}
                \end{subfigure}
                \&
                \begin{subfigure}{0.35\columnwidth}
                    \centering
                    \resizebox{\linewidth}{!}{
                        \begin{tikzpicture}
                            \node (img)  {\includegraphics[width=\textwidth]{figs/Effect_of_B_C/predict_death_from_apache/500/5000/0.2/dense.pdf}};
                        \end{tikzpicture}
                    }
                    \vspace{-0.7cm}
                    \caption{\scriptsize Scenario (14) \\$ (l = 500, u = 5000)$} 
                    \label{fig:bc1_subfig6}
                \end{subfigure}
                \&
                \\
                \begin{subfigure}{0.35\columnwidth}
                    \centering
                    \resizebox{\linewidth}{!}{
                        \begin{tikzpicture}
                            \node (img)  {\includegraphics[width=\textwidth]{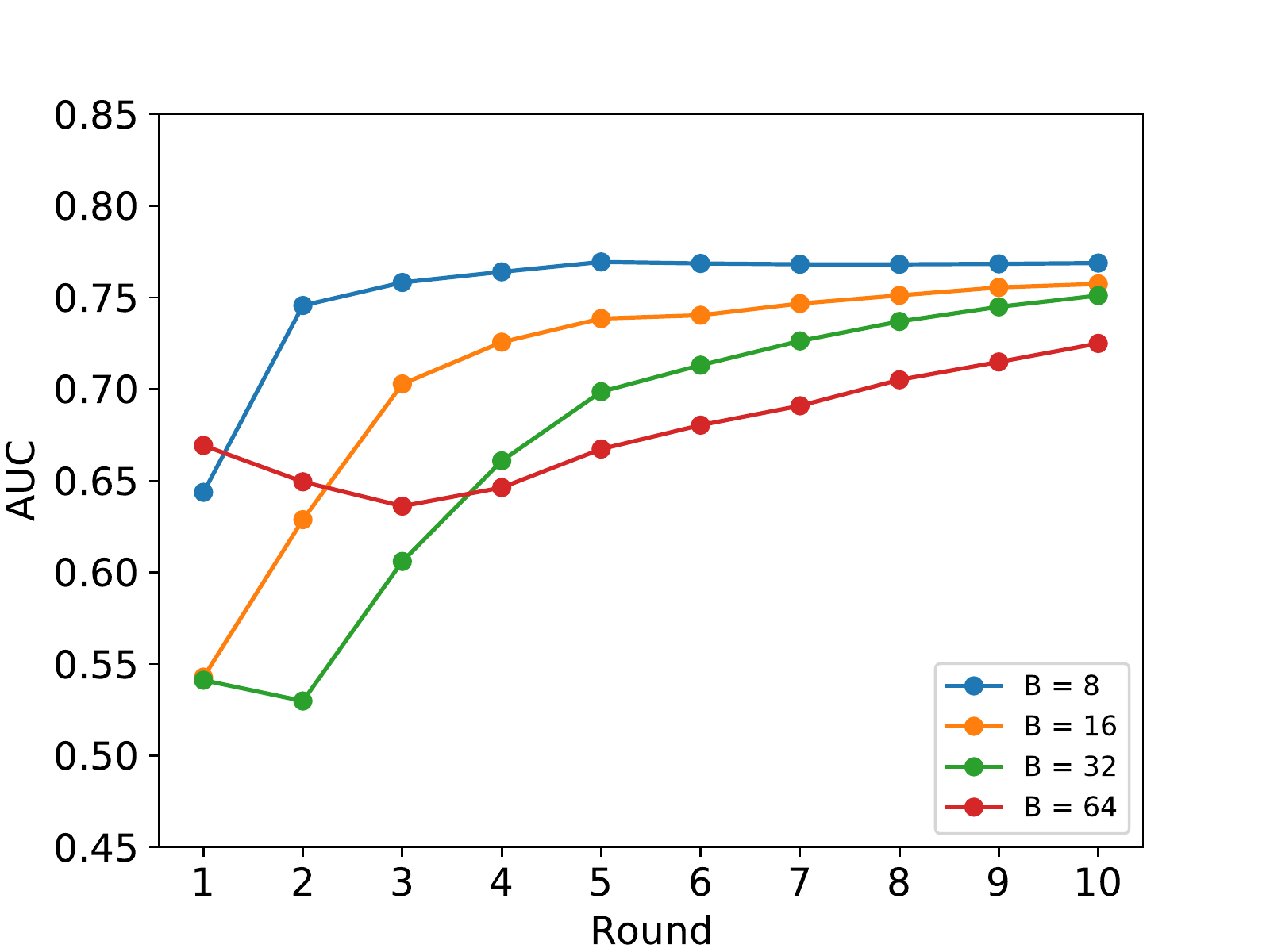}};
                        \end{tikzpicture}
                    }
                    \vspace{-0.7cm}
                    \caption{\scriptsize Scenario (16) \\$ (l = 5, u = 50)$} 
                    \label{fig:bc1_subfig7}
                \end{subfigure}
                \&
               \begin{subfigure}{0.35\columnwidth}
                    \centering
                    \resizebox{\linewidth}{!}{
                        \begin{tikzpicture}
                            \node (img)  {\includegraphics[width=\textwidth]{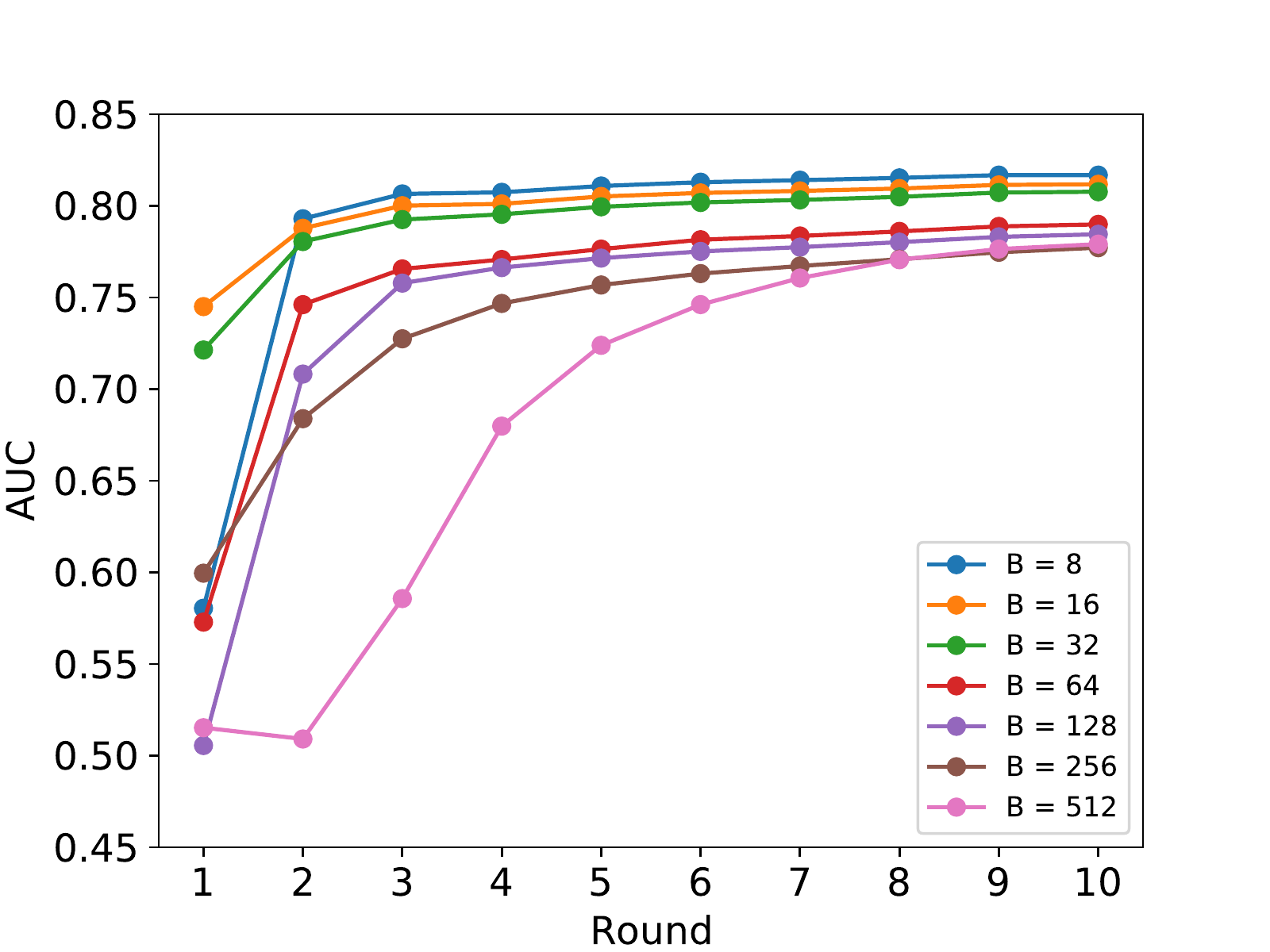}};
                        \end{tikzpicture}
                    }
                    \vspace{-0.7cm}
                    \caption{\scriptsize Scenario (17) \\$ (l = 5, u = 500)$} 
                    \label{fig:bc1_subfig8}
                \end{subfigure}
                \&
                \begin{subfigure}{0.35\columnwidth}
                    \centering
                    \resizebox{\linewidth}{!}{
                        \begin{tikzpicture}
                            \node (img)  {\includegraphics[width=\textwidth]{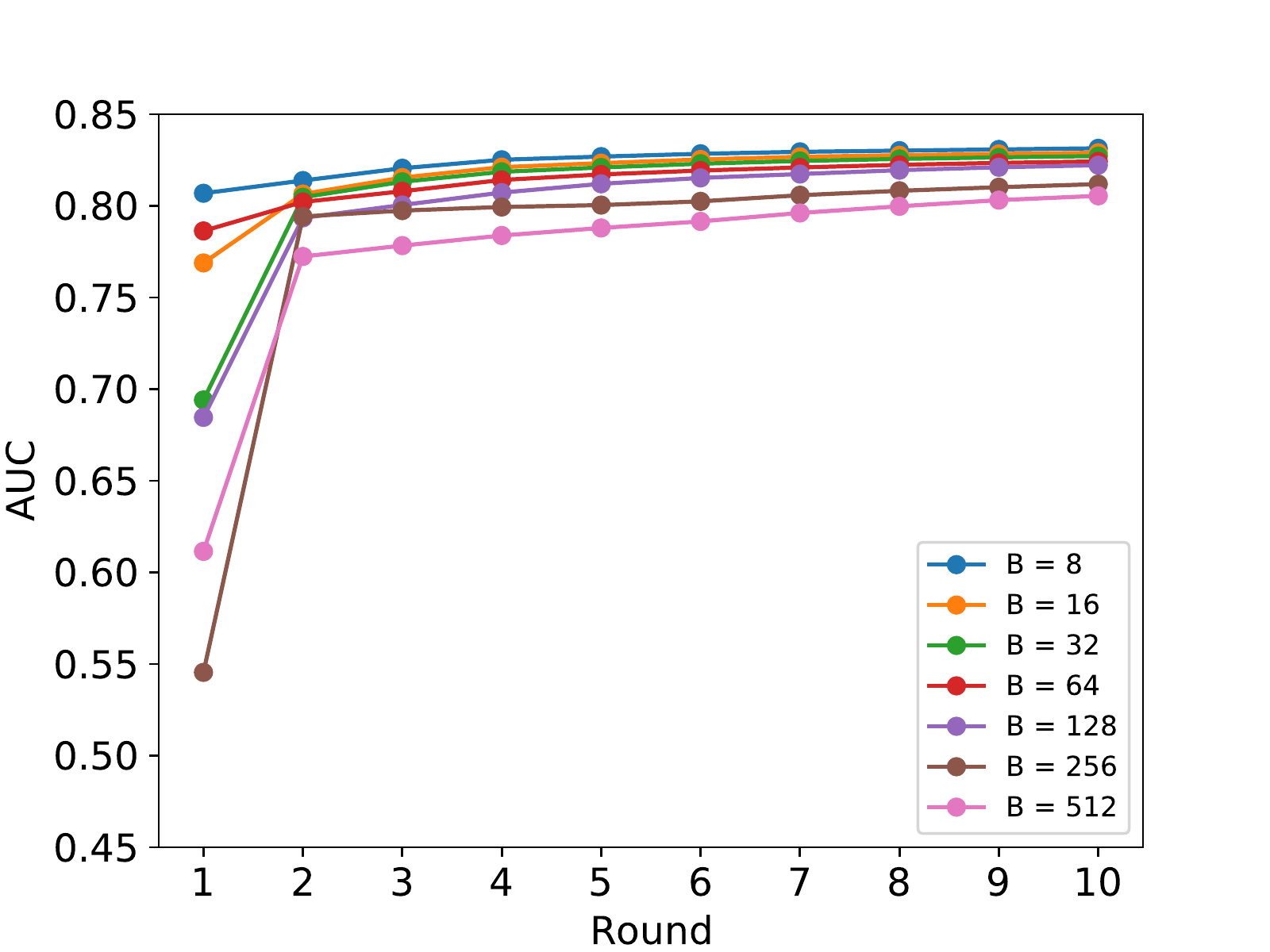}};
                        \end{tikzpicture}
                    }
                    \vspace{-0.7cm}
                    \caption{\scriptsize Scenario (18) \\$ (l = 5, u = 5000)$} 
                    \label{fig:bc1_subfig9}
                \end{subfigure}
                \&
                \\
                \begin{subfigure}{0.35\columnwidth}
                    \centering
                    \resizebox{\linewidth}{!}{
                        \begin{tikzpicture}
                            \node (img)  {\includegraphics[width=\textwidth]{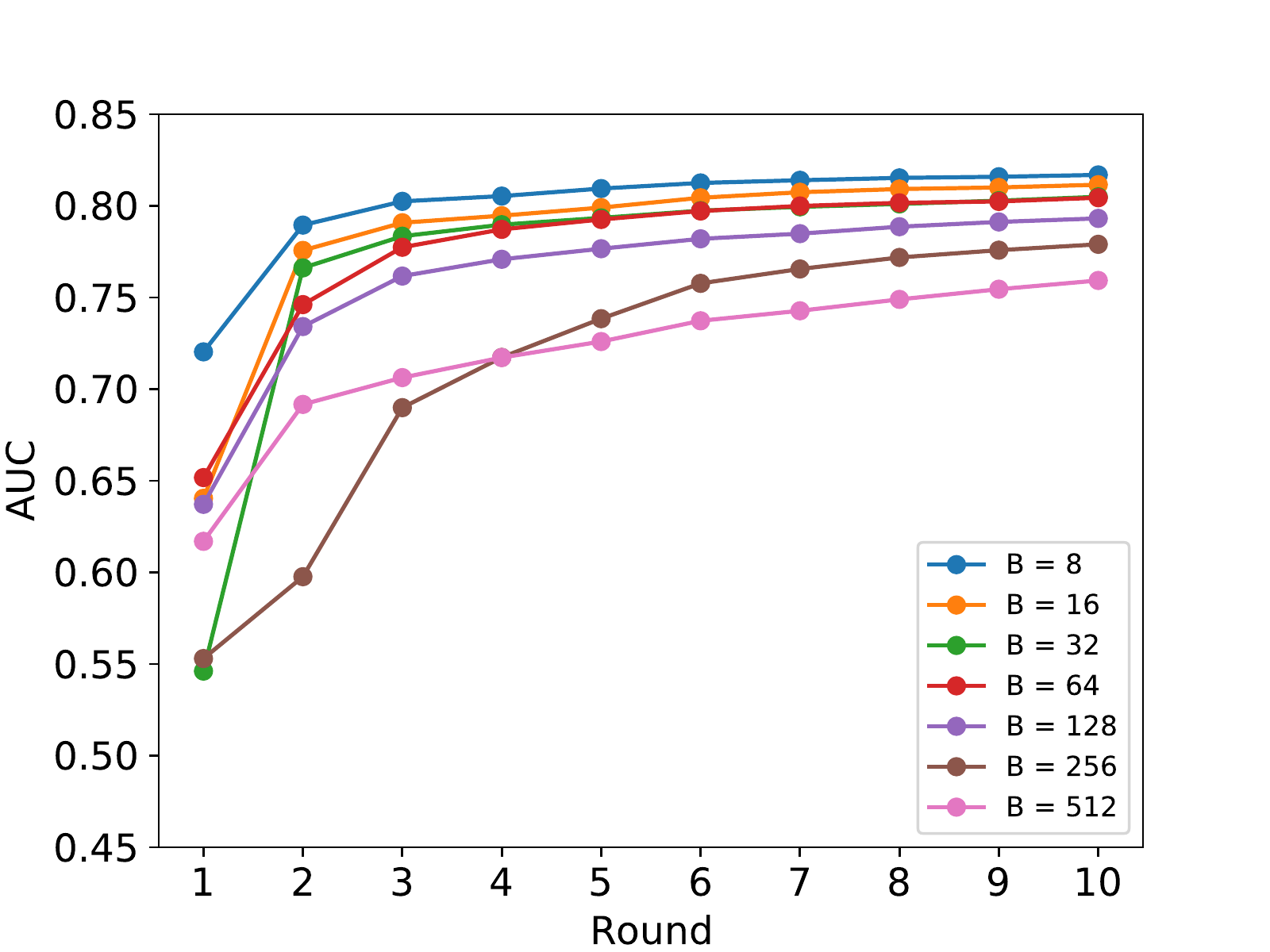}};
                        \end{tikzpicture}
                    }
                    \vspace{-0.7cm}
                    \caption{\scriptsize Scenario (7) \\$ (l = 50, u = 500)$} 
                    \label{fig:bc1_subfig10}
                \end{subfigure}
                \&
                \begin{subfigure}{0.35\columnwidth}
                    \centering
                    \resizebox{\linewidth}{!}{
                        \begin{tikzpicture}
                            \node (img)  {\includegraphics[width=\textwidth]{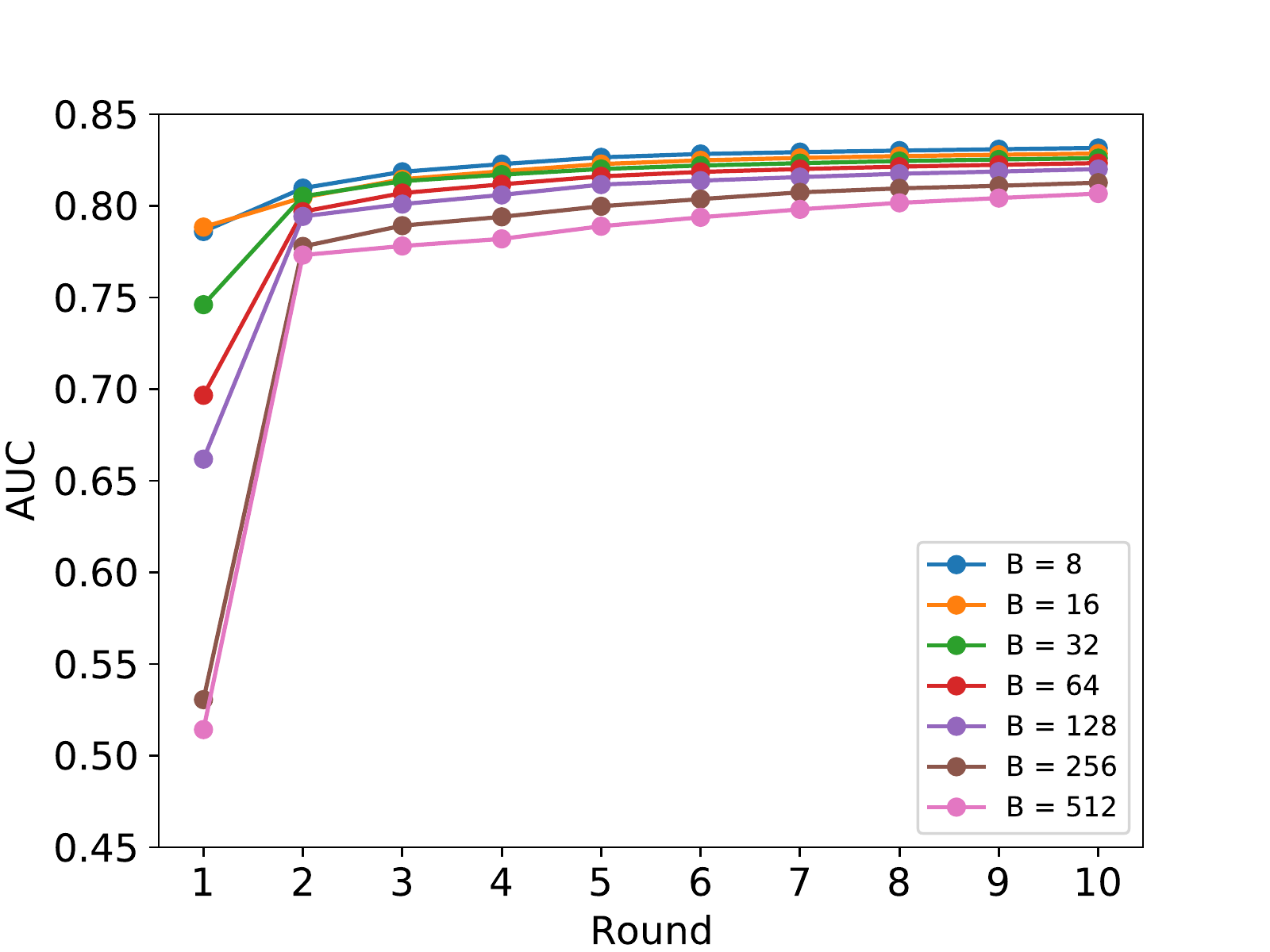}};
                        \end{tikzpicture}
                    }
                    \vspace{-0.7cm}
                    \caption{\scriptsize Scenario (9) \\$ (l = 50, u = 5000)$} 
                    \label{fig:bc1_subfig11}
                \end{subfigure}
                \&
                \begin{subfigure}{0.35\columnwidth}
                    \centering
                    \resizebox{\linewidth}{!}{
                        \begin{tikzpicture}
                            \node (img)  {\includegraphics[width=\textwidth]{figs/Effect_of_B_C/predict_death_from_apache/500/5000/0.4/dense.pdf}};
                        \end{tikzpicture}
                    }
                    \vspace{-0.7cm}
                    \caption{\scriptsize Scenario (14) \\$ (l = 500, u = 5000)$} 
                    \label{fig:bc1_subfig12}
                \end{subfigure}
                \&
                \\
                \begin{subfigure}{0.35\columnwidth}
                    \centering
                    \resizebox{\linewidth}{!}{
                        \begin{tikzpicture}
                            \node (img)  {\includegraphics[width=\textwidth]{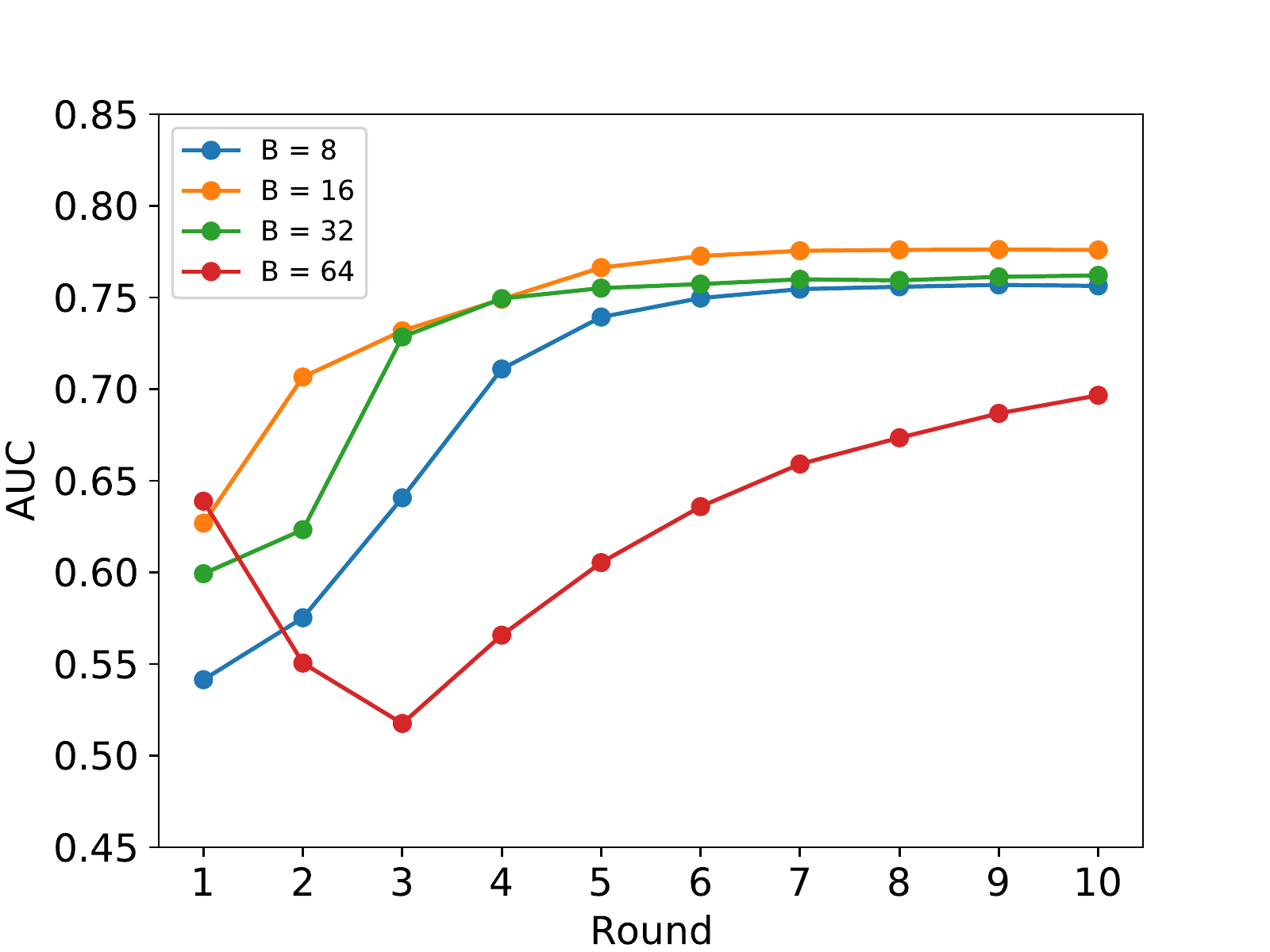}};
                        \end{tikzpicture}
                    }
                    \vspace{-0.7cm}
                    \caption{\scriptsize Scenario (16) \\$ (l = 5, u = 50)$} 
                    \label{fig:bc1_subfig13}
                \end{subfigure}
                \&
                \begin{subfigure}{0.35\columnwidth}
                    \centering
                    \resizebox{\linewidth}{!}{
                        \begin{tikzpicture}
                            \node (img)  {\includegraphics[width=\textwidth]{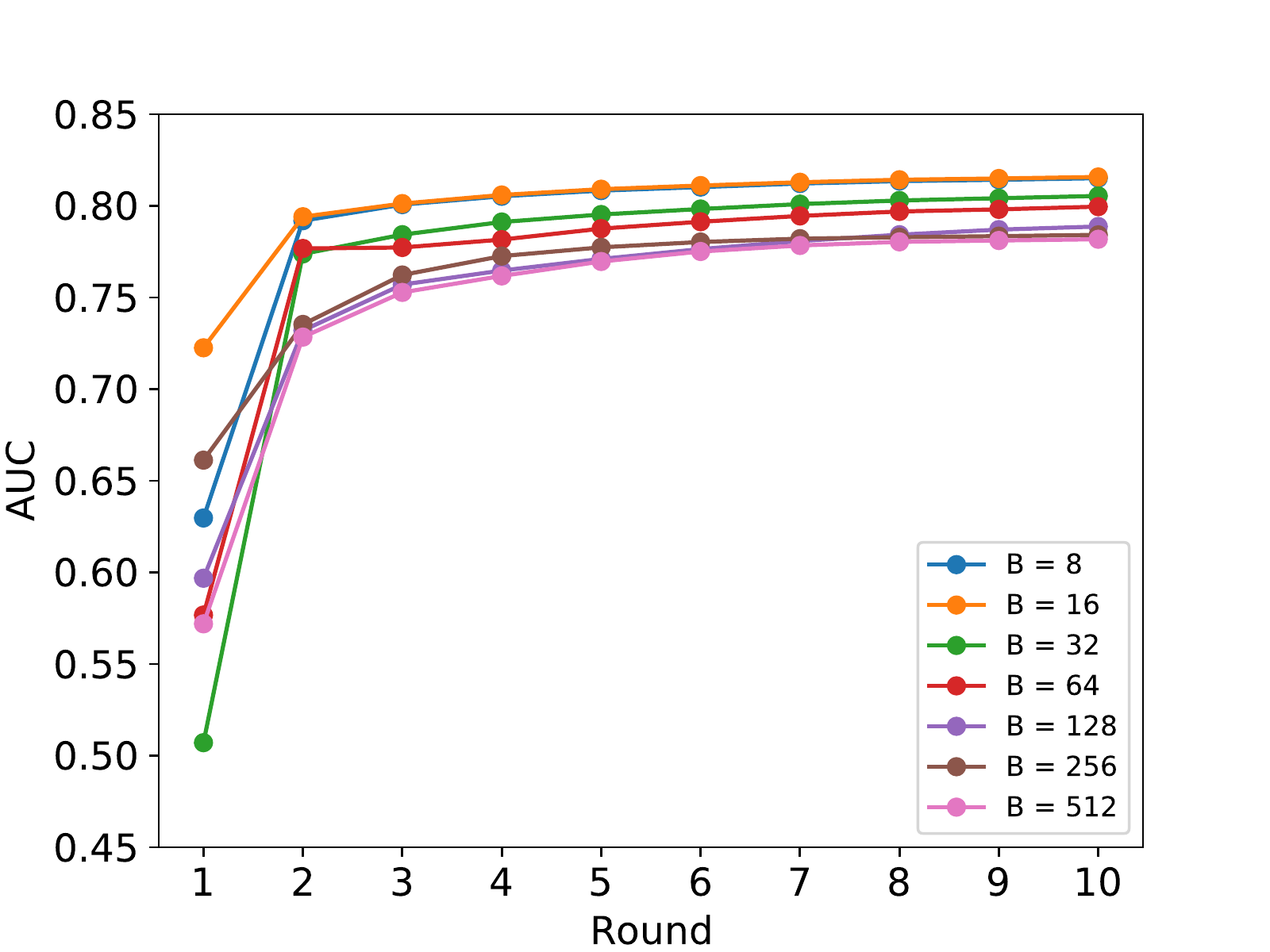}};
                        \end{tikzpicture}
                    }
                    \vspace{-0.7cm}
                    \caption{\scriptsize Scenario (17) \\$ (l = 5, u = 500)$} 
                    \label{fig:bc1_subfig14}
                \end{subfigure}
                \&
                \begin{subfigure}{0.35\columnwidth}
                    \centering
                    \resizebox{\linewidth}{!}{
                        \begin{tikzpicture}
                            \node (img)  {\includegraphics[width=\textwidth]{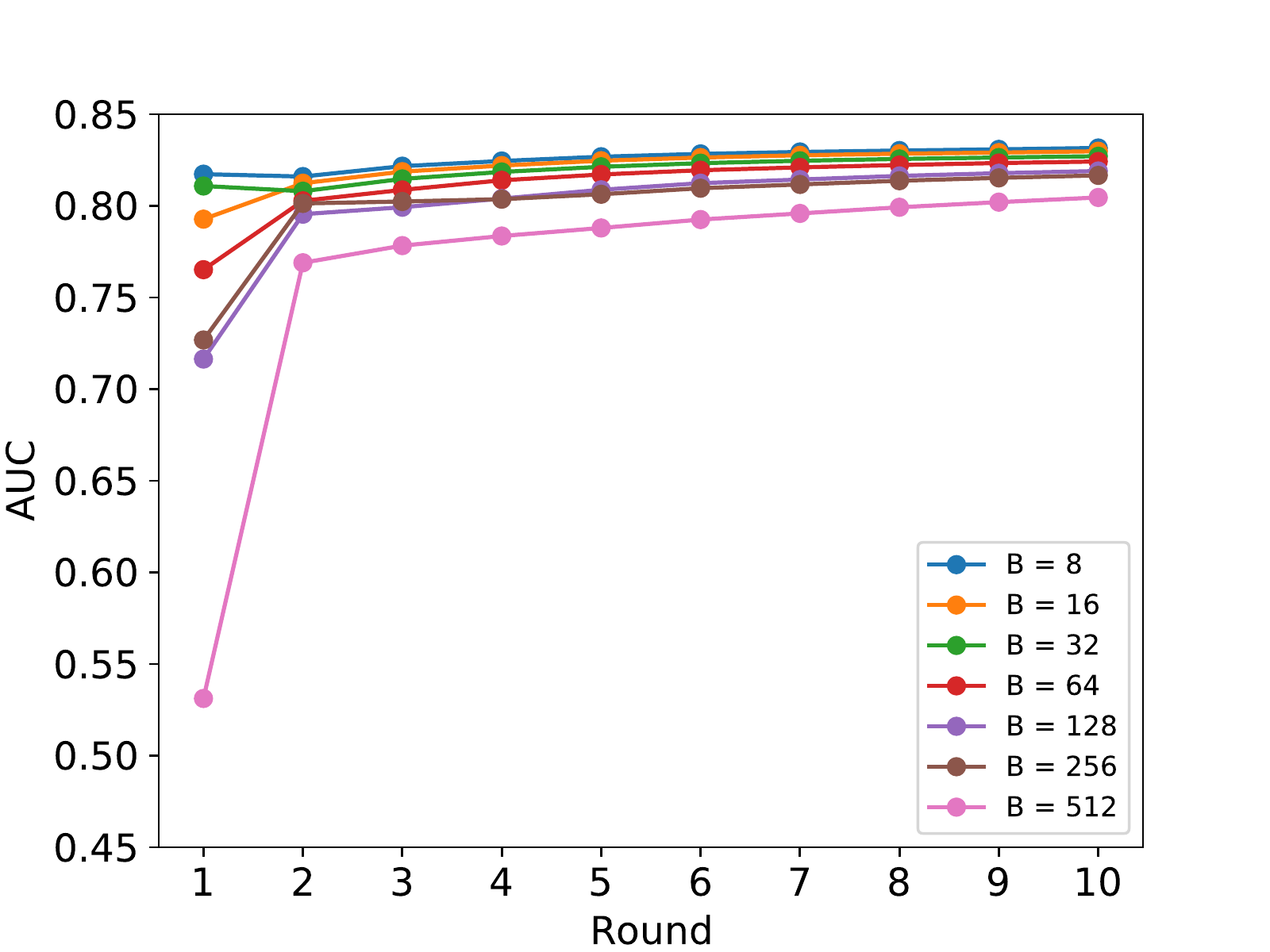}};
                        \end{tikzpicture}
                    }
                    \vspace{-0.7cm}
                    \caption{\scriptsize Scenario (18) \\$ (l = 5, u = 5000)$} 
                    \label{fig:bc1_subfig15}
                \end{subfigure}
                \&
                \\
            \\
            };
            \node [draw=none, rotate=90] at ([xshift=-1mm, yshift=2mm]fig-1-1.west) {\small $C=0.2$ };
            \node [draw=none, rotate=90] at ([xshift=-1mm, yshift=2mm]fig-2-1.west) {\small $C=0.2$ };
            \node [draw=none, rotate=90] at ([xshift=-1mm, yshift=2mm]fig-3-1.west) {\small $C=0.4$ };
            \node [draw=none, rotate=90] at ([xshift=-1mm, yshift=2mm]fig-4-1.west) {\small $C=0.4$ };
            \node [draw=none, rotate=90] at ([xshift=-1mm, yshift=2mm]fig-5-1.west) {\small $C=0.6 $ };
        \end{tikzpicture}}
        \vspace{-0.5mm}
        \caption{Effect of B \& C on Federated Learning (1st Part).}
        \vspace{-5mm}
        \label{fig:bc1}
    \end{figure*}

 \begin{figure*}
    \vspace{-1.5cm}
        \centering
        \makebox[0.6\paperwidth]{%
            \begin{tikzpicture}[ampersand replacement=\&]
            \matrix (fig) [matrix of nodes, row sep=-1.1em, column sep=-3em]{ 
                \begin{subfigure}{0.35\columnwidth}
                    \centering
                    \resizebox{\linewidth}{!}{
                        \begin{tikzpicture}
                            \node (img)  {\includegraphics[width=\textwidth]{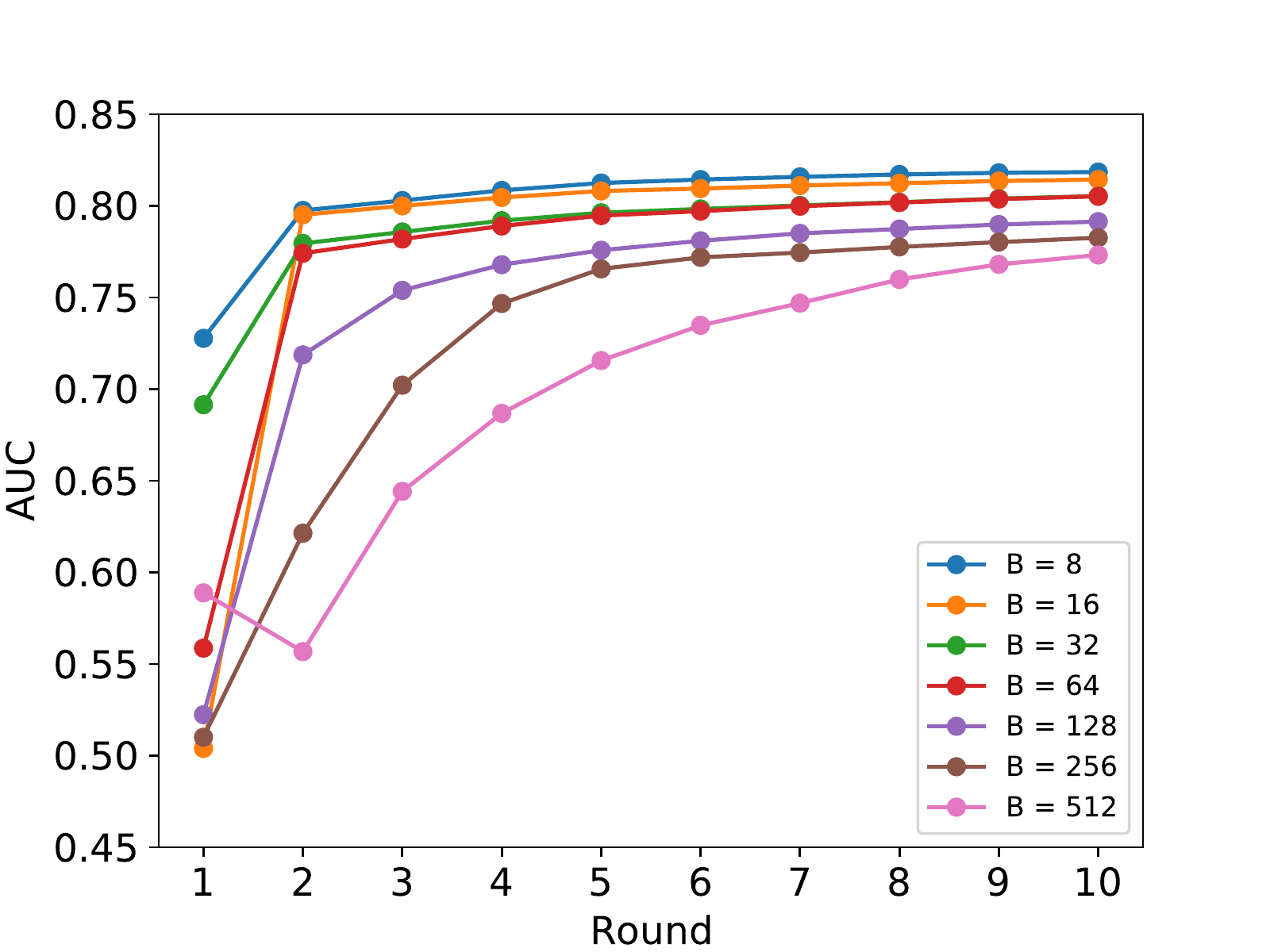}};
                        \end{tikzpicture}
                    }
                    \vspace{-0.7cm}
                    \caption{\scriptsize Scenario (7) \\ $(l = 50, u = 500)$} 
                    \label{fig:bc2_subfig1}
                \end{subfigure}
                \&
                \begin{subfigure}{0.35\columnwidth}
                    \centering
                    \resizebox{\linewidth}{!}{
                        \begin{tikzpicture}
                            \node (img)  {\includegraphics[width=\textwidth]{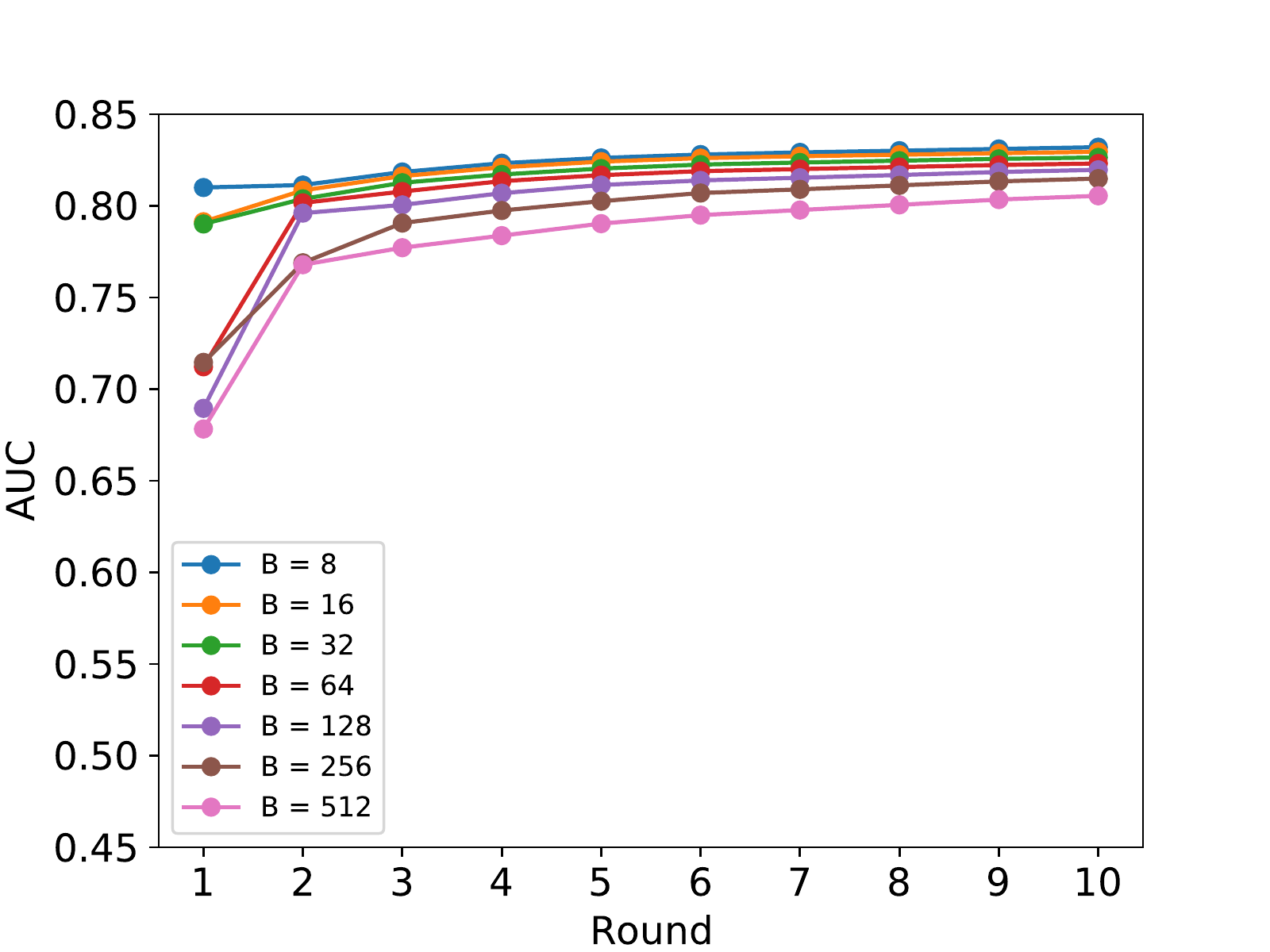}};
                        \end{tikzpicture}
                    }
                    \vspace{-0.7cm}
                    \caption{\scriptsize Scenario (9) \\$ (l = 50, u = 5000)$} 
                    \label{fig:bc2_subfig2}
                \end{subfigure}
                \&
                \begin{subfigure}{0.35\columnwidth}
                    \centering
                    \resizebox{\linewidth}{!}{
                        \begin{tikzpicture}
                            \node (img)  {\includegraphics[width=\textwidth]{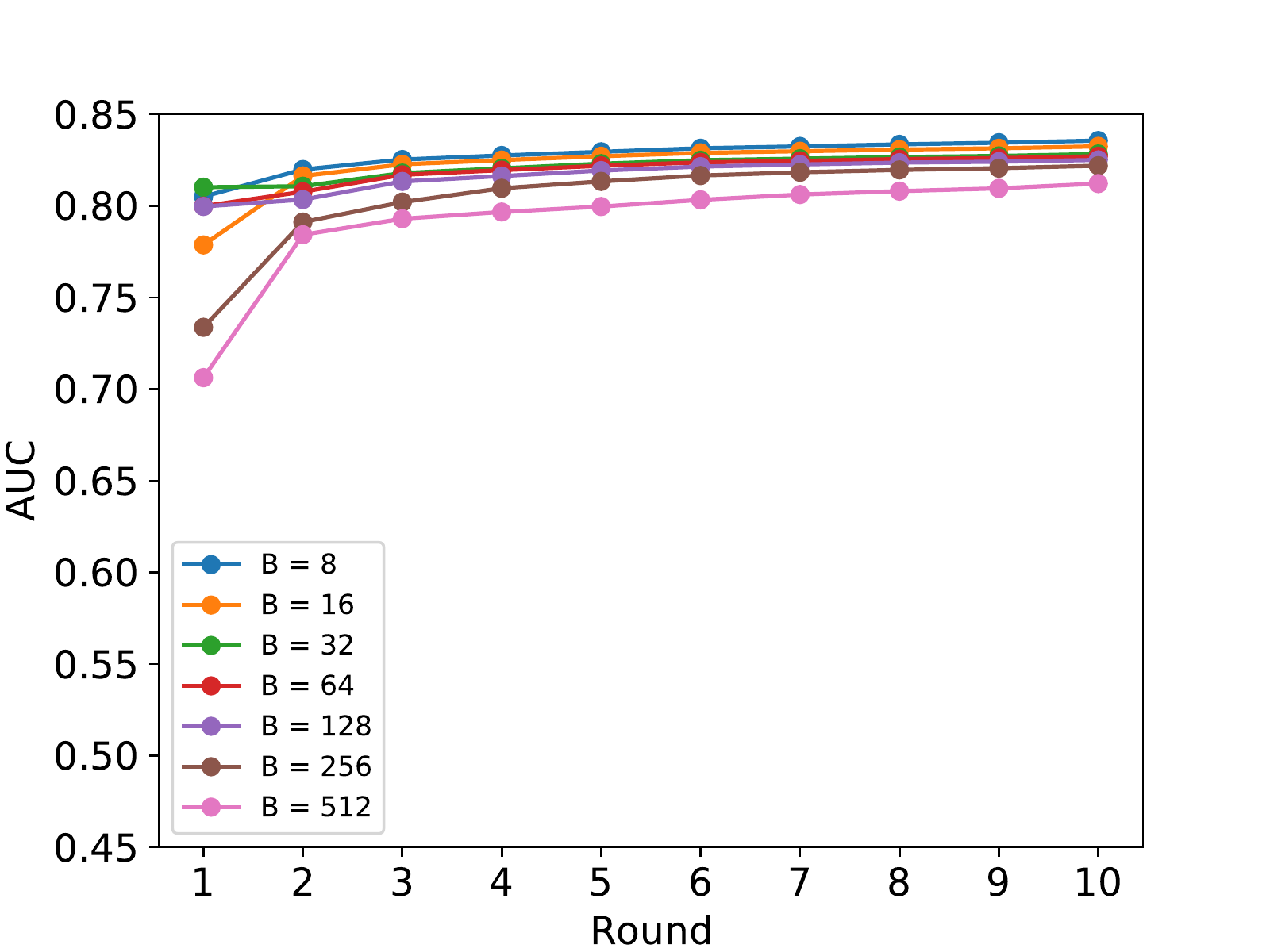}};
                        \end{tikzpicture}
                    }
                    \vspace{-0.7cm}
                    \caption{\scriptsize Scenario (14) \\$ (l = 500, u = 5000)$} 
                    \label{fig:bc2_subfig3}
                \end{subfigure}
                \&
            \\
                \begin{subfigure}{0.35\columnwidth}
                    \centering
                    \resizebox{\linewidth}{!}{
                        \begin{tikzpicture}
                            \node (img)  {\includegraphics[width=\textwidth]{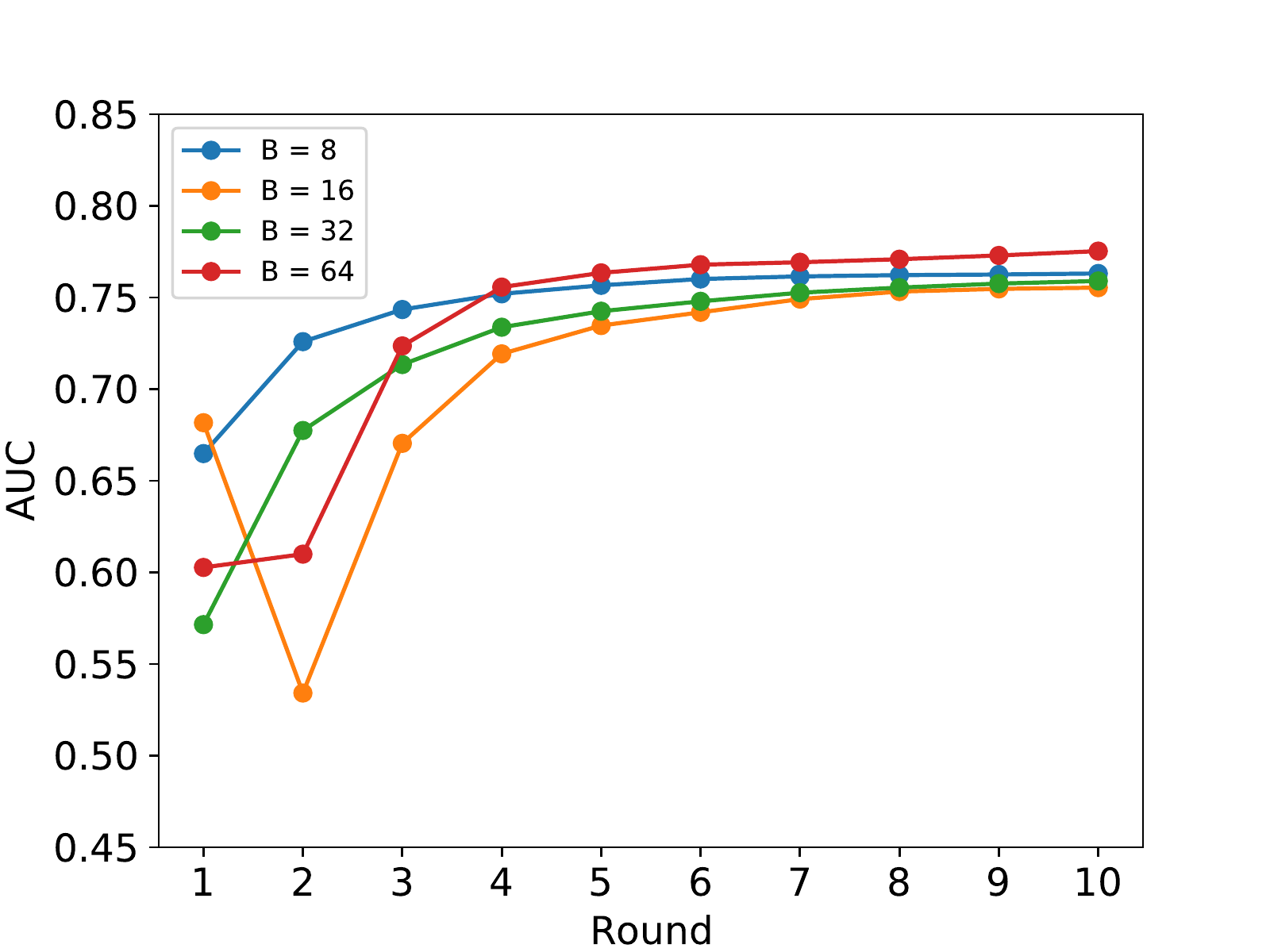}};
                        \end{tikzpicture}
                    }
                    \vspace{-0.7cm}
                    \caption{\scriptsize Scenario (16) \\$ (l = 5, u = 50)$} 
                    \label{fig:bc2_subfig4}
                \end{subfigure}
                \&
                \begin{subfigure}{0.35\columnwidth}
                    \centering
                    \resizebox{\linewidth}{!}{
                        \begin{tikzpicture}
                            \node (img)  {\includegraphics[width=\textwidth]{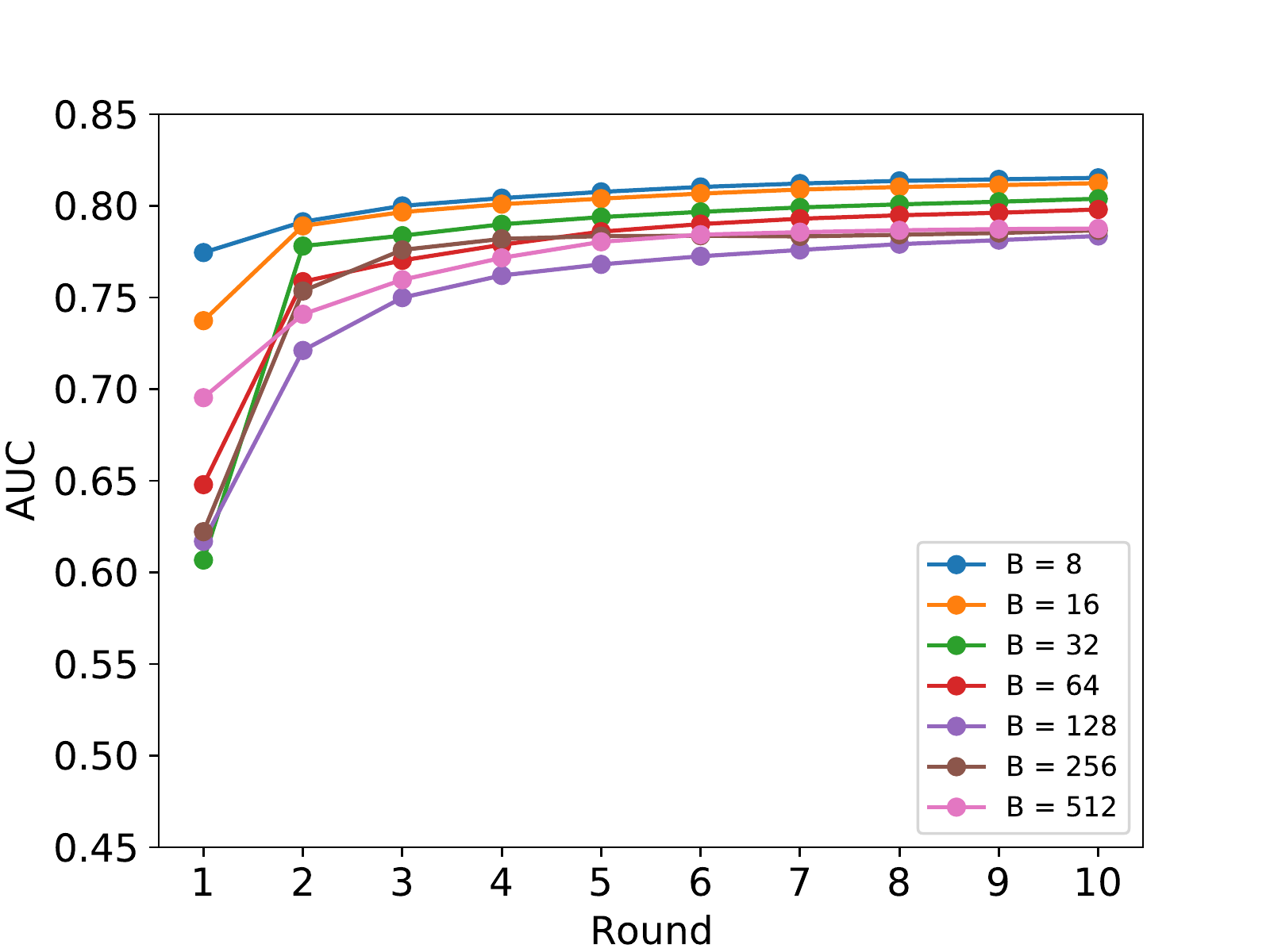}};
                        \end{tikzpicture}
                    }
                    \vspace{-0.7cm}
                    \caption{\scriptsize Scenario (17) \\$ (l = 5, u = 500)$} 
                    \label{fig:bc2_subfig5}
                \end{subfigure}
                \&
                \begin{subfigure}{0.35\columnwidth}
                    \centering
                    \resizebox{\linewidth}{!}{
                        \begin{tikzpicture}
                            \node (img)  {\includegraphics[width=\textwidth]{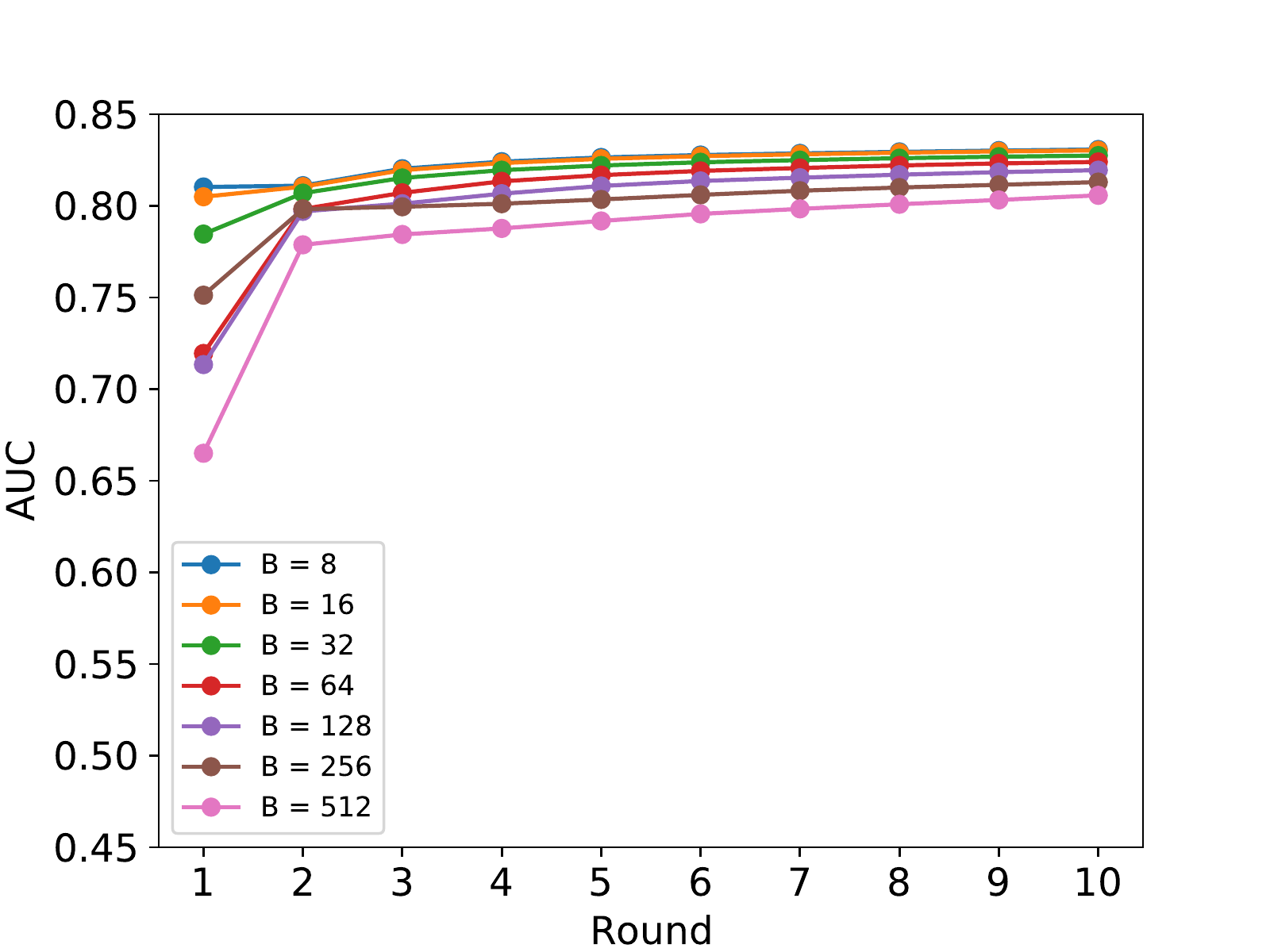}};
                        \end{tikzpicture}
                    }
                    \vspace{-0.7cm}
                    \caption{\scriptsize Scenario (18) \\$ (l = 5, u = 5000)$} 
                    \label{fig:bc2_subfig6}
                \end{subfigure}
                \&
                \\
                \begin{subfigure}{0.35\columnwidth}
                    \centering
                    \resizebox{\linewidth}{!}{
                        \begin{tikzpicture}
                            \node (img)  {\includegraphics[width=\textwidth]{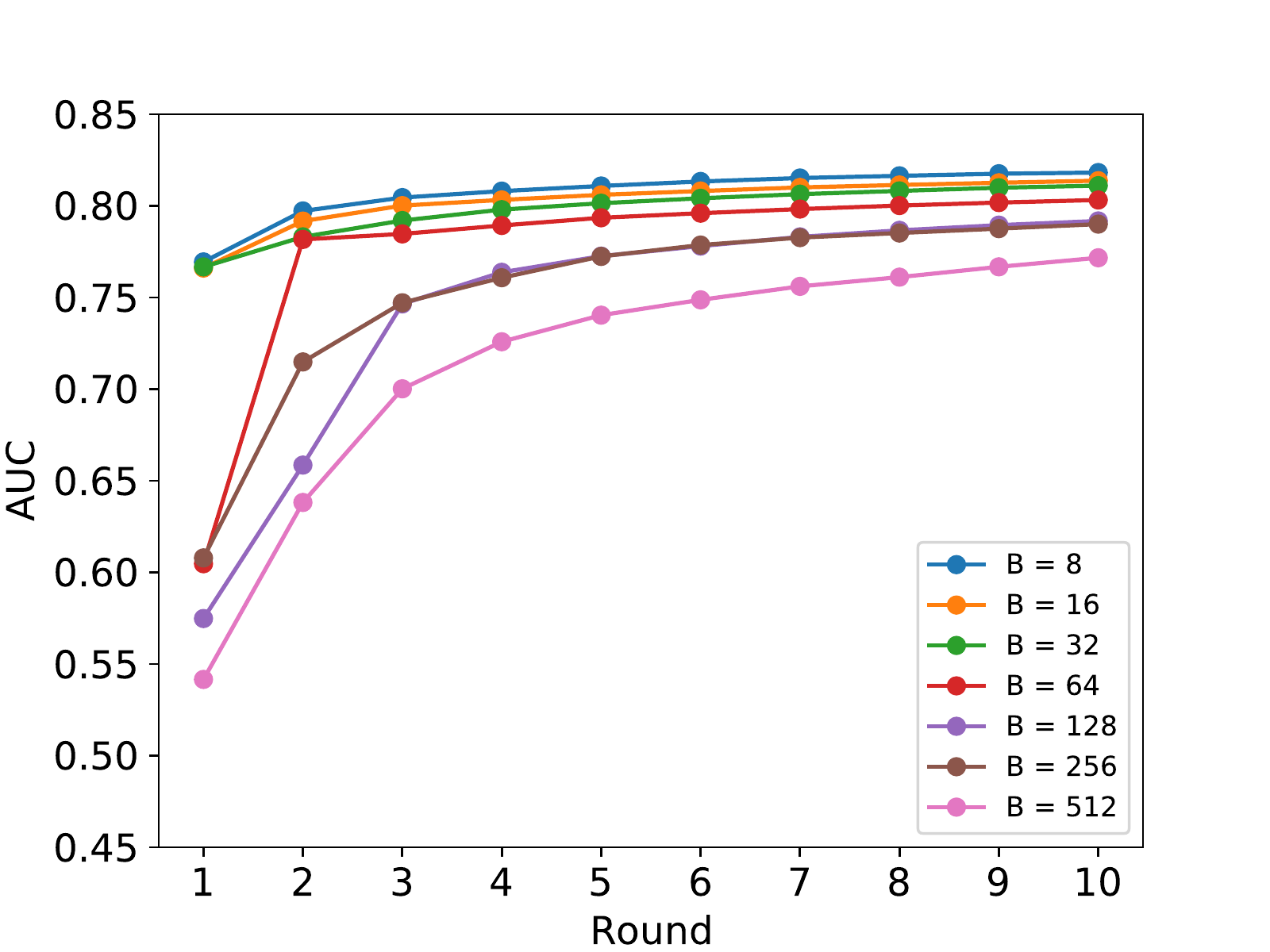}};
                        \end{tikzpicture}
                    }
                    \vspace{-0.7cm}
                    \caption{\scriptsize Scenario (7) \\$ (l = 50, u = 500)$} 
                    \label{fig:bc2_subfig7}
                \end{subfigure}
                \&
               \begin{subfigure}{0.35\columnwidth}
                    \centering
                    \resizebox{\linewidth}{!}{
                        \begin{tikzpicture}
                            \node (img)  {\includegraphics[width=\textwidth]{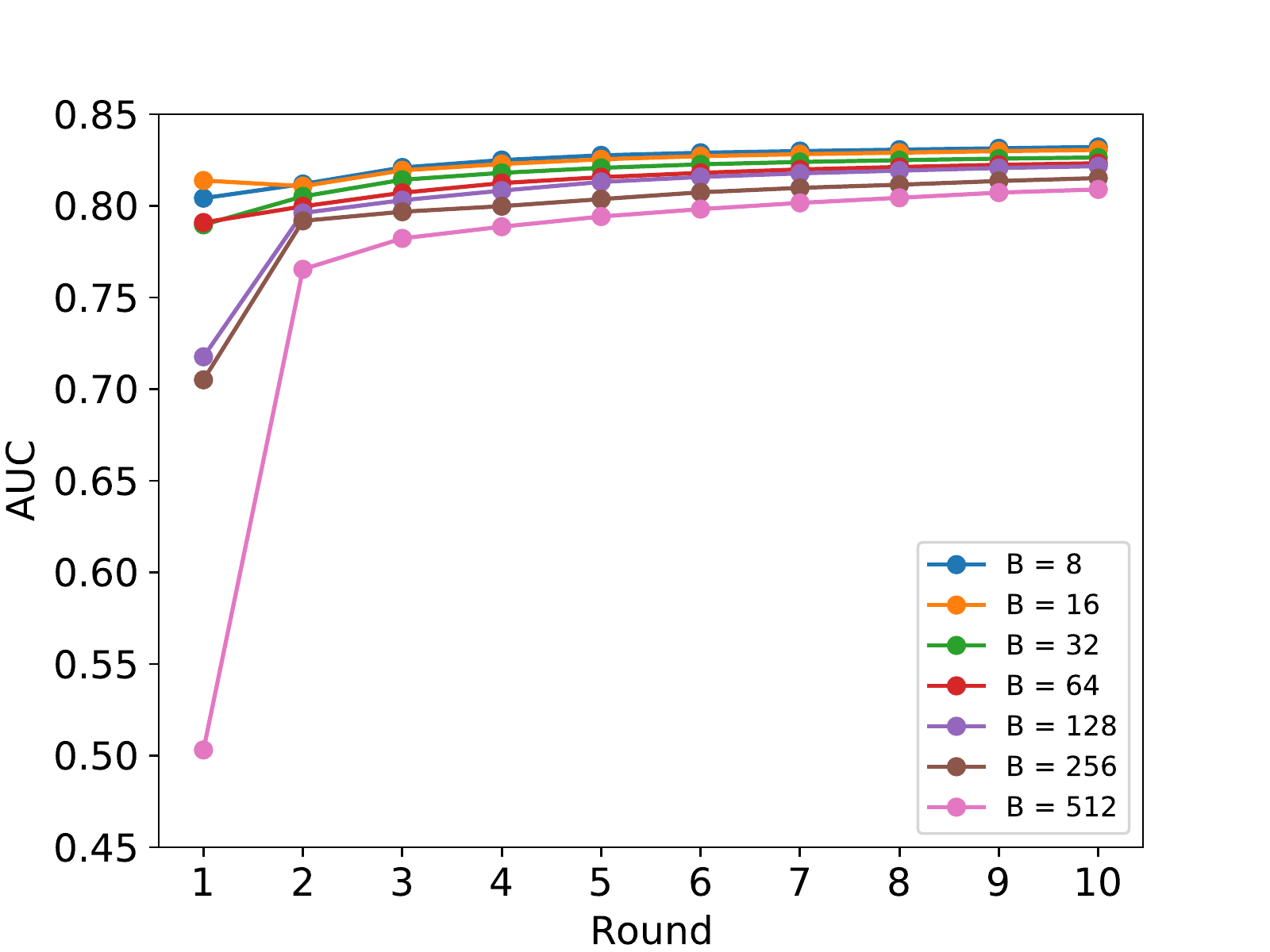}};
                        \end{tikzpicture}
                    }
                    \vspace{-0.7cm}
                    \caption{\scriptsize Scenario (9) \\$ (l = 50, u = 5000)$} 
                    \label{fig:bc2_subfig8}
                \end{subfigure}
                \&
                \begin{subfigure}{0.35\columnwidth}
                    \centering
                    \resizebox{\linewidth}{!}{
                        \begin{tikzpicture}
                            \node (img)  {\includegraphics[width=\textwidth]{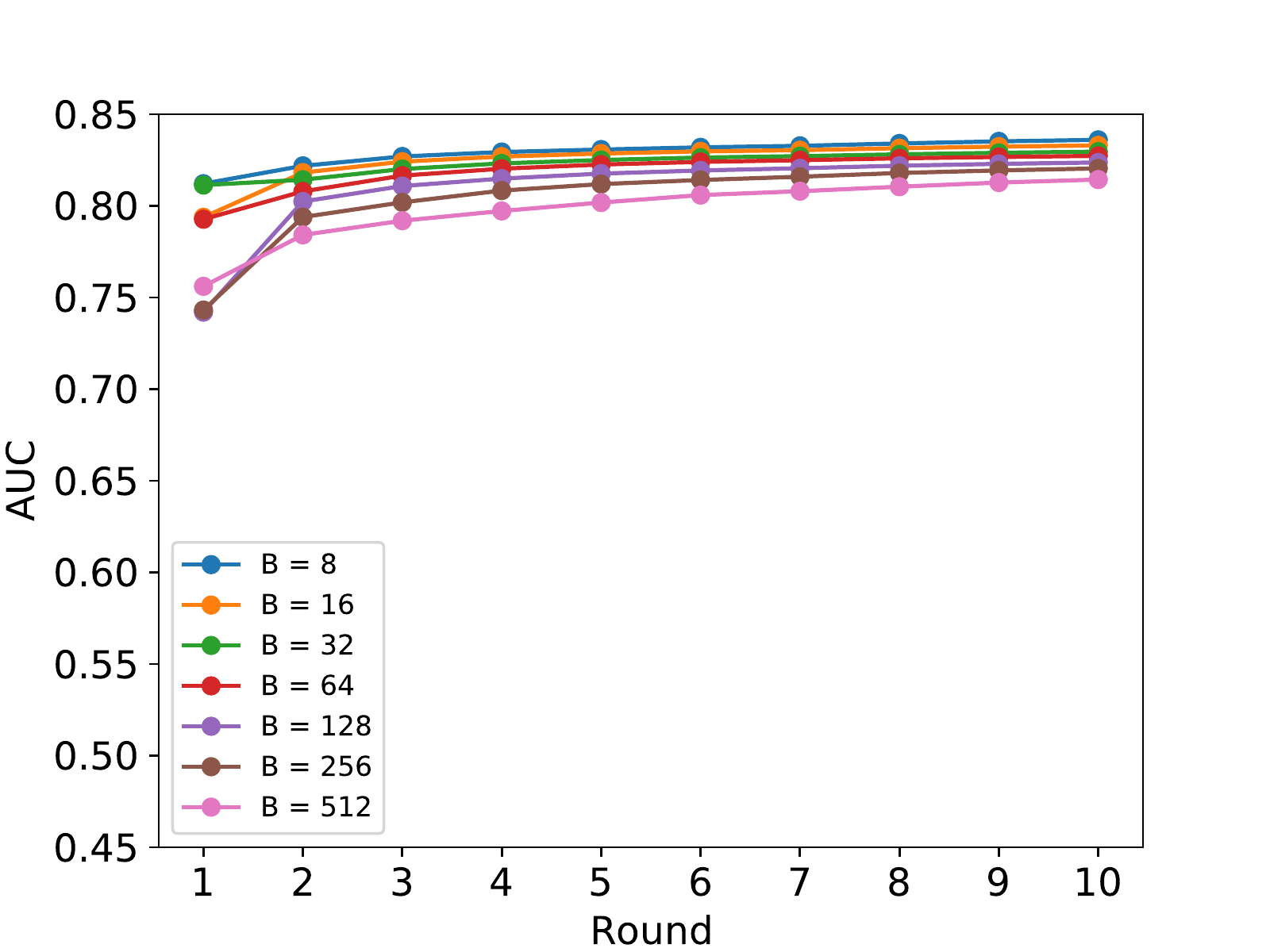}};
                        \end{tikzpicture}
                    }
                    \vspace{-0.7cm}
                    \caption{\scriptsize Scenario (14) \\$ (l = 500, u = 5000)$} 
                    \label{fig:bc2_subfig9}
                \end{subfigure}
                \&
                \\
                \begin{subfigure}{0.35\columnwidth}
                    \centering
                    \resizebox{\linewidth}{!}{
                        \begin{tikzpicture}
                            \node (img)  {\includegraphics[width=\textwidth]{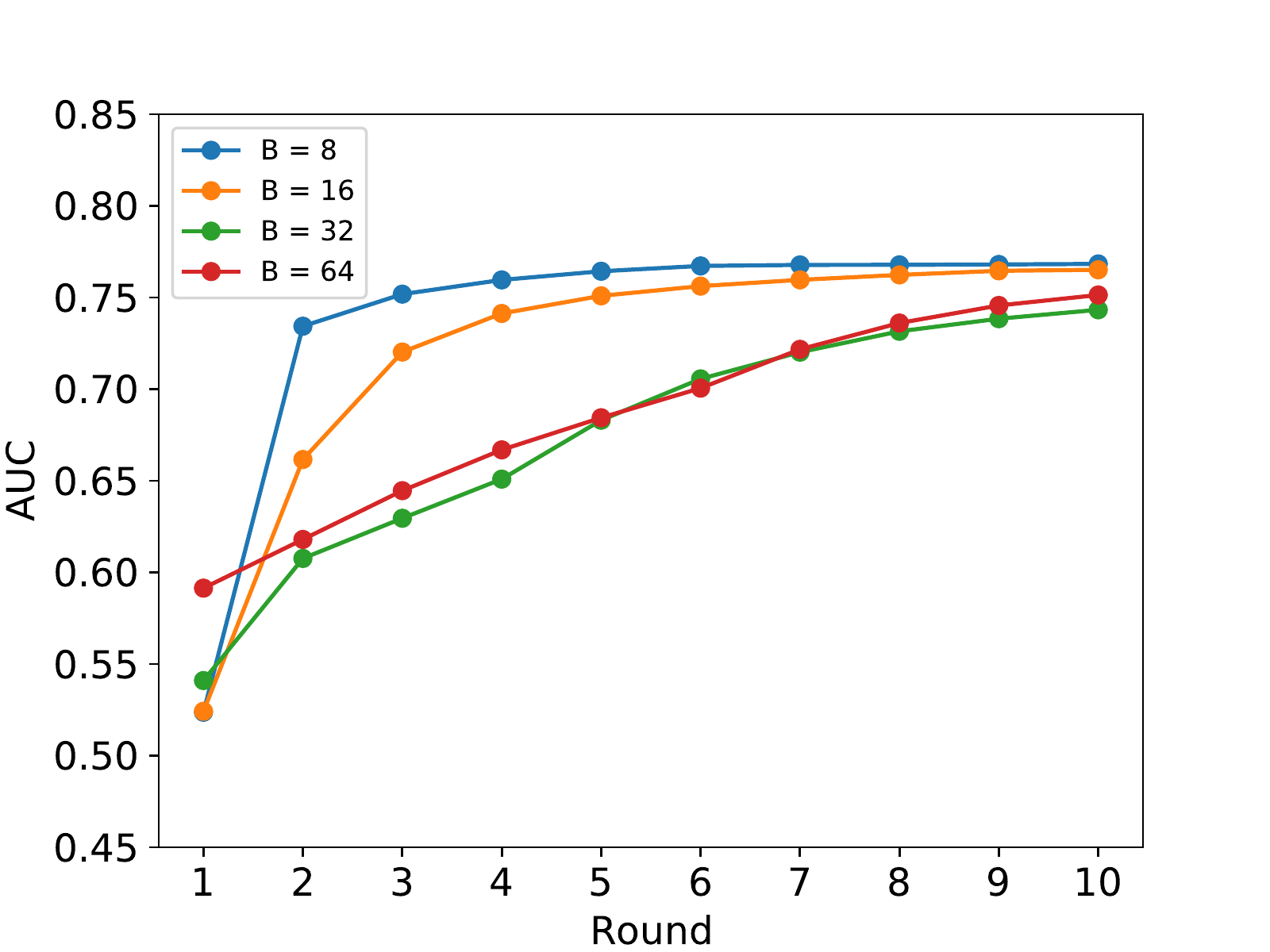}};
                        \end{tikzpicture}
                    }
                    \vspace{-0.7cm}
                    \caption{\scriptsize Scenario (16) \\$ (l = 5, u = 50)$} 
                    \label{fig:bc2_subfig10}
                \end{subfigure}
                \&
                \begin{subfigure}{0.35\columnwidth}
                    \centering
                    \resizebox{\linewidth}{!}{
                        \begin{tikzpicture}
                            \node (img)  {\includegraphics[width=\textwidth]{figs/Effect_of_B_C/predict_death_from_apache/5/500/1/dense.pdf}};
                        \end{tikzpicture}
                    }
                    \vspace{-0.7cm}
                    \caption{\scriptsize Scenario (17) \\$ (l = 5, u = 500)$} 
                    \label{fig:bc2_subfig11}
                \end{subfigure}
                \&
                \begin{subfigure}{0.35\columnwidth}
                    \centering
                    \resizebox{\linewidth}{!}{
                        \begin{tikzpicture}
                            \node (img)  {\includegraphics[width=\textwidth]{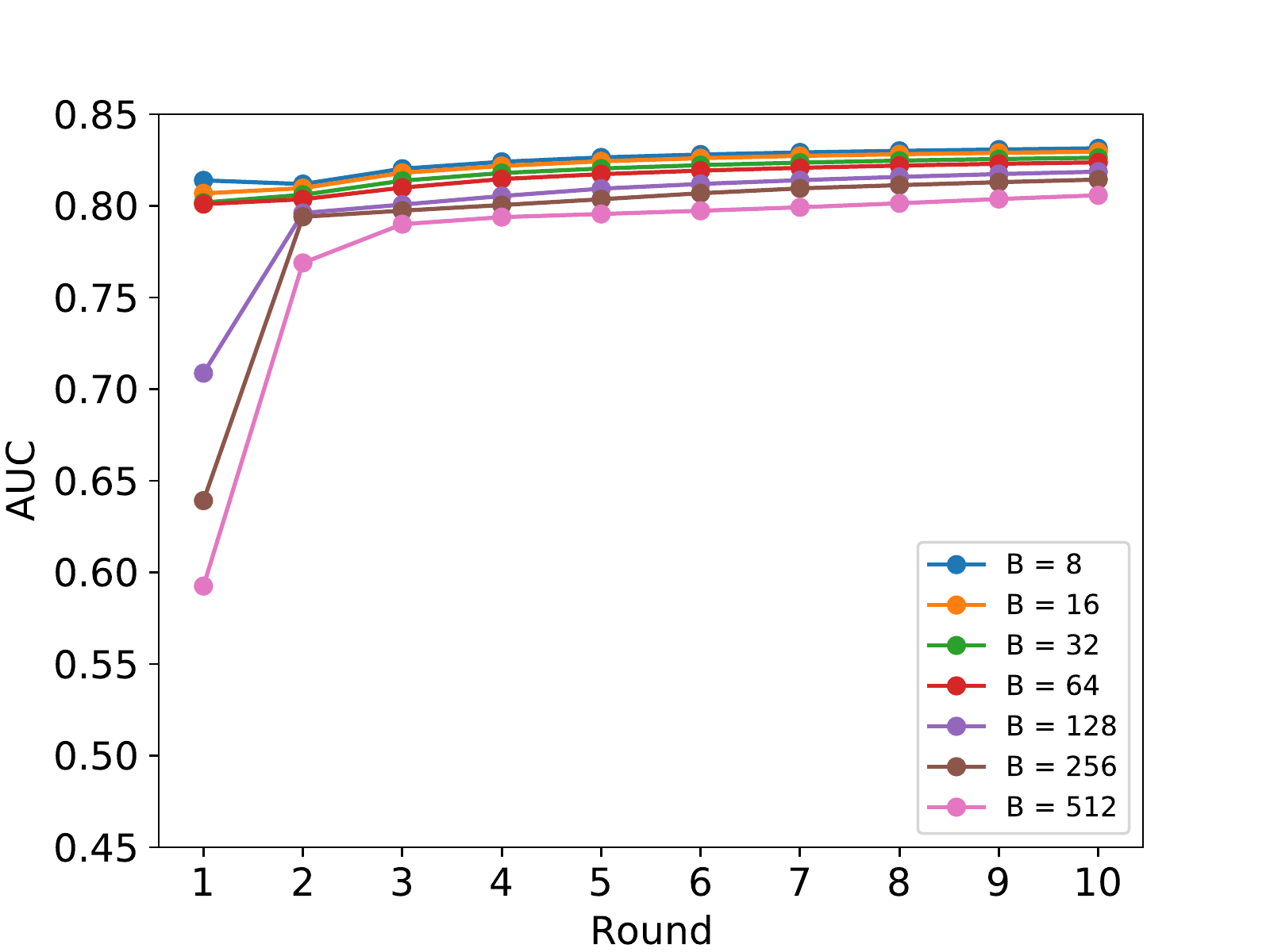}};
                        \end{tikzpicture}
                    }
                    \vspace{-0.7cm}
                    \caption{\scriptsize Scenario (18) \\$ (l = 5, u = 5000)$} 
                    \label{fig:bc2_subfig12}
                \end{subfigure}
                \&
                \\
                \begin{subfigure}{0.35\columnwidth}
                    \centering
                    \resizebox{\linewidth}{!}{
                        \begin{tikzpicture}
                            \node (img)  {\includegraphics[width=\textwidth]{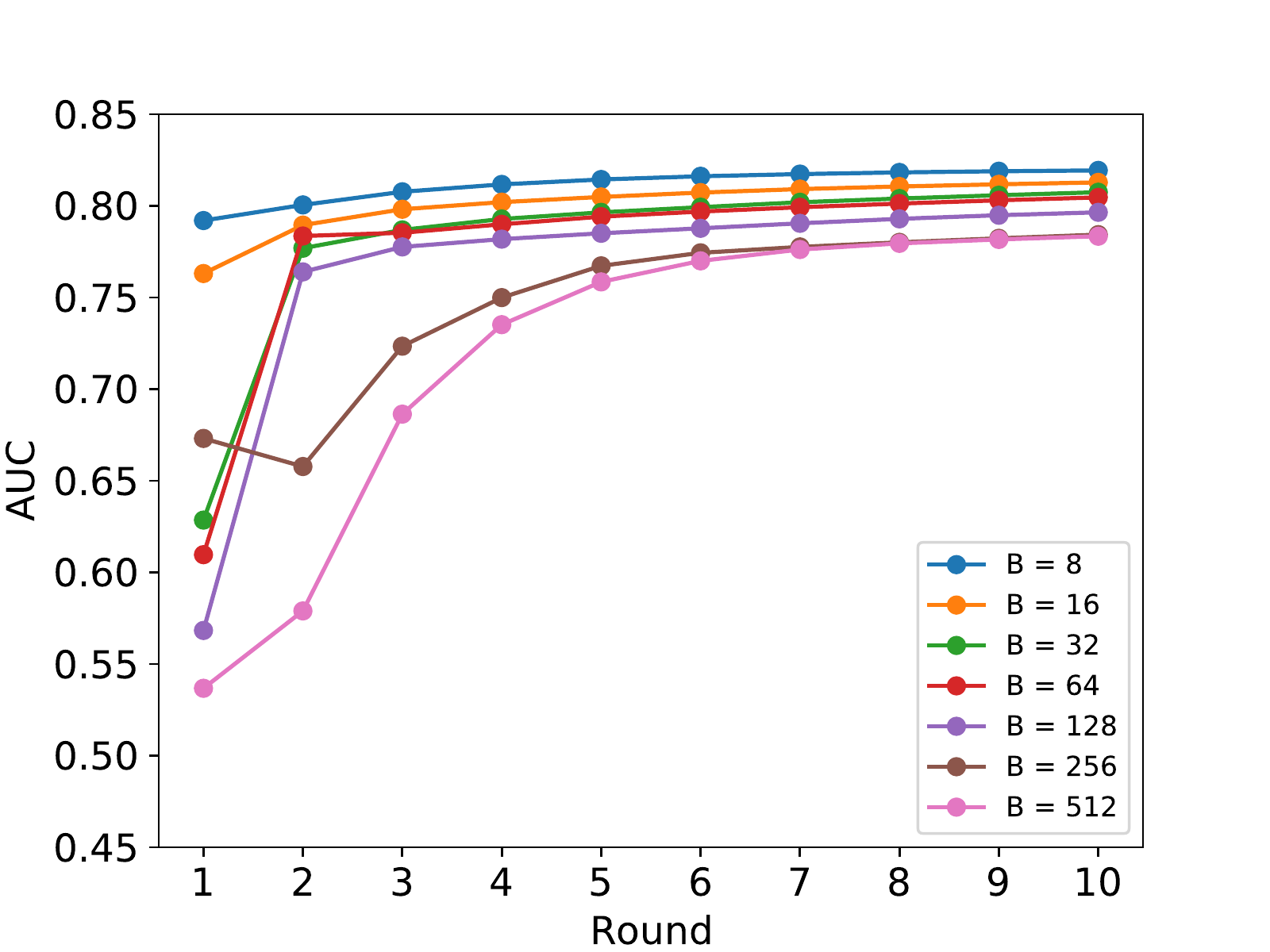}};
                        \end{tikzpicture}
                    }
                    \vspace{-0.7cm}
                    \caption{\scriptsize Scenario (7) \\$ (l = 50, u = 500)$} 
                    \label{fig:bc2_subfig13}
                \end{subfigure}
                \&
                \begin{subfigure}{0.35\columnwidth}
                    \centering
                    \resizebox{\linewidth}{!}{
                        \begin{tikzpicture}
                            \node (img)  {\includegraphics[width=\textwidth]{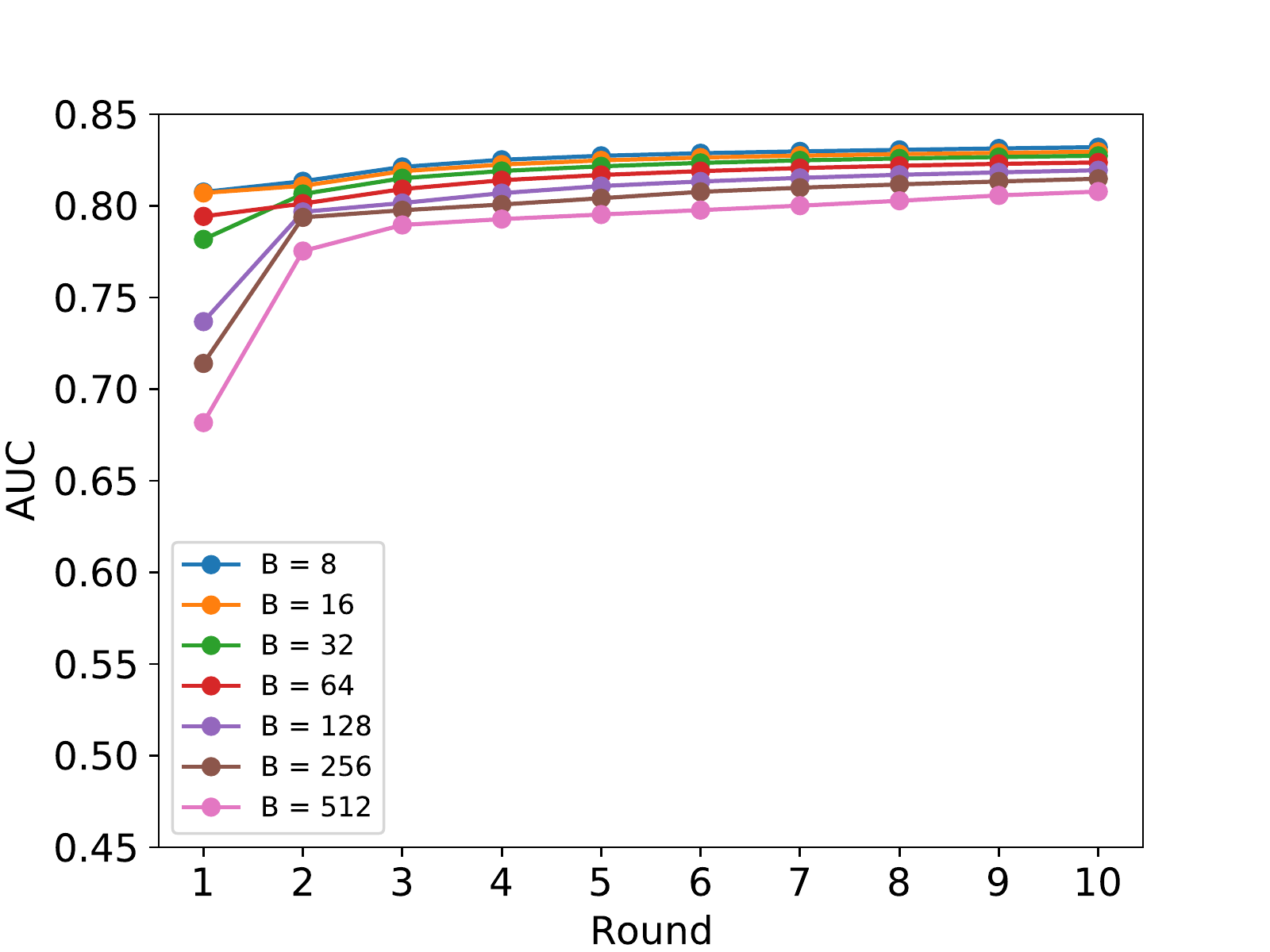}};
                        \end{tikzpicture}
                    }
                    \vspace{-0.7cm}
                    \caption{\scriptsize Scenario (9) \\$ (l = 50, u = 5000)$} 
                    \label{fig:bc2_subfig14}
                \end{subfigure}
                \&
                \begin{subfigure}{0.35\columnwidth}
                    \centering
                    \resizebox{\linewidth}{!}{
                        \begin{tikzpicture}
                            \node (img)  {\includegraphics[width=\textwidth]{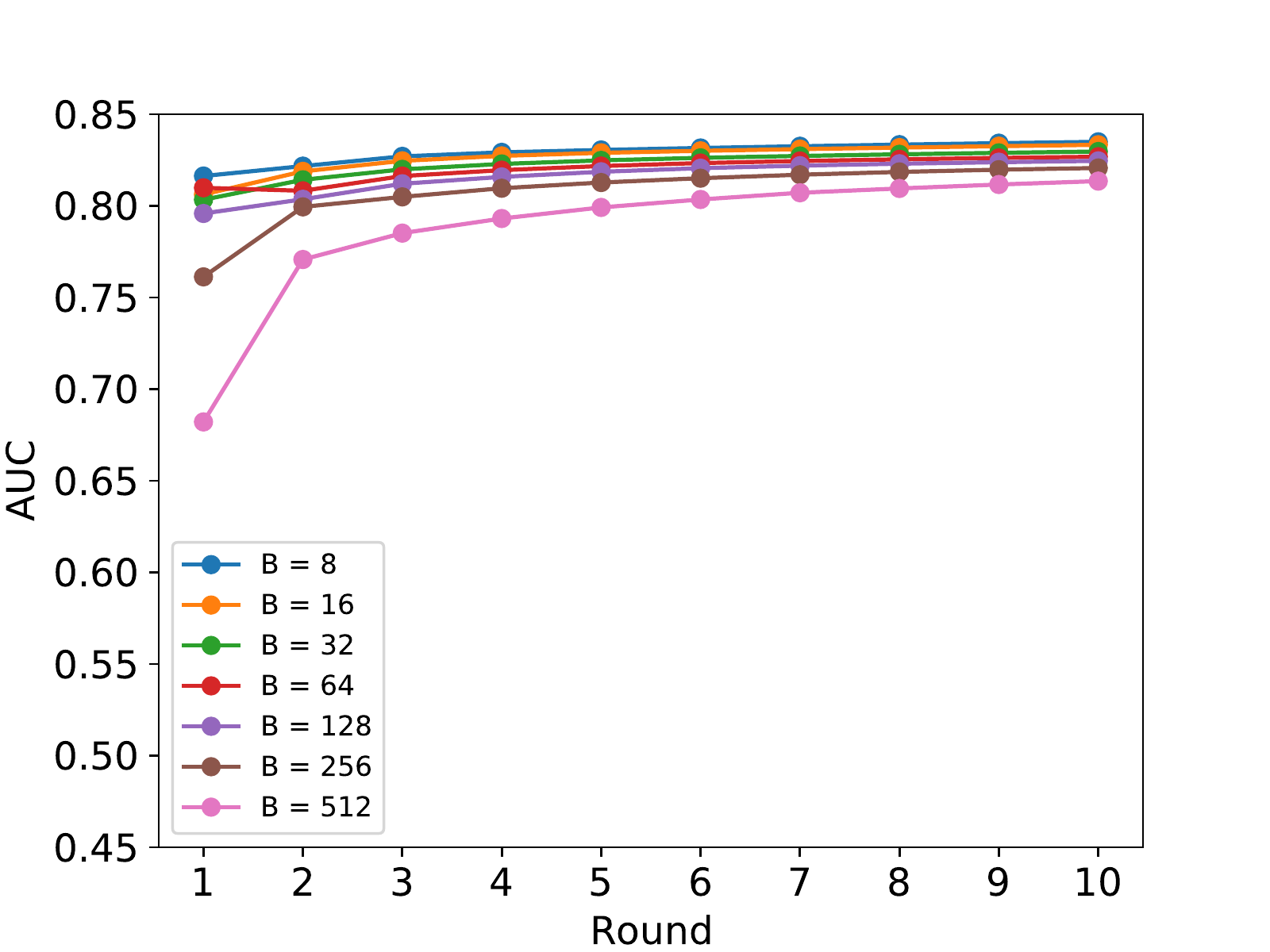}};
                        \end{tikzpicture}
                    }
                    \vspace{-0.7cm}
                    \caption{\scriptsize Scenario (14) \\$ (l = 500, u = 5000)$} 
                    \label{fig:bc2_subfig15}
                \end{subfigure}
                \&
                \\
            \\
            };
            \node [draw=none, rotate=90] at ([xshift=-1mm, yshift=2mm]fig-1-1.west) {\small $C=0.6$ };
            \node [draw=none, rotate=90] at ([xshift=-1mm, yshift=2mm]fig-2-1.west) {\small $C=0.8$ };
            \node [draw=none, rotate=90] at ([xshift=-1mm, yshift=2mm]fig-3-1.west) {\small $C=0.8$ };
            \node [draw=none, rotate=90] at ([xshift=-1mm, yshift=2mm]fig-4-1.west) {\small $C=1$ };
            \node [draw=none, rotate=90] at ([xshift=-1mm, yshift=2mm]fig-5-1.west) {\small $C=1$ };
        \end{tikzpicture}}
        \vspace{-0.5mm}
        \caption{Effect of B \& C on Federated Learning (2nd Part).}
        \vspace{-5mm}
        \label{fig:bc2}
    \end{figure*}